\documentclass[sigconf]{acmart}
\AtBeginDocument{%
  }

\setcopyright{cc}
\setcctype{by}
\copyrightyear{2026}
\acmYear{2026}
\acmDOI{10.1145/3767308.3835166}

\acmConference[MM '26]{Proceedings of the 34th ACM International Conference on Multimedia}{November 10--14, 2026}{Rio de Janeiro, Brazil}
\acmBooktitle{Proceedings of the 34th ACM International Conference on Multimedia (MM '26), November 10--14, 2026, Rio de Janeiro, Brazil}

\acmISBN{979-8-4007-2213-4/2026/11}

\acmSubmissionID{1509}

\definecolor{cvprblue}{rgb}{0.21,0.49,0.74}

\newcommand{\TODO}[1]{\textbf{\color{red}[TODO: #1]}}

\newcommand{\rr}[1]{{\textcolor{red}{#1}}}
\newcommand{\bb}[1]{{\textcolor{cvprblue}{#1}}}
\renewcommand{\TODO}[1]{}

\usepackage{color}

\usepackage{soul}
\usepackage{pifont}
\usepackage{amsthm}

\usepackage{colortbl}
\usepackage{makecell}

\newcommand{\cmark}{\ding{51}}%
\newcommand{\xmark}{\ding{55}}%

\definecolor{sh_gray}{rgb}{0.84,0.84,0.84}
\definecolor{sh_gray2}{rgb}{1,0.89,0.75}
\definecolor{color3}{rgb}{0.95,0.95,0.95}
\definecolor{color4}{rgb}{0.96,0.96,0.86}
\definecolor{color5}{rgb}{0.90,0.90,0.90}

\DeclareRobustCommand{\eg}{\emph{e.g.}}
\DeclareRobustCommand{\ie}{\emph{i.e.}}

\usepackage{adjustbox}
\usepackage{subcaption}

\usepackage{tabularx}
\usepackage{multirow}

\setuldepth{foobar}

\begin{document}

\title{HonestFace: Towards Honest Face Restoration with One-Step Diffusion Model}

\author{Jingkai Wang}
\email{jingkaiwang100@gmail.com}
\author{Wu Miao}
\email{miaowu123@sjtu.edu.cn}
\affiliation{%
  \department{School of Computer Science}
  \institution{Shanghai Jiao Tong University}
  \city{Shanghai}
  \country{China}}

\author{Jue Gong}
\email{juegong.me@gmail.com}
\author{Zheng Chen}
\email{zhengchen.cse@gmail.com}
\affiliation{%
  \department{School of Computer Science}
  \institution{Shanghai Jiao Tong University}
  \city{Shanghai}
  \country{China}}

\author{Xing Liu}
\email{xing.liu@vivo.com}
\author{Hong Gu}
\email{guhong@vivo.com}
\affiliation{%
  \department{vivo BlueImage Lab}
  \institution{vivo Mobile Communication Co., Ltd.}
  \city{Hangzhou}
  \country{China}}

\author{Yutong Liu}
\authornote{Corresponding authors: Yutong Liu and Yulun Zhang.}
\affiliation{%
  \department{School of Computer Science}
  \institution{Shanghai Jiao Tong University}
  \city{Shanghai}
  \country{China}}
\email{isabelleliu@sjtu.edu.cn}

\author{Yulun Zhang}
\authornotemark[1]
\affiliation{%
  \department{School of Computer Science}
  \institution{Shanghai Jiao Tong University}
  \city{Shanghai}
  \country{China}}
\email{yulun100@gmail.com}

\renewcommand{\shortauthors}{Wang et al.}

\begin{abstract}
Face restoration has achieved significant advancements through the years of development. However, maintaining high fidelity and authenticity while avoiding artifacts remains challenging, especially in extreme degradation scenarios. This highlights the need for models that are more ``honest'' in their reconstruction from low-quality inputs, accurately reflecting original characteristics.
In this work, we propose \textbf{HonestFace}, a novel approach for honest face restoration with identity consistency and realistic textures. To achieve this, HonestFace incorporates several key components. First, we propose an identity embedder to effectively capture and preserve crucial identity features from both the low-quality input and multiple reference faces. Second, a masked face alignment method is presented to enhance fine-grained details and textural authenticity.
Furthermore, we present a new real-world reference-based face restoration dataset, \textbf{MultiRefCeleb-Test}, with multiple high-quality references and low-quality inputs.
Leveraging these contributions within a one-step diffusion model framework, HonestFace delivers excellent restoration results in terms of facial fidelity and realism. Experiments demonstrate that our approach surpasses state-of-the-art methods, achieving superior performance in both visual quality and quantitative assessments. The code and pre-trained models are available at \url{https://github.com/jkwang28/HonestFace}.
\end{abstract}

\begin{teaserfigure}
  \includegraphics[width=\textwidth]{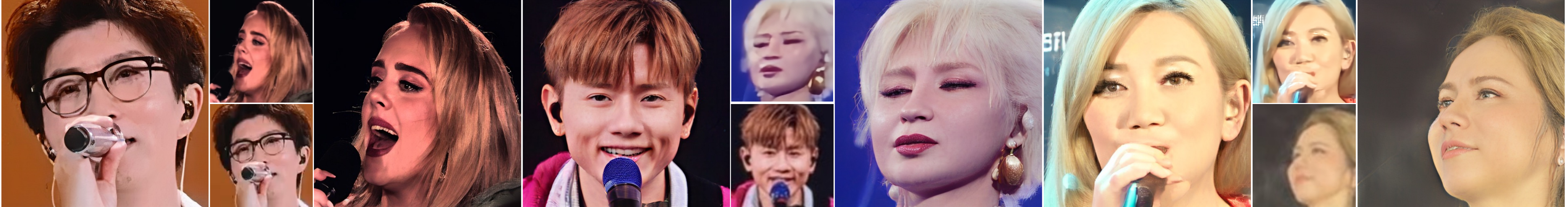}
  \vspace{-6mm}
  \caption{Real-world concert scenes captured by smartphones. From left to right: Buyi Mao, Adele, Jason Zhang, Bibi Zhou, Fish Leong, and G.E.M. Concerts are typically held in the evening, and photos are predominantly taken using telephoto lenses. For celebrities like these, who have abundant high-definition images available online, smartphone manufacturers can directly embed high-resolution ID information, enabling automatic matching and background processing when users take photos.}
  \vspace{2mm}
  \label{fig:teaser}
\end{teaserfigure}

\maketitle

\vspace{-2mm}
\section{Introduction}
\label{sec:intro}

Face restoration is an ill-posed problem that aims to recover a high-quality (HQ) face image from a given low-quality (LQ) input. In practice, LQ face images often suffer from severe degradations (\eg, blur, noise, compression artifacts) that result in significant loss of detail. This loss of information makes accurate restoration extremely challenging, particularly when recovering fine details and preserving the subject’s identity. Existing methods seek to improve face restoration performance, including models based on Transformers~\cite{zhou2022codeformer,wang2023restoreformer++,xie2024pltrans,tsai2024daefr}, GANs~\cite{ChenPSFRGAN,wang2021gfpgan,Yang2021GPEN,chan2021glean}, and diffusion models~\cite{wang2023dr2,chen2023BFRffusion,qiu2023diffbfr,Suin2024CLRFace,wu2024osediff,lin2024diffbir,yue2024difface,ying2024restorerid,wang2025osdface,zhang2025instantrestore}. These methods have significantly improved the quality of restored faces, marking notable milestones in low-level vision and practical portrait enhancement.

\begin{figure}[t]
\begin{center}
\small
\scalebox{1}{
    \hspace{-0.4cm}
    \begin{adjustbox}{valign=t}
    \begin{tabular}{cccc}
    \includegraphics[width=0.24\columnwidth]{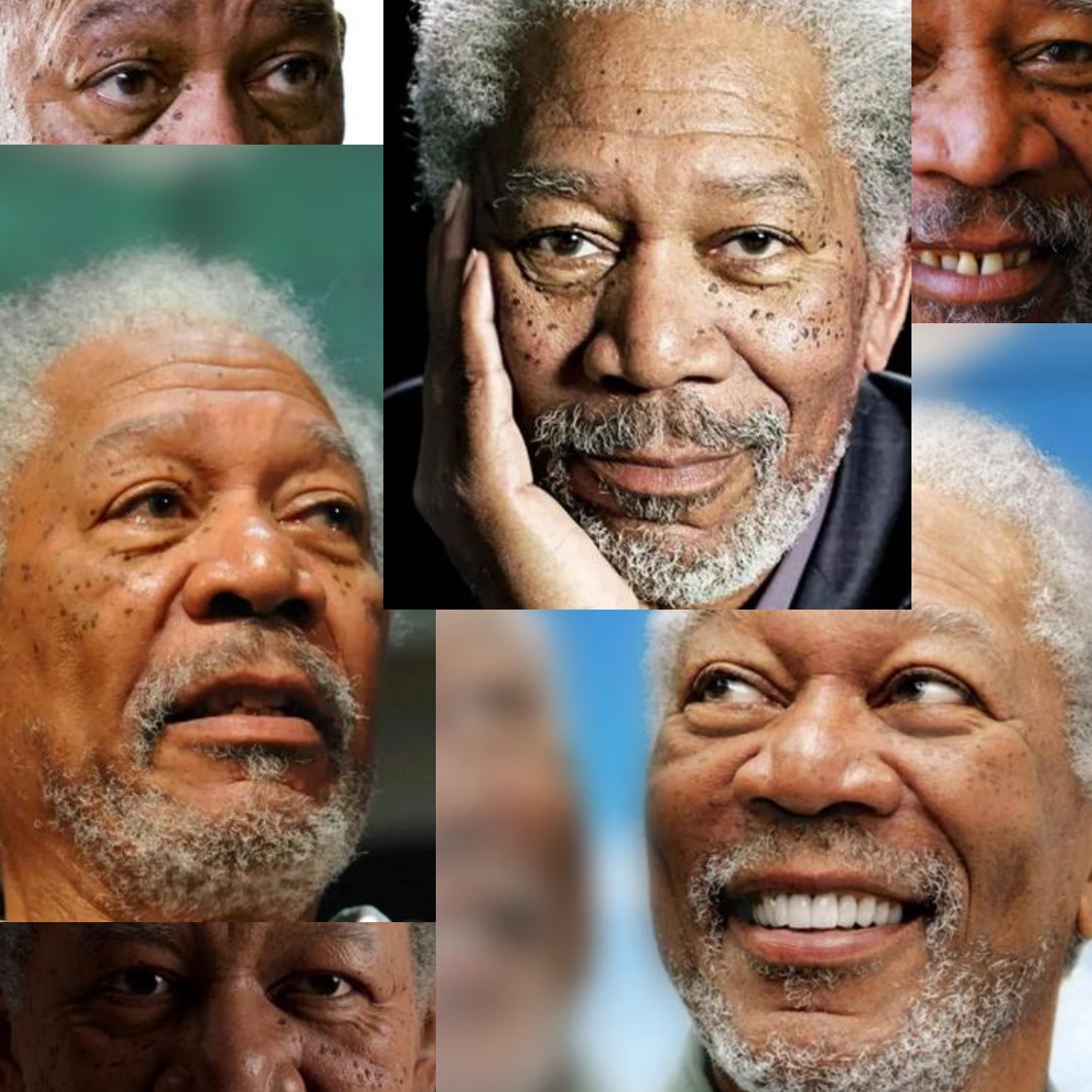} \hspace{-4mm} &
    \includegraphics[width=0.24\columnwidth]{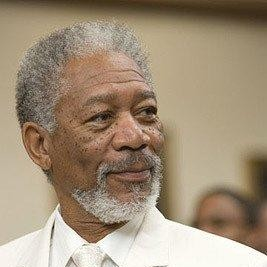} \hspace{-4mm} &
    \includegraphics[width=0.24\columnwidth]{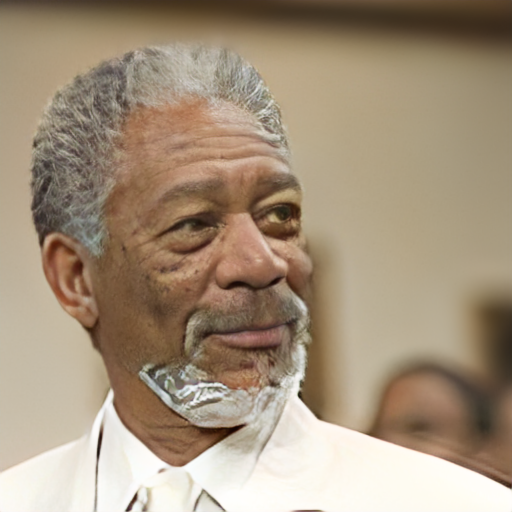} \hspace{-4mm} &
    \includegraphics[width=0.24\columnwidth]{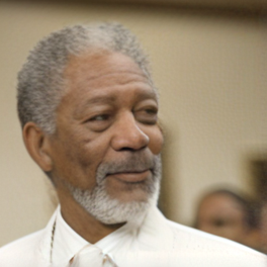} \hspace{-4mm}
    \vspace{-0.5mm} \\
    Reference \hspace{-4mm} &
    LQ (512$\times$512) \hspace{-4mm} &
    ASFFNet~\cite{li2020asffnet512} \hspace{-4mm} &
    FaceMe~\cite{liu2025faceme} \hspace{-4mm} \\
    \includegraphics[width=0.24\columnwidth]{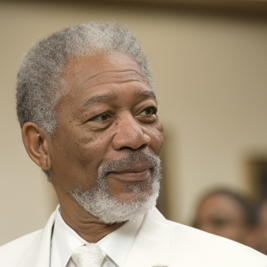} \hspace{-4mm} &
    \includegraphics[width=0.24\columnwidth]{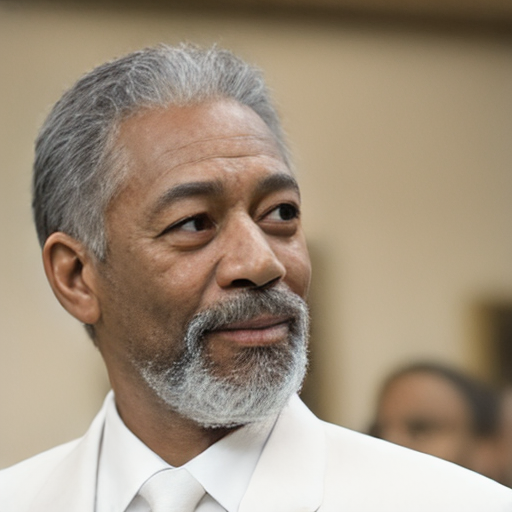} \hspace{-4mm} &
    \includegraphics[width=0.24\columnwidth]{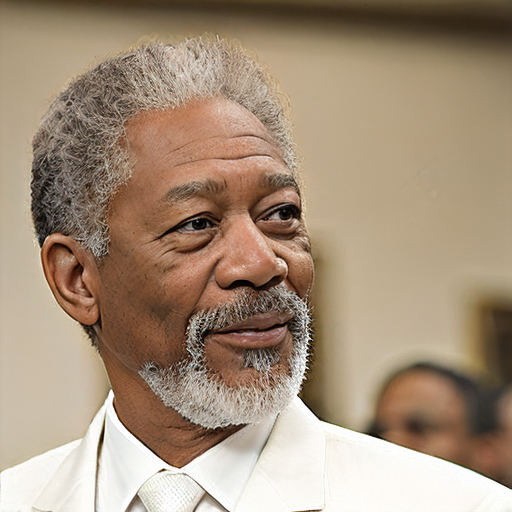} \hspace{-4mm} &
    \includegraphics[width=0.24\columnwidth]{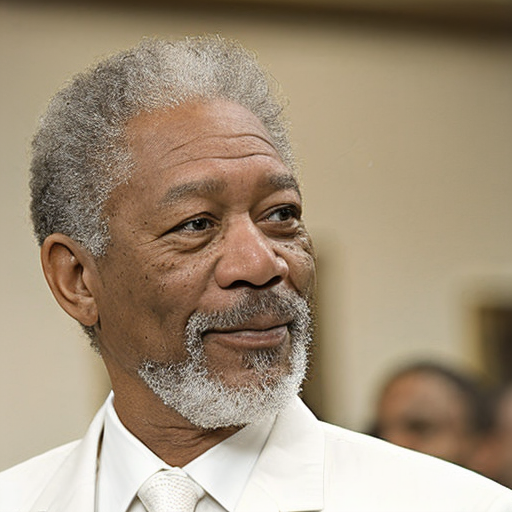} \hspace{-4mm} 
    \vspace{-0.5mm} \\
    DiffBIR~\cite{lin2024diffbir} \hspace{-4mm} &
    OSEDiff~\cite{wu2024osediff} \hspace{-4mm} &
    OSDFace~\cite{wang2025osdface} \hspace{-4mm} &
    HonestFace (ours) \hspace{-4mm} \\
    \end{tabular}
    \end{adjustbox}
}
\end{center}
\vspace{-4mm}
\caption{Comparisons of face restoration methods. HonestFace restores a more natural, faithful and ID-consistent face.}
\label{fig:page1_compare}
\vspace{-7mm}
\end{figure}

However, a key requirement for an ``honest'' restored face is a natural appearance with authentic textures. As shown in Fig.~\ref{fig:page1_compare}, in practice, many generative restoration models produce results that, while sharp, appear unnatural. A common issue is over-smoothing: skin may look overly airbrushed, lacking pores and fine wrinkles, and hair often appears too uniform, missing the irregular details of real hair. For example, the diffusion-based model DiffBIR~\cite{lin2024diffbir}, despite its powerful generative prior, often yields overly smooth, even portrait-like faces. Similarly, recent transformer-based models like CodeFormer~\cite{zhou2022codeformer} and DAEFR~\cite{tsai2024daefr} sometimes produce artifacts in certain regions. In particular, their restored hair textures can appear artificial or repetitive, lacking the natural randomness of real hair strands. Another common issue is color shift, where the restored image color deviates significantly from the input. Some methods~\cite{liu2025faceme,wu2024osediff} attempt to address this using AdaIN or wavelet-based techniques, but the problem remains largely unresolved. These limitations highlight the ongoing challenge of generating truly natural-looking results without sacrificing detail.

Moreover, a practical face restoration model for real-world applications must handle diverse levels of degradation. Maintaining identity fidelity is especially challenging when dealing with severely degraded inputs. The loss of identity cues and the absence of strong prior constraints render the solution space underdetermined. This leads to numerous plausible local optima and causes restored faces to deviate from the true appearance. To address this, reference-based methods have been proposed~\cite{li2018gfrnet,li2020asffnet512,li2020blind,tao2025overcoming,varanka2024pfstorer}, which utilize an additional high-quality reference image to guide restoration. Incorporating multiple references can offer even richer identity information. Existing approaches typically select the most reliable single reference~\cite{dogan2019exemplar,li2020asffnet512} or average multiple references~\cite{nitzan2022mystyle}; however, these methods do not fully exploit all available reference data. In summary, an ``honest'' restoration must achieve two key objectives: perceptual naturalness (producing realistic, natural textures) and identity fidelity (maintaining consistency with the target ID).

In this paper, we propose a novel face restoration method, \textit{HonestFace}, designed to address the challenges mentioned above. 
\textbf{First}, to enhance identity consistency, we present the identity embedder (IDE), which combines a facial feature extractor capturing fine textures (e.g., wrinkles, eye color), and a face identity encoder preserving overall identity. This dual-path design effectively maintains comprehensive identity information and person-specific details.
\textbf{Second}, we propose masked face alignment (MFA) to enhance perceptual realism. MFA generates a heatmap of facial landmarks and uses it as a mask in the alpha channel. This approach emphasizes regions crucial to human perception, like wrinkles and skin structure, producing more authentic textures.
\textbf{Third}, we introduce a new real-world reference-based face restoration dataset, \textit{MultiRefCeleb-Test}, which includes diverse degradation types and multiple HQ reference images per LQ input. This dataset serves as a valuable benchmark for evaluating performance on authentic degradations, rather than synthetic ones.
\textbf{Finally}, with our proposed techniques, we establish HonestFace, a one-step diffusion model that accepts multiple reference images. In particular, our method avoids the over-smoothed, ``plastic'' appearance and the repetitive artifacts commonly seen in prior methods. Our method achieves state-of-the-art results, preserving identity fidelity while generating realistic skin textures and hair details across diverse degradation levels.

Our main contributions are summarized as follows.
\vspace{-0.3mm}\begin{itemize}
\item We introduce the identity embedder (IDE), consisting of a facial feature extractor and a face identity encoder, to preserve and extract the key ID facial features.
\item We present the masked face alignment (MFA) method to enhance the naturalness of the restored images. By applying a heatmap-based mask on the alpha channel, MFA ensures that restoration efforts are strategically concentrated on perceptually important areas throughout the restoration process.
\item We publish \textit{MultiRefCeleb-Test}, the first large-scale real-world reference-based benchmark (400+ identities, 9,000+ images) for face restoration with authentic degradations.
\item We propose \textit{HonestFace}, a one-step diffusion model that uses multiple reference images. Our method prevents common issues such as over-smoothing and repetitive artifacts, delivering ``honest'' face restoration that simultaneously preserves both identity and realistic textures.
\end{itemize}

\vspace{-2.8mm}
\section{Related Work}
\label{sec:related}
\vspace{-0.8mm}
\subsection{Reference-based Face Restoration}
\vspace{-0.2mm}
Faces are arguably the most detail-sensitive part of the human body and have been widely studied in image restoration~\cite{yu2018super,chen2018fsrnet,kim2019progressive,shen2018deep,menon2020pulse,gu2020image,wan2020bringing,yang2020hifacegan,wang2021gfpgan}. However, traditional methods often struggle with challenging tasks. To address identity information loss in LQ inputs, reference-based approaches leverage HQ images of the same individual as supplementary guidance~\cite{li2018learning,dogan2019exemplar}. For instance, ASFFNet~\cite{li2020asffnet512} selects an optimal reference and adaptively fuses features. DMDNet~\cite{li2022dmdnet} employs dual memory dictionaries to store general facial priors and identity-specific cues. MyStyle~\cite{nitzan2022mystyle} fine-tunes a pre-trained face generator to create personalized priors, while MyStyle++~\cite{zeng2023mystylepp} improves latent optimization. PFStorer~\cite{varanka2024pfstorer} fine-tunes new models using reference faces for identity-specific restoration. These methods enable high-fidelity reconstruction and even allow flexible attribute editing of specific facial features.

\begin{figure*}[htbp]
\centering
\includegraphics[width=\linewidth]{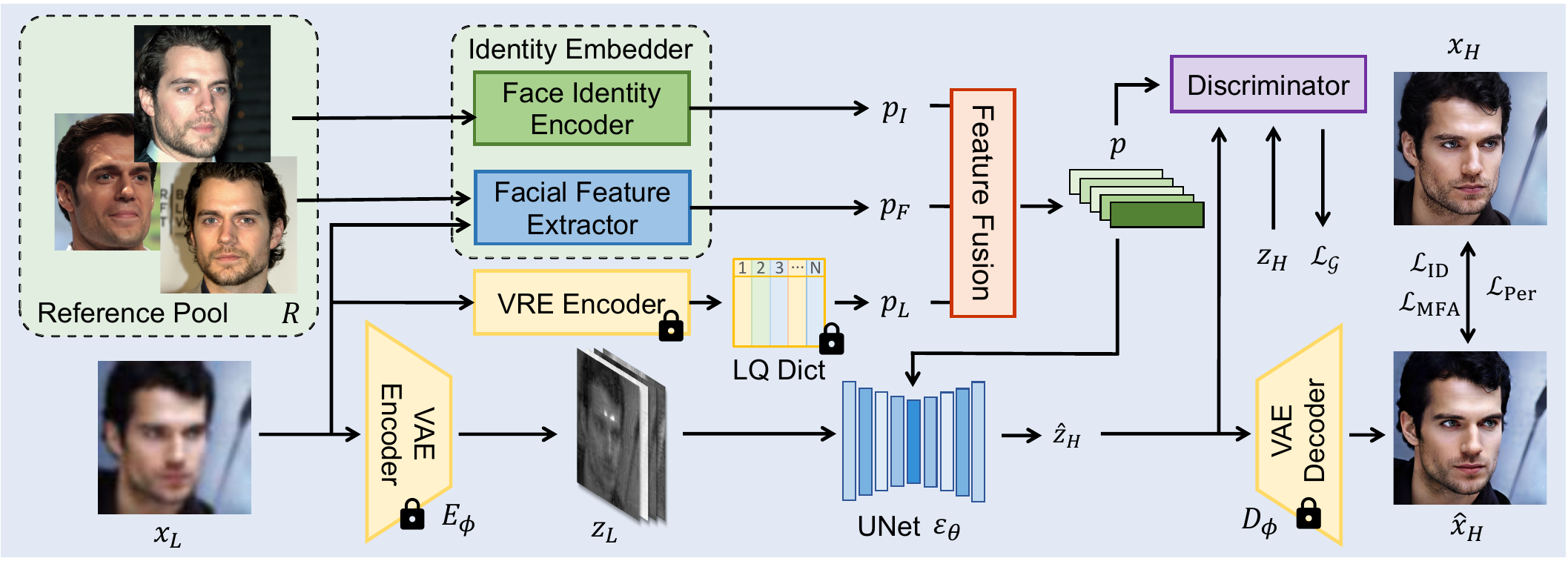}
\vspace{-5mm}
\caption{Overall training pipeline of HonestFace. First, the LQ input $x_L$ is encoded into $z_L$ by the VAE encoder $E_\phi$. Meanwhile, $x_L$ and HQ references $R = \{r_i\}$ pass through IDE and VRE, then fused into the prompt embedding $p$. Next, the UNet predicts $\varepsilon_\theta$ to estimate $\hat{z}_H$. Finally, the VAE decoder $D_\phi$ reconstructs the output $\hat{x}_H$. Generator and discriminator are trained alternately. }
\label{fig:overall}
\vspace{-2mm}
\end{figure*}

\vspace{-1.5mm}
\subsection{Diffusion Models}
\vspace{-0.2mm}

Latent diffusion models~\cite{rombach2022ldm} and related works~\cite{pmlr-v139-ramesh21a,chen2024pixartalpha,flux2024,xie2024sana} have driven progress in high-resolution image synthesis. Meanwhile, pre-trained diffusion models have been widely applied in low-level vision tasks~\cite{wang2024stablesr,wu2024seesr,yu2024supir,sun2024pisasr,dong2024tsdsr,li2025fluxsr,li2024dfosd,wu2025sasw,wang2026strsr}, including face image restoration.
For example, \citet{yang2023pgdiff} proposed partial guidance for diffusion models, focusing on modeling high-quality image properties instead of degradation. \citet{ding2024restoration} fine-tuned a pre-trained diffusion model with high-quality image sets to constrain the generative space. 
Ref-LDM~\cite{hsiao2024refldm} proposed CacheKV, which can be utilized repeatedly in each of the denoising timesteps.
FaceMe~\cite{liu2025faceme} introduced an identity encoder to guide restoration, while \citet{tao2025overcoming} combined text attributes, reference images, and identity information through a dual-control adapter. \citet{zhang2025instantrestore} applied a one-step diffusion model with an attention-sharing mechanism to achieve fast, personalized restoration without per-identity fine-tuning. \citet{wang2025osdface} introduced a one-step diffusion model with a visual representation embedder to capture prior information and input features better. These methods significantly improve face restoration performance in real-world scenarios.

\vspace{-0.5mm}
\section{Methods}
\label{sec:methods}
\vspace{-0.5mm}

\subsection{Model Formulation}\label{sec:model_formulation}

Face restoration (FR) aims to reconstruct a high-quality (HQ) face $x_H \in \mathbb{R}^{H \times W \times 3}$ from a low-quality (LQ) input $x_L \in \mathbb{R}^{H \times W \times 3}$. For reference-based FR, this process is augmented by a set of $N$ HQ reference images $R=\{r_1, r_2, \dots, r_N\}$, where each $r_i \in \mathbb{R}^{H \times W \times 3}$.

Our approach operates within a latent diffusion framework~\cite{rombach2022ldm}. LQ image $x_L$ is encoded to its latent representation $z_L = E_{\phi}(x_L)$, and similarly, HQ image $x_H$ is encoded to $z_{H} = E_{\phi}(x_H)$ for training. 

The forward diffusion process progressively introduces Gaussian noise to a clean latent vector $z_0$ (\eg, $z_{H}$ in our context) over a sequence of $T$ timesteps:
\begin{equation} \label{eq:forward_diffusion}
    z_t = \sqrt{\bar{\alpha}_t} \, z_0 + \sqrt{1 - \bar{\alpha}_t} \, \varepsilon,
\end{equation}
where $\varepsilon \sim \mathcal{N}(0, \mathbf{I})$ is the noise. The parameters $\alpha_t = 1 - \beta_t$ are determined by a predefined noise schedule $\beta_t \in (0, 1)$, and $\bar{\alpha}_t = \prod_{s=1}^{t} \alpha_s$ denotes their cumulative product.

The reverse diffusion process aims to denoise $z_t$ and recover an estimate of $z_0$. This is achieved by training network $\varepsilon_\theta(z_t, p, t)$ to approximate noise $\varepsilon$, conditioned on the noisy latent $z_t$, prompt embedding $p=p(x_L, R)$, and timestep $t$. The estimated clean latent $\hat{z}_0$ is then derived from $z_t$ and the predicted noise $\hat{\varepsilon} = \varepsilon_{\theta}(z_t, p, t)$ as:
\begin{equation} \label{eq:reverse_denoising}
    \hat{z}_0 = \frac{z_t - \sqrt{1 - \bar{\alpha}_t} \, \hat{\varepsilon}}{\sqrt{\bar{\alpha}_t}}.
\end{equation}
Once $\hat{z}_0$ (which corresponds to the estimated HQ latent $\hat{z}_H$) is obtained,  it is projected back into the pixel space by the VAE decoder to synthesize the final restored image: $\hat{x}_H = D_{\phi}(\hat{z}_H)$.%

During inference, the LQ input $x_L$ is encoded to its latent form $z_L = E_{\phi}(x_L)$. 
One-step inference~\cite{wu2024osediff,sun2024pisasr} is performed using a fixed timestep $T_L$(where $0$\,$\leq$\,$T_L$$<$\,$T$).
The target HQ latent vector $\hat{z}_H$ is subsequently computed in a single step:
\begin{equation} \label{eq:onestep_generate_zh}
    \hat{z}_H = \frac{z_L - \sqrt{1 - \bar{\alpha}_{T_L}} \varepsilon_{\theta} (z_L, p, T_L)}{\sqrt{\bar{\alpha}_{T_L}}}.
\end{equation}
This formulation enables an end-to-end, single-step restoration from the LQ latent $z_L$ to the HQ latent $\hat{z}_H$. 
Denoting the entire model as the generator $\mathcal{G}_{\psi}$, the overall procedure is
\begin{equation}
    \hat{x}_H = \mathcal{G}_{\psi}({x}_L, p). 
\end{equation}

\begin{figure*}[htbp]
    \centering
    \scalebox{0.9}{
    \vspace{-1mm}
    \begin{subfigure}{0.40\linewidth}
        \centering
        \includegraphics[width=\linewidth]{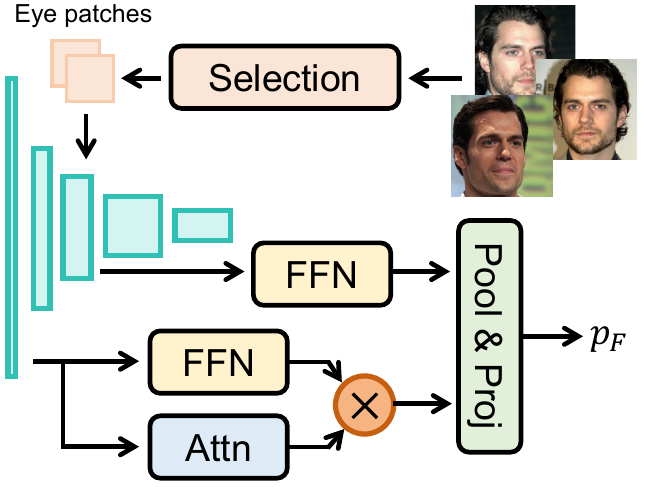}
        \phantomcaption
        \label{fig:ffe}
    \end{subfigure}
    \hspace{4mm}
    \begin{subfigure}{0.235\linewidth}
        \centering
        \includegraphics[width=\linewidth]{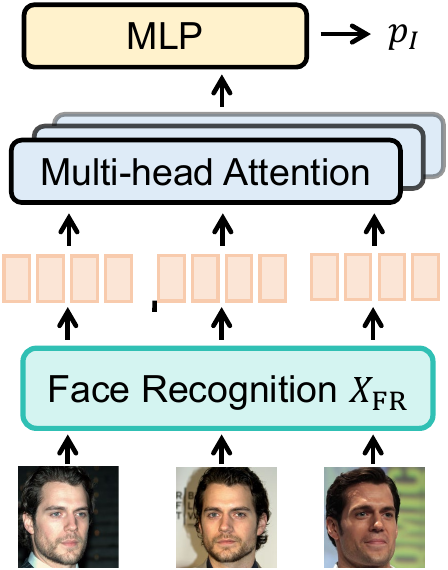}
        \phantomcaption
        \label{fig:fie}
    \end{subfigure}
    \hspace{4.5mm}
    \begin{subfigure}{0.30\linewidth}
        \centering
        \includegraphics[width=\linewidth]{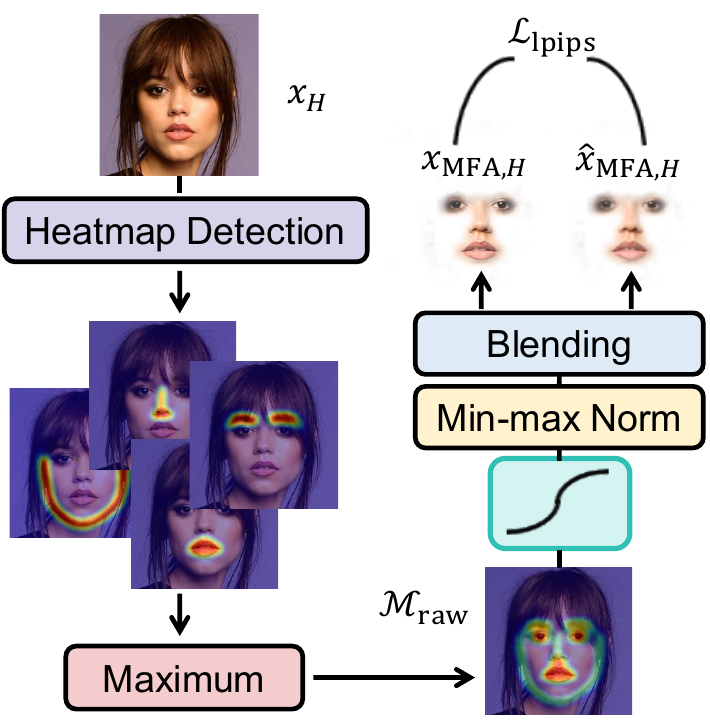}
        \phantomcaption
        \label{fig:mfa}
    \end{subfigure}
    }
    \vspace{-4mm}
    \caption{Key modules in HonestFace pipeline. (\textit{left}, a) Facial feature extractor: Generates a feature embedding $p_F$ from the most geometrically similar eye patches, selected from the optimal references; (\textit{middle}, b) Face identity encoder: Encodes multiple reference faces to produce a combined identity representation $p_I$; (\textit{right}, c) Masked face alignment: A novel perceptual loss that strategically focuses the restoration on perceptually crucial areas, ensuring the fidelity of key textures and natural colors.}
    \vspace{-3mm}
    \label{fig:modules}
\end{figure*}

To optimize the one-step diffusion model $\varepsilon_{\theta}$ training for high-fidelity generation, we employ an adversarial distillation strategy. The generative adversarial network (GAN) discriminator~\cite{Sauer2023ADD,Sauer2024LADD,li2024dfosd}, and the variational score distillation (VSD)~\cite{luo2023diffinstruct,wang2023prolificdreamer,wu2024osediff} are widely recognized for significantly enhancing the performance of one-step diffusion models. By introducing the discriminator presented in OSDFace~\cite{wang2025osdface}, we integrate the GAN discriminator to jointly optimize both the generator $\mathcal{G}_{\psi}$ and the discriminator $\mathcal{D}_{\psi}$. The corresponding loss functions are defined as:
\begin{equation}
\begin{aligned}
\mathcal{L}_\mathcal{G} =& -\mathbb{E}_t \left[\log \mathcal{D}_{\psi} \left( F(\hat{z}_H, t) \right)\right], \\
\mathcal{L}_\mathcal{D} =& -\mathbb{E}_t \left[\log \left( 1 - \mathcal{D}_{\psi} \left( F(\hat{z}_H, t) \right) \right)\right] \\&- \mathbb{E}_t \left[\log \mathcal{D}_{\psi} \left( F(E_\phi(x_H), t) \right)\right], \\
\end{aligned}
\end{equation}
where $F(\cdot, t)$ denotes the forward diffusion process applied to the input at timestep $t \in [0, T]$ during adversarial training.

\vspace{-1.5mm}
\subsection{Conditional Embeddings}\label{sec:conditional_input}
\vspace{-0.2mm}

The prompt embedding $p$ in Eq.~\eqref{eq:onestep_generate_zh} plays a key role in guiding the face restoration process. It provides targeted semantic information to support the generation of high-quality latent features $\hat{z}_H$. To avoid semantic ambiguities that arise from direct analysis of LQ input~\cite{wu2024seesr,wu2024osediff}, and to eliminate the need for additional textual prompts~\cite{chen2023promptsr,tao2025overcoming}, our framework integrates two specialized prompt embedding modules: our proposed identity embedder (IDE) and visual representation embedder (VRE)~\cite{wang2025osdface}. These modules collaboratively extract conditioning information directly from the input LQ image $x_L$ and reference images $R$. For convenience, we assume the final prompt embedding $p$ is a sequence of $n$ tokens, each of dimension $k$, \ie, $p \in \mathbb{R}^{n \times k}$, for efficient conditioning.

\vspace{1mm}
\noindent\textbf{Sorted Reference Image Set}.
Since we utilize multiple reference images, their guidance can sometimes be conflicting. Therefore, their presentation order becomes crucial. During training, we sort the reference images based on their similarity to the input image. Specifically, references with the most similar geometric pose are ranked first, as they are often more informative. During inference, a user can choose the same sorting scheme or adopt a custom order.

To implement this similarity-based sorting, we must select a robust metric. Standard pixel-wise L2 distance is not good enough, as it is highly sensitive to minor global misalignments (\eg, small translations, rotations, or scaling). We therefore choose the affine landmark distance $d_\text{A-LD}$~\cite{li2020asffnet512}. This metric computes a more perceptually relevant distance between two faces after applying an optimal affine transformation to align their landmarks. 

Given landmarks $L=\{l_k\}_{k=1}^N$, target HQ landmarks $H=\{h_k\}_{k=1}^N$, and corresponding weights $w_k$, we calculate the minimal weighted squared distance $d_\text{A-LD}$ after an optimal 2D affine alignment:
\begin{equation}
d_\text{A-LD}(L, H; W) = \min_{A}\sum_{k=1}^{N} w_k|A l_k' - h_k|^2,
\end{equation}
where $A \in \mathbb{R}^{2\times 3}$ is the affine transformation matrix and $l_k'$ denotes $l_k$ in homogeneous coordinates. 
Further analysis of the weighted least-squares solution $A$ and a complete definition of $d_\text{A-LD}$ are provided in the supplementary material.

In our context of reference selection, the $d_\text{A-LD}$ is particularly effective precisely because it is alignment-invariant. A high distance score implies that no simple affine transformation can align the reference to the input. This reliably identifies and discards poor references, such as those with severe occlusions or extreme pose variations, which would otherwise provide erroneous guidance.

\begin{figure*}[t]
\begin{center}
\includegraphics[width=\textwidth]{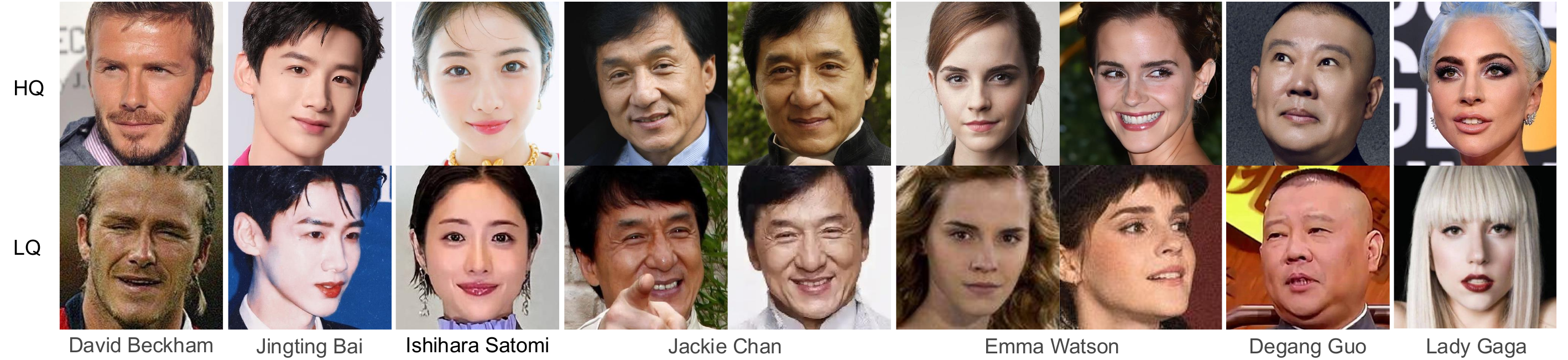}
\end{center}
\vspace{-3mm}
\caption{Overview of the MultiRefCeleb-Test benchmark. Each identity features multiple high-quality reference images alongside several low-quality inputs. The dataset spans diverse ethnicities and age ranges. It incorporates a wide spectrum of authentic real-world degradations and challenging cases, such as occlusions, varied illumination, and diverse facial expressions.}
\label{fig:dataset}
\vspace{-3mm}
\end{figure*}

\vspace{1mm}
\noindent\textbf{Identity Embedder (IDE)}.
The identity embedder (IDE) is principally focused on enhancing identity fidelity. As shown in {Fig.~\ref{fig:overall}}, the IDE comprises two parallel components: a facial feature extractor and a face identity encoder. 

\vspace{1mm}
\noindent\textit{Facial Feature Extractor}. 
Key facial characteristics, such as the eyes, nose, and mouth, are important for human identity perception. Variations in eye color, skin texture, and edge structure strongly influence an individual’s appearance. Given that the LQ input often suffers from considerable degradation, extracting such detailed identity solely from $x_L$ is unreliable. Consequently, leveraging information from high-quality reference images becomes essential.

We observe that certain facial attributes, particularly ocular features, remain highly consistent for the same individual across images. This allows us to select an optimal reference image that closely matches the input. Let $R = \{r_1, r_2, \dots, r_N\}$ denote the set of available references. The optimal reference $r_0$ could be identified by
\begin{equation}
    r_0 = \arg \min_{r_i \in R} d_\text{A-LD}(r_i, x_L),
\end{equation}
where $d_\text{A-LD}(\cdot, \cdot)$ denotes the affine landmark distance.

From the selected reference $r_0$, we crop patches corresponding to the left eye $e_l$ and the right eye $e_r$. These patches are processed using a pre-trained VGG16 network~\cite{simonyan2015vgg} to extract distinct color and structural features for identity-aware representation.

Drawing inspiration from studies on VGG representations~\cite{Gatys2015Texture,Mahendran2015Understanding}, we use features from multiple depths. Early layers capture low-level details like color and fine textures, while mid-level layers capture more complex patterns and structures. As shown in Fig.~\ref{fig:ffe}, for an eye patch $e_p$ (where $p \in \{l, r\}$), its layer 0 feature $f_{\text{vgg},0}(e_p)$ yields a color embedding $f_{\text{color},p}$, and its layer 2 feature $f_{\text{vgg},2}(e_p)$ yields a structural embedding $f_{\text{struct},p}$. These four embeddings are concatenated and projected to form the output $p_F \in \mathbb{R}^{n_1 \times k}$.

\vspace{1mm}
\noindent\textit{Face Identity Encoder}. 
Extracting robust global identity information is a key challenge in face recognition. 
Many face recognition models aim to keep embeddings from the same identity clustered close, while pushing apart those from different identities.
We process the aligned reference set $R$\,$=$\,$\{r_1, r_2, \dots, r_N\}$ using such a pre-trained face recognition model $X_{\text{FR}}$. This yields a set of identity embeddings $E_{\text{id}}$\,$=$\,$\{e_{\text{id},1}, e_{\text{id},2}, \dots, e_{\text{id},N}\}$, where each $e_{\text{id},i}$\,$=$\,$X_{\text{FR}}(r_i) \in \mathbb{R}^{d_{\text{id}}}$.
$E_{\text{id}}$ is then aggregated by an attention-based module, shown in Fig.~\ref{fig:fie}, to selectively extract the most important identity cues. It effectively pools identity information without distorting the learned feature space, thus producing the final output $p_I \in \mathbb{R}^{n_2 \times k}$.

\vspace{1mm}
\noindent\textbf{Visual Representation Embedder (VRE)}.
Extracting pertinent information directly from the LQ input $x_L$, despite its degradation, is also vital. Visual representation embedder~\cite{wang2025osdface} typically consists of a pre-trained VQ-VAE architecture, featuring an encoder $E_{\text{VRE}}$ specialized for LQ inputs. This encoder tokenizes $x_L$ and queries a learned codebook $\text{Dict}_{\text{LQ}}$ via nearest-neighbor lookup to produce the visual prompt $p_L$, capturing distilled representations from $x_L$:
\begin{equation}
    p_L = \text{Dict}_{\text{LQ}}(E_{\text{VRE}}(x_L)) \in \mathbb{R}^{n_3 \times k}.
\end{equation}

Finally, the parallel IDE and VRE produce feature sequences $p_F \in \mathbb{R}^{n_1 \times k}$, $p_I \in \mathbb{R}^{n_2 \times k}$, and $p_L \in \mathbb{R}^{n_3 \times k}$. 
These embeddings are fused via a trainable attention layer to yield the final prompt embedding $p \in \mathbb{R}^{n \times k}$, which is then fed into the diffusion model.

\begin{figure*}[t]
\footnotesize
\centering
\newcommand{\imgid}{00009}
\newcommand{\imgnote}{009}
\scalebox{0.98}{
    \hspace{-0.4cm}
    \begin{adjustbox}{valign=t}
    \begin{tabular}{ccccccccccc}
    \includegraphics[width=0.098\textwidth]{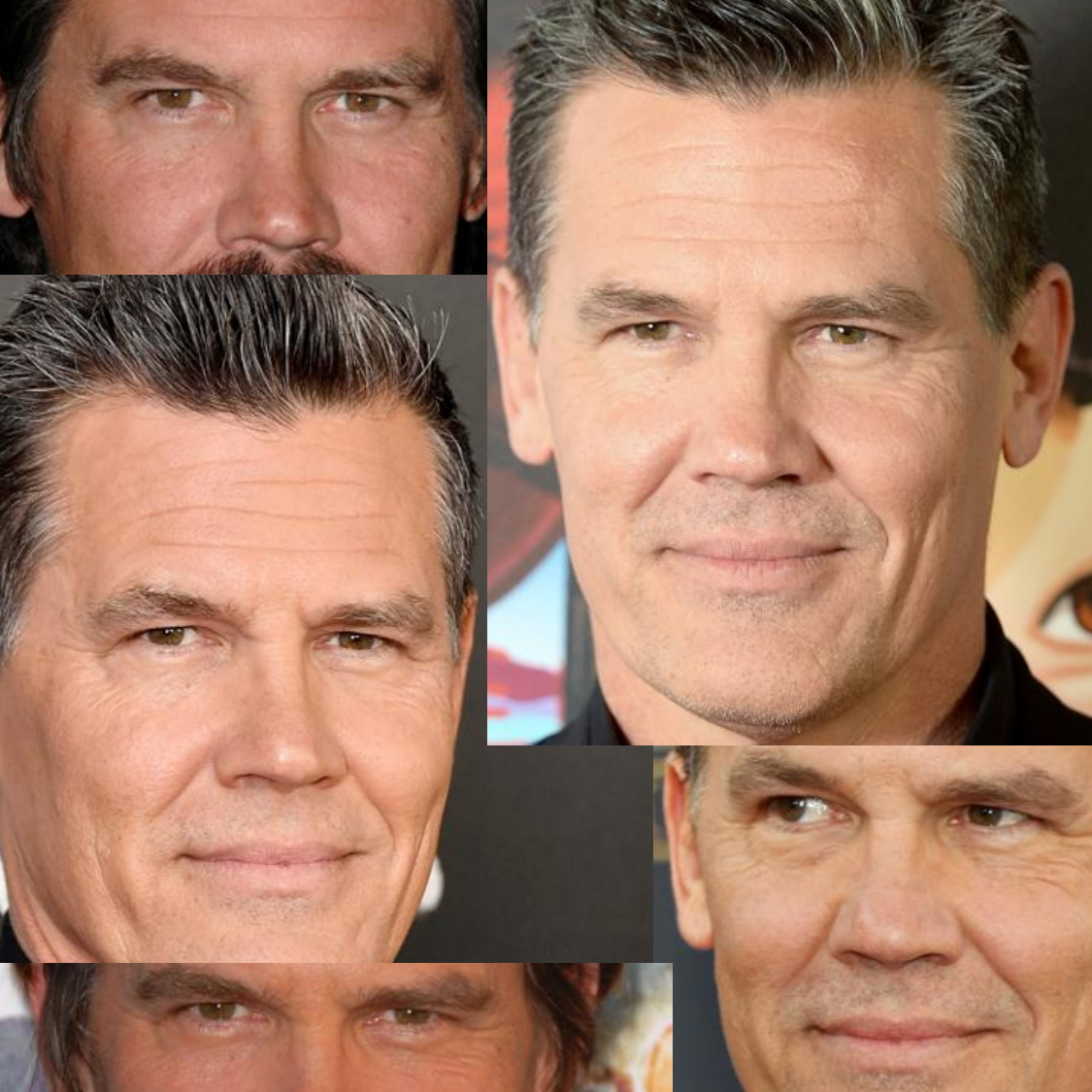} \hspace{-4mm} &
    \includegraphics[width=0.098\textwidth]{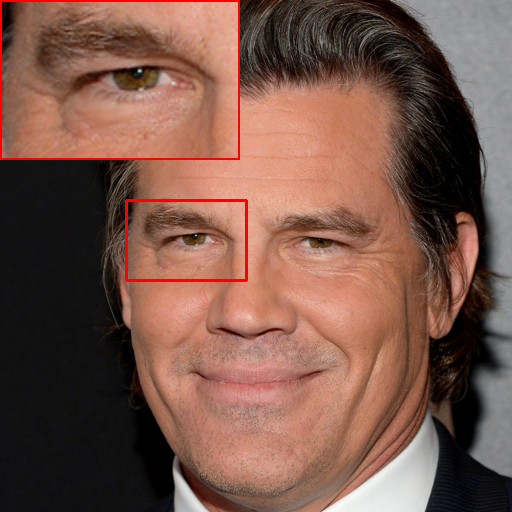} \hspace{-4mm} &
    \includegraphics[width=0.098\textwidth]{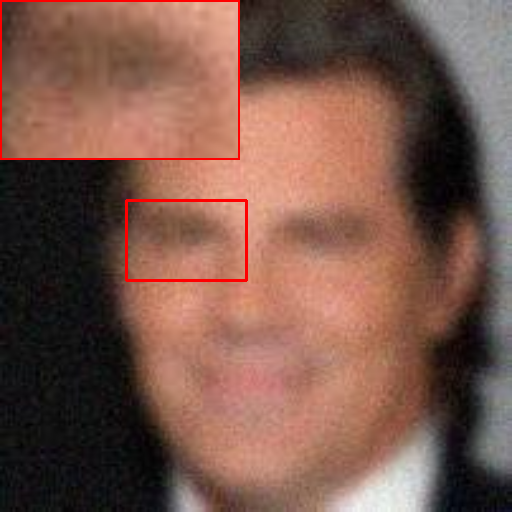} \hspace{-4mm} &
    \includegraphics[width=0.098\textwidth]{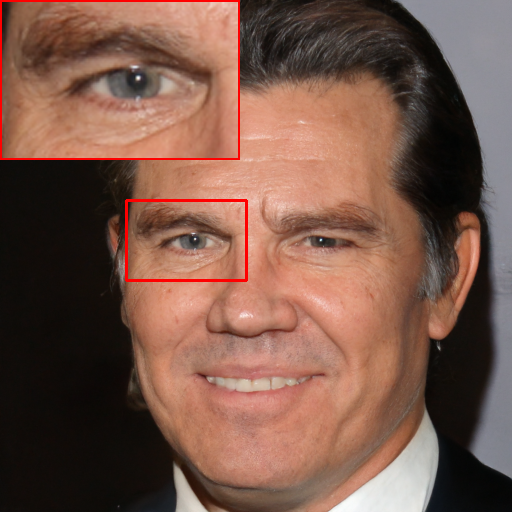} \hspace{-4mm} &
    \includegraphics[width=0.098\textwidth]{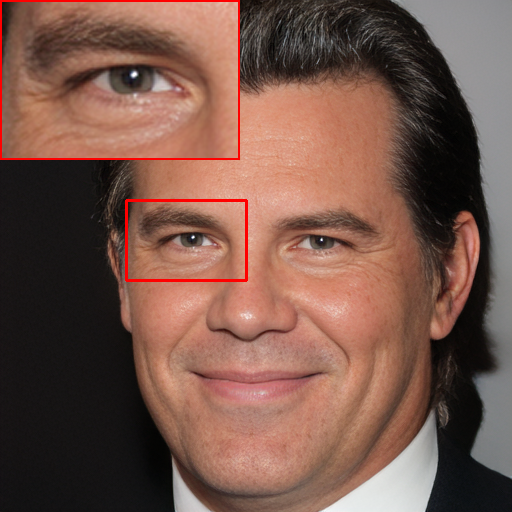} \hspace{-4mm} &
    \includegraphics[width=0.098\textwidth]{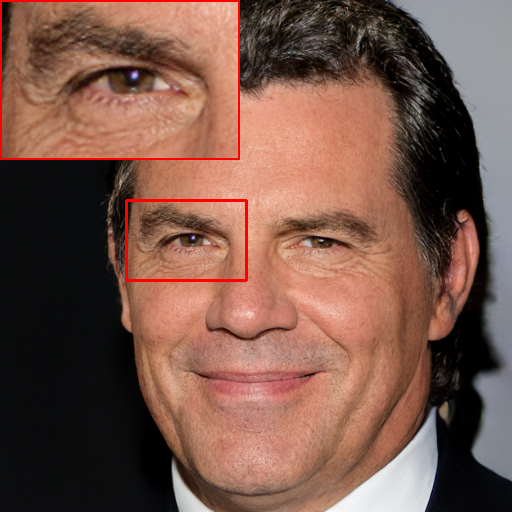} \hspace{-4mm} &
    \includegraphics[width=0.098\textwidth]{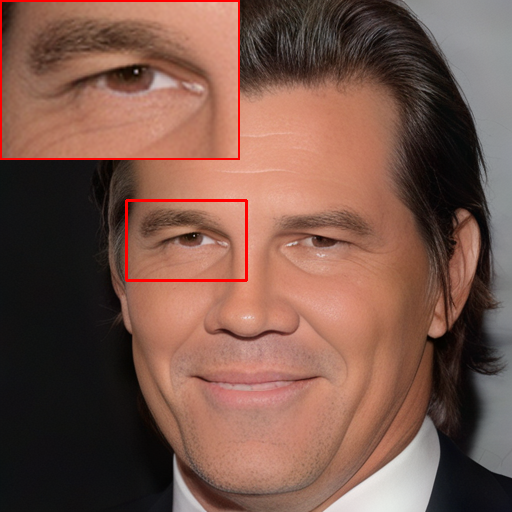} \hspace{-4mm} &
    \includegraphics[width=0.098\textwidth]{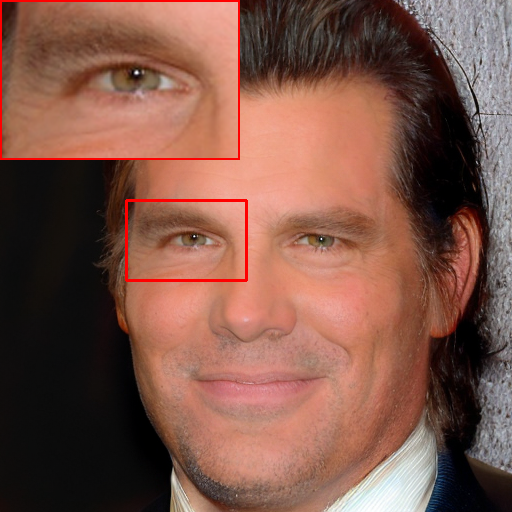} \hspace{-4mm} &
    \includegraphics[width=0.098\textwidth]{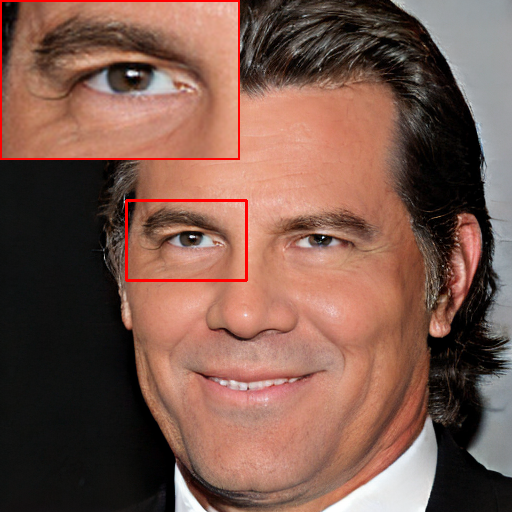} \hspace{-4mm} &
    \includegraphics[width=0.098\textwidth]{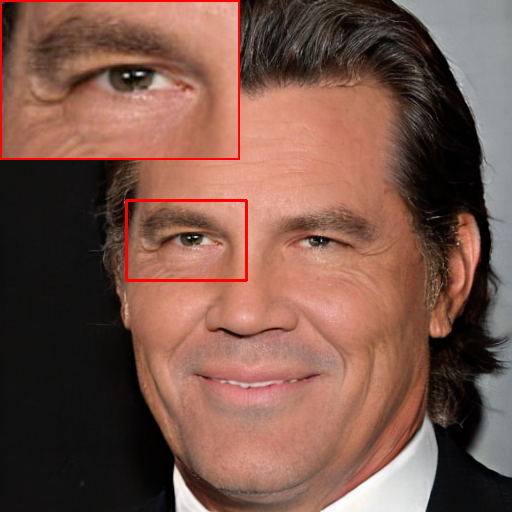} \hspace{-4mm} 
    \\
    Ref - \imgnote \hspace{-4mm} &
    HQ - \imgnote \hspace{-4mm} &
    LQ - \imgnote \hspace{-4mm} &
    PGDiff~\cite{yang2023pgdiff} \hspace{-4mm} &
    CodeFormer~\cite{zhou2022codeformer} \hspace{-4mm} & 
    DAEFR~\cite{tsai2024daefr} \hspace{-4mm} &
    FaceMe~\cite{liu2025faceme} \hspace{-4mm} & 
    MGFR~\cite{tao2025overcoming} \hspace{-4mm} &
    OSDFace~\cite{wang2025osdface} \hspace{-4mm} &
    HonestFace (ours) \hspace{-4mm}
    \\
    \end{tabular}
    \end{adjustbox}
}
\renewcommand{\imgid}{00094}
\renewcommand{\imgnote}{094}
\scalebox{0.98}{
    \hspace{-0.4cm}
    \begin{adjustbox}{valign=t}
    \begin{tabular}{ccccccccccc}
    \includegraphics[width=0.098\textwidth]{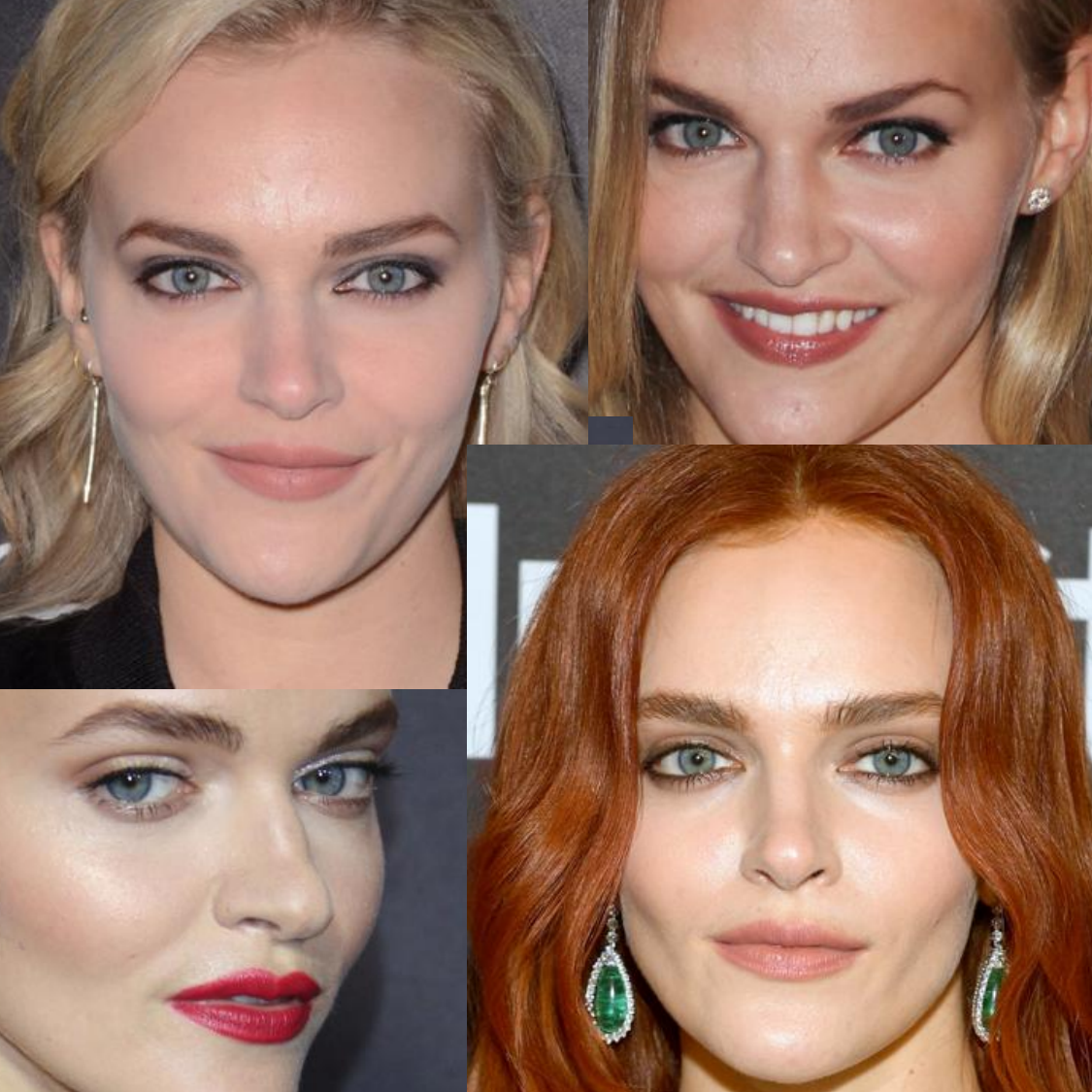} \hspace{-4mm} &
    \includegraphics[width=0.098\textwidth]{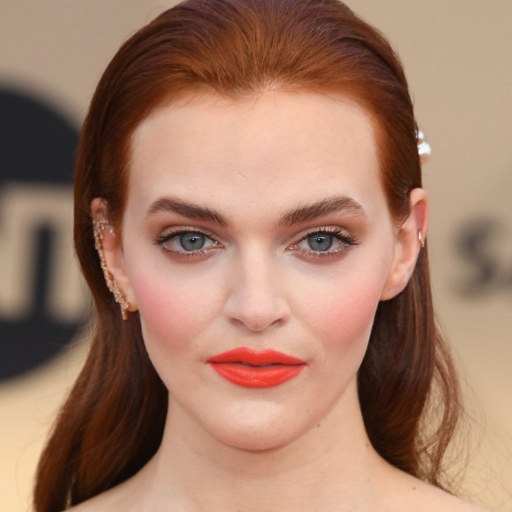} \hspace{-4mm} &
    \includegraphics[width=0.098\textwidth]{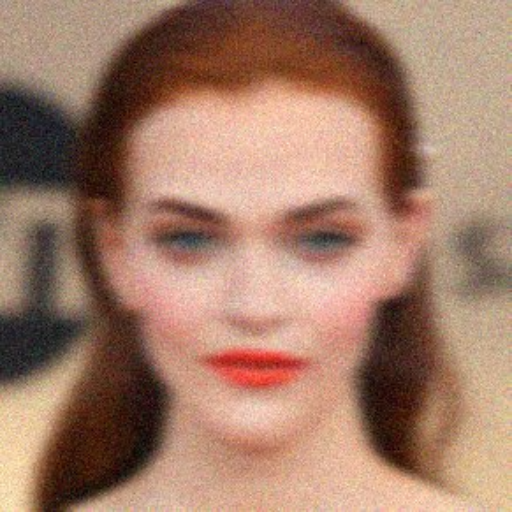} \hspace{-4mm} &
    \includegraphics[width=0.098\textwidth]{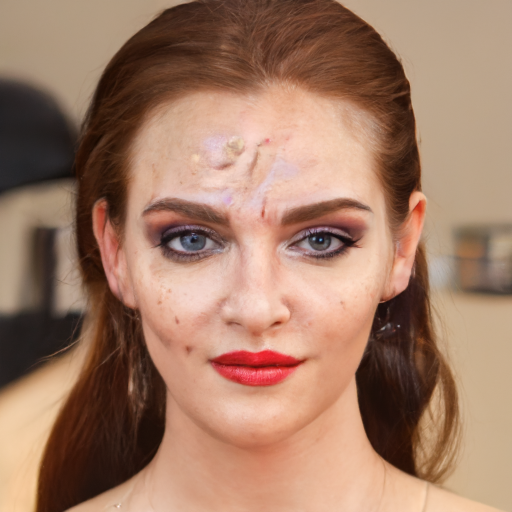} \hspace{-4mm} &
    \includegraphics[width=0.098\textwidth]{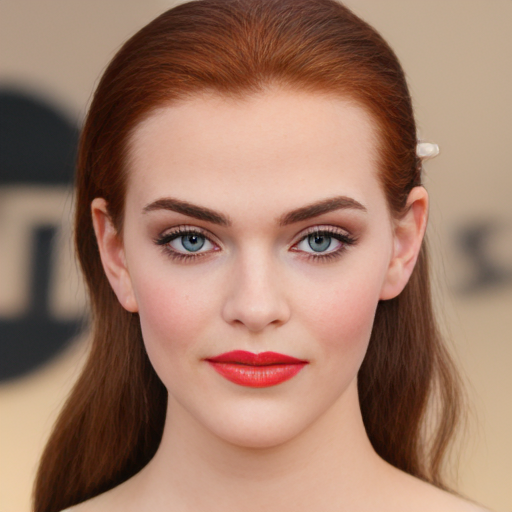} \hspace{-4mm} &
    \includegraphics[width=0.098\textwidth]{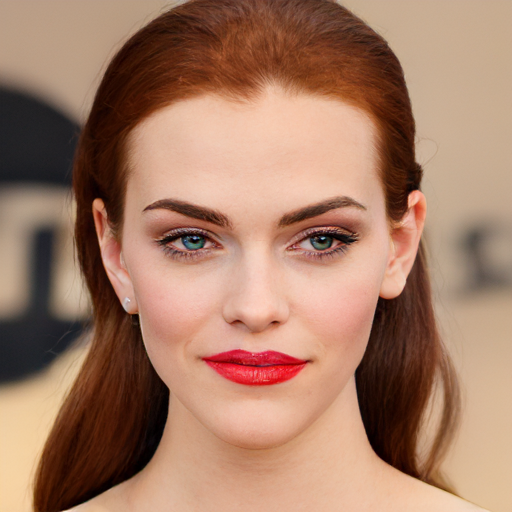} \hspace{-4mm} &
    \includegraphics[width=0.098\textwidth]{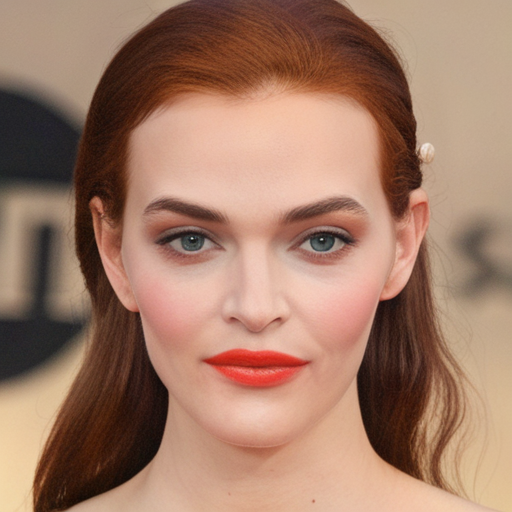} \hspace{-4mm} &
    \includegraphics[width=0.098\textwidth]{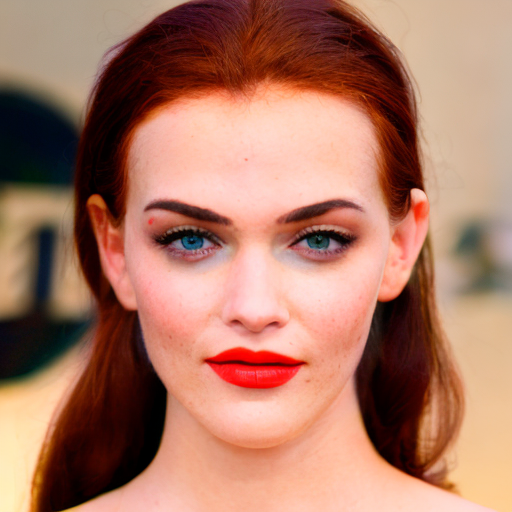} \hspace{-4mm} &
    \includegraphics[width=0.098\textwidth]{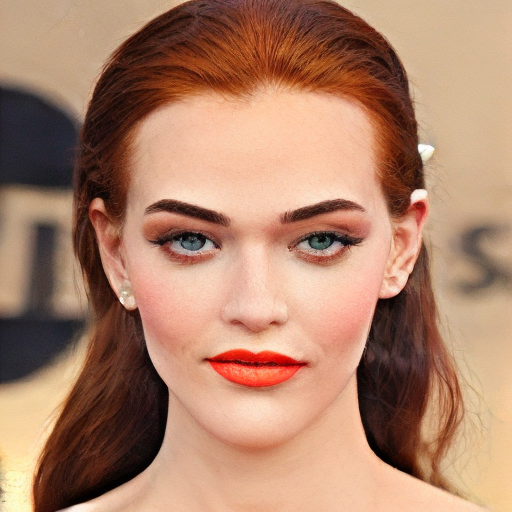} \hspace{-4mm} &
    \includegraphics[width=0.098\textwidth]{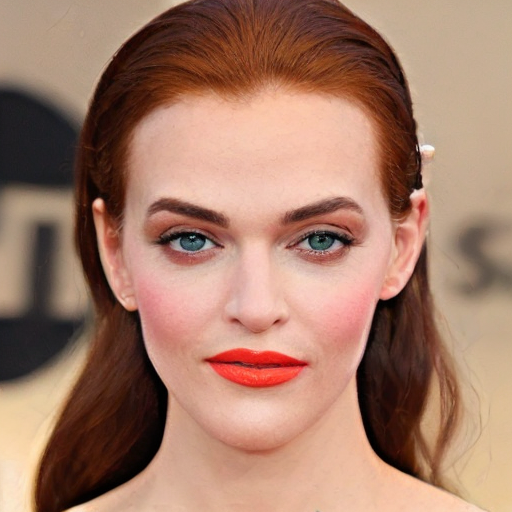} \hspace{-4mm} 
    \\
    Ref - \imgnote \hspace{-4mm} &
    HQ - \imgnote \hspace{-4mm} &
    LQ - \imgnote \hspace{-4mm} &
    PGDiff~\cite{yang2023pgdiff} \hspace{-4mm} &
    CodeFormer~\cite{zhou2022codeformer} \hspace{-4mm} & 
    DAEFR~\cite{tsai2024daefr} \hspace{-4mm} &
    FaceMe~\cite{liu2025faceme} \hspace{-4mm} & 
    MGFR~\cite{tao2025overcoming} \hspace{-4mm} &
    OSDFace~\cite{wang2025osdface} \hspace{-4mm} &
    HonestFace (ours) \hspace{-4mm}
    \\
    \end{tabular}
    \end{adjustbox}
}
\renewcommand{\imgid}{00059}
\renewcommand{\imgnote}{059}
\scalebox{0.98}{
    \hspace{-0.4cm}
    \begin{adjustbox}{valign=t}
    \begin{tabular}{ccccccccccc}
    \includegraphics[width=0.098\textwidth]{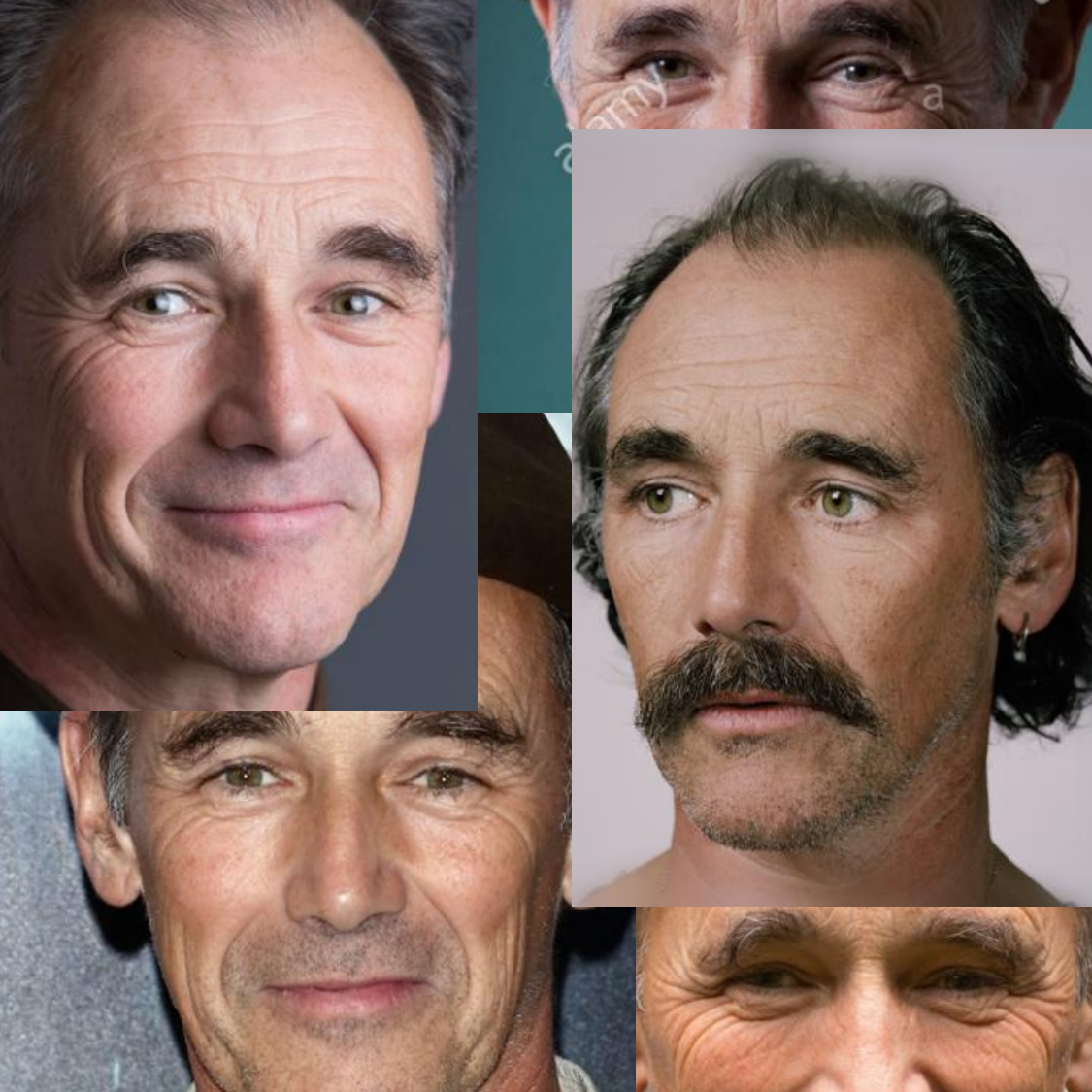} \hspace{-4mm} &
    \includegraphics[width=0.098\textwidth]{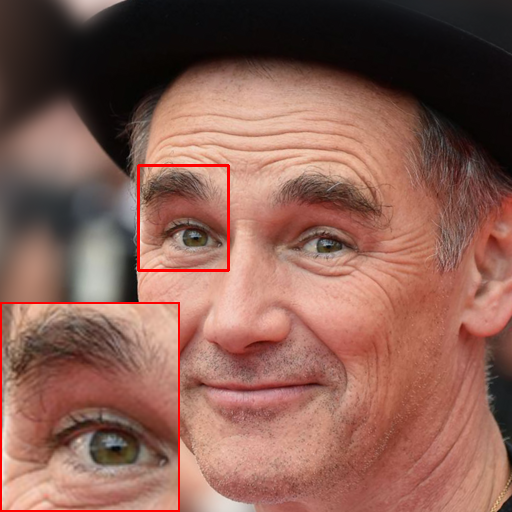} \hspace{-4mm} &
    \includegraphics[width=0.098\textwidth]{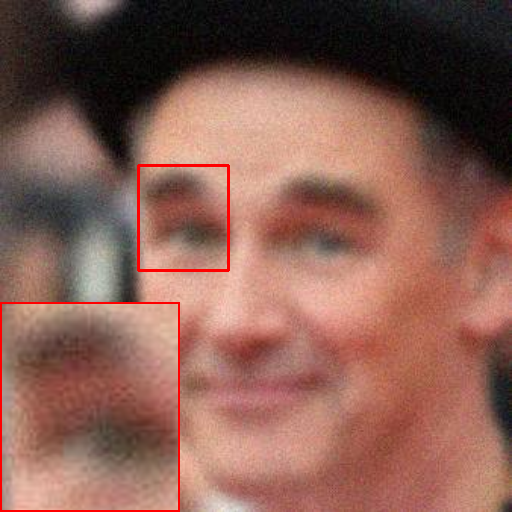} \hspace{-4mm} &
    \includegraphics[width=0.098\textwidth]{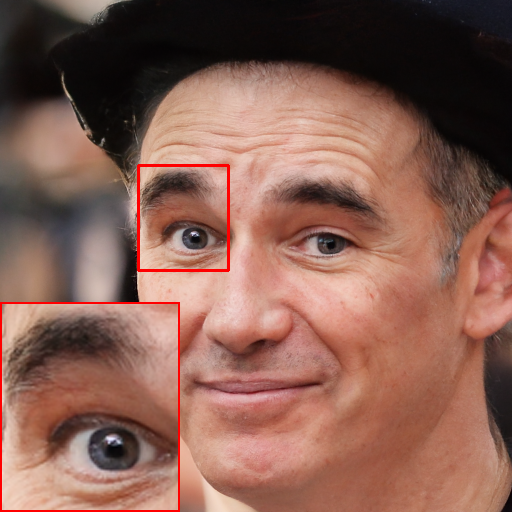} \hspace{-4mm} &
    \includegraphics[width=0.098\textwidth]{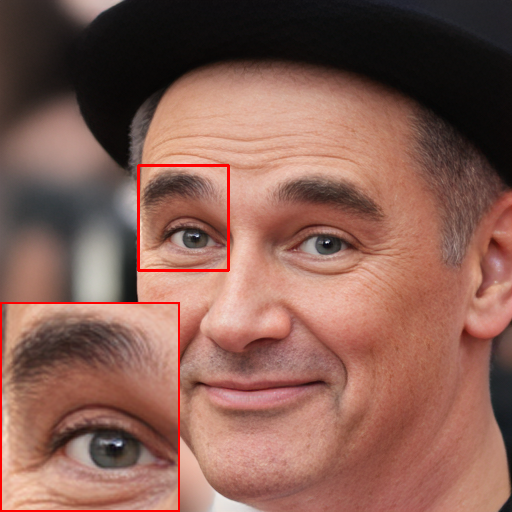} \hspace{-4mm} &
    \includegraphics[width=0.098\textwidth]{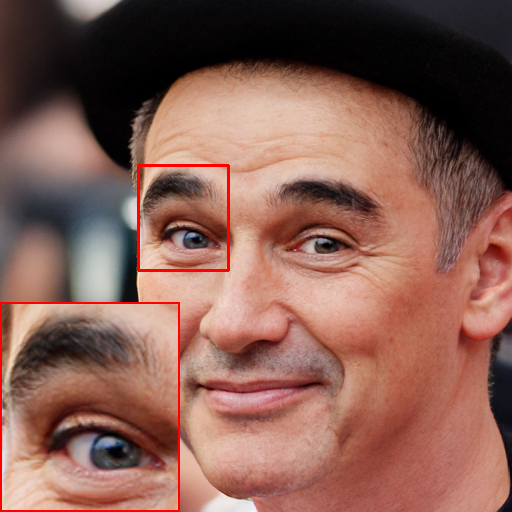} \hspace{-4mm} &
    \includegraphics[width=0.098\textwidth]{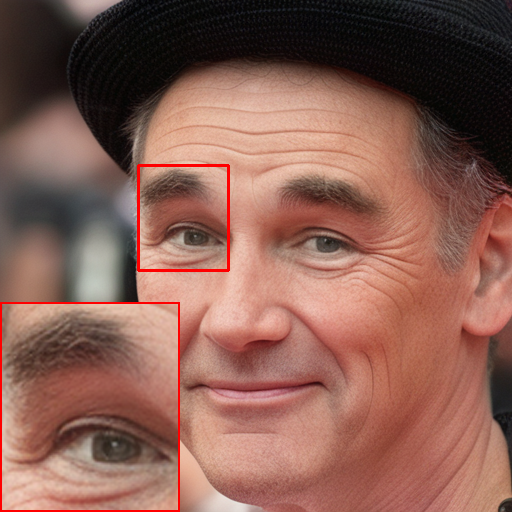} \hspace{-4mm} &
    \includegraphics[width=0.098\textwidth]{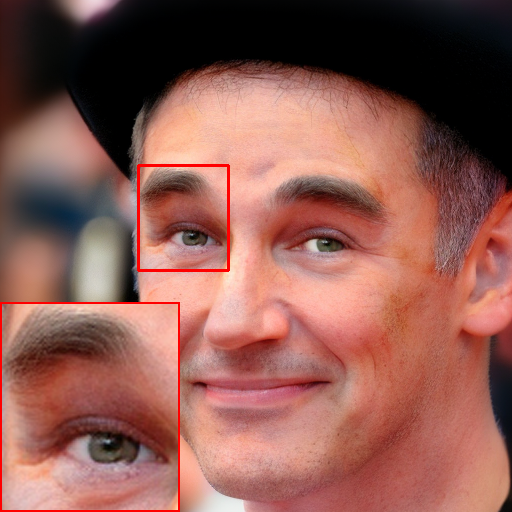} \hspace{-4mm} &
    \includegraphics[width=0.098\textwidth]{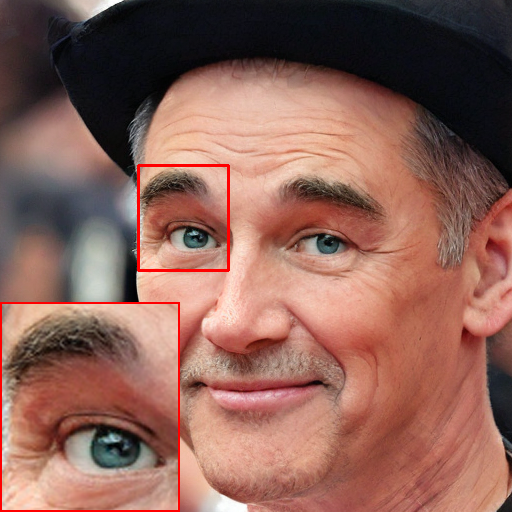} \hspace{-4mm} &
    \includegraphics[width=0.098\textwidth]{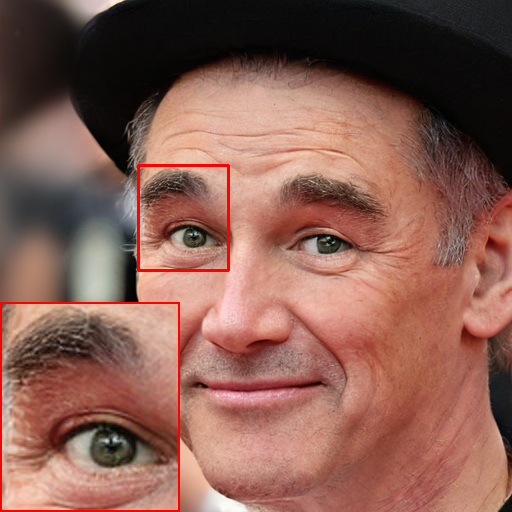} \hspace{-4mm} 
    \\
    Ref - \imgnote \hspace{-4mm} &
    HQ - \imgnote \hspace{-4mm} &
    LQ - \imgnote \hspace{-4mm} &
    PGDiff~\cite{yang2023pgdiff} \hspace{-4mm} &
    CodeFormer~\cite{zhou2022codeformer} \hspace{-4mm} & 
    DAEFR~\cite{tsai2024daefr} \hspace{-4mm} &
    FaceMe~\cite{liu2025faceme} \hspace{-4mm} & 
    MGFR~\cite{tao2025overcoming} \hspace{-4mm} &
    OSDFace~\cite{wang2025osdface} \hspace{-4mm} &
    HonestFace (ours) \hspace{-4mm}
    \\
    \end{tabular}
    \end{adjustbox}
}
\renewcommand{\imgid}{112}
\renewcommand{\imgnote}{112}
\scalebox{0.98}{
    \hspace{-0.4cm}
    \begin{adjustbox}{valign=t}
    \begin{tabular}{ccccccccccc}
    \includegraphics[width=0.098\textwidth]{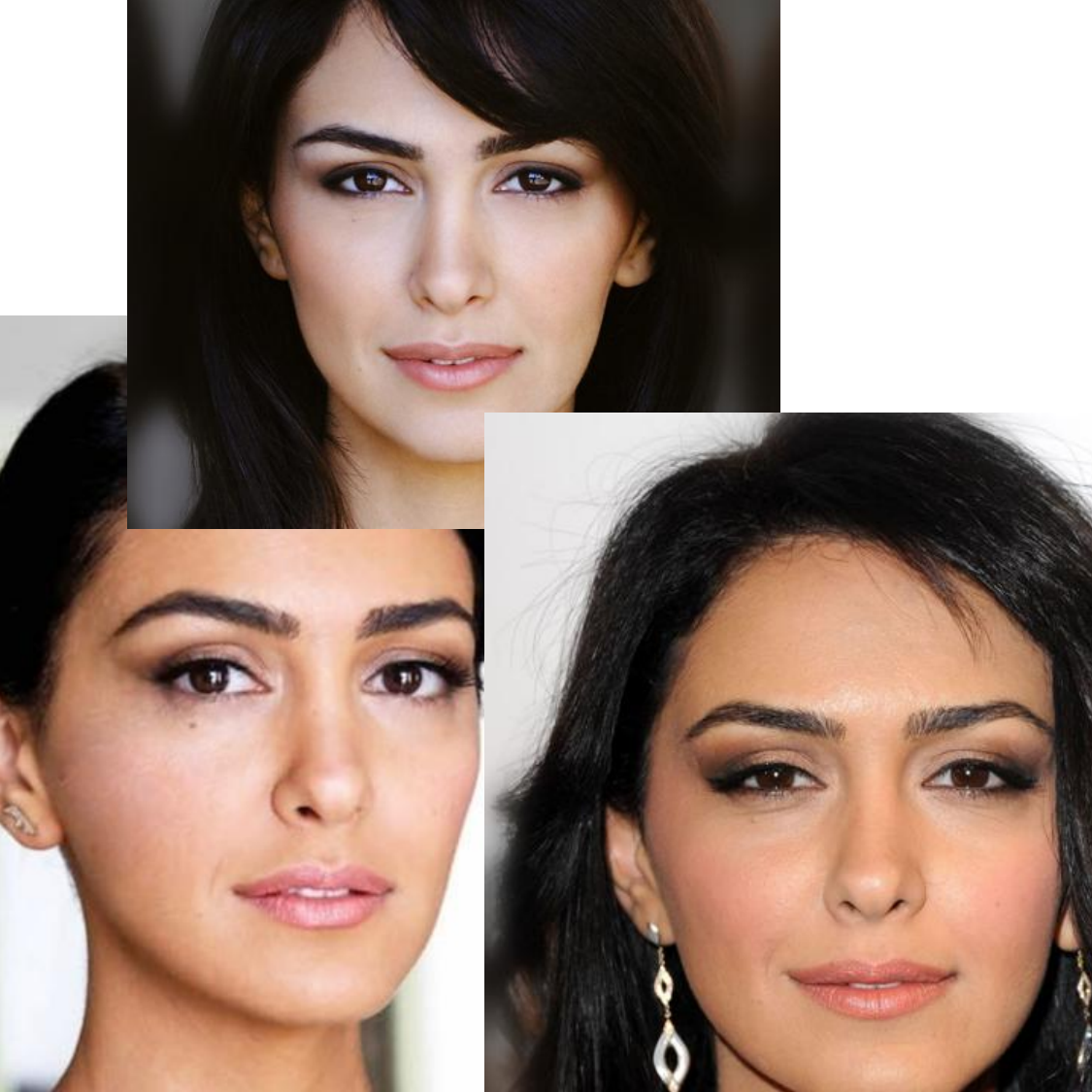} \hspace{-4mm} &
    \includegraphics[width=0.098\textwidth]{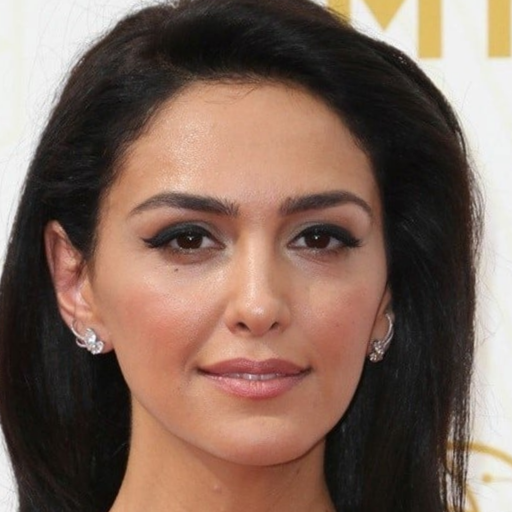} \hspace{-4mm} &
    \includegraphics[width=0.098\textwidth]{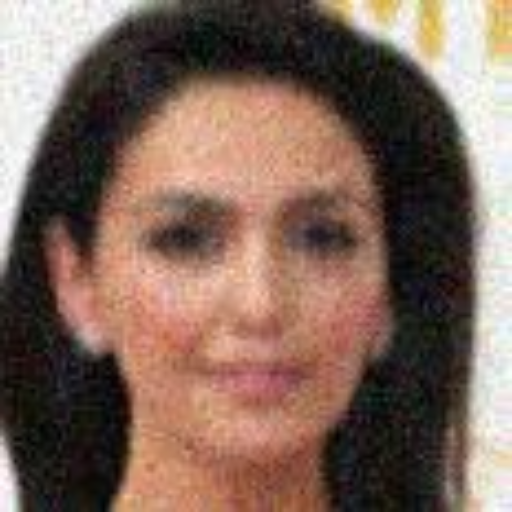} \hspace{-4mm} &
    \includegraphics[width=0.098\textwidth]{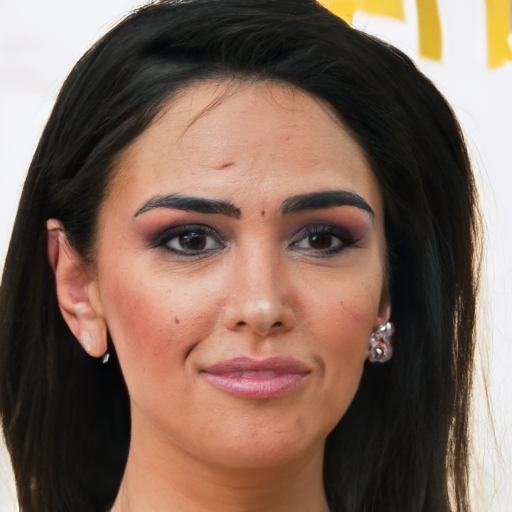} \hspace{-4mm} &
    \includegraphics[width=0.098\textwidth]{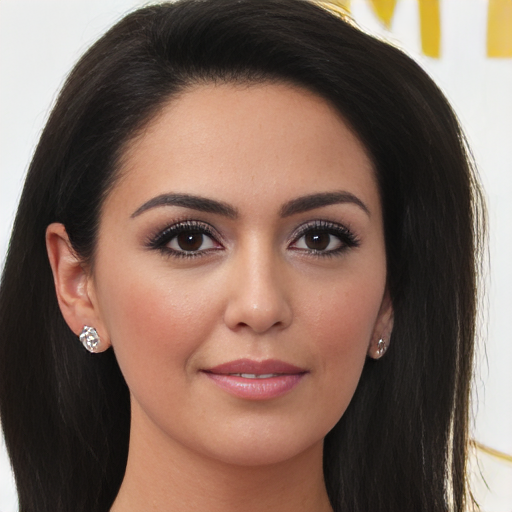} \hspace{-4mm} &
    \includegraphics[width=0.098\textwidth]{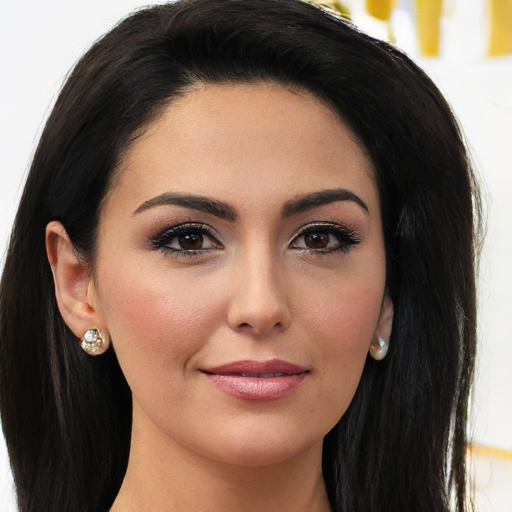} \hspace{-4mm} &
    \includegraphics[width=0.098\textwidth]{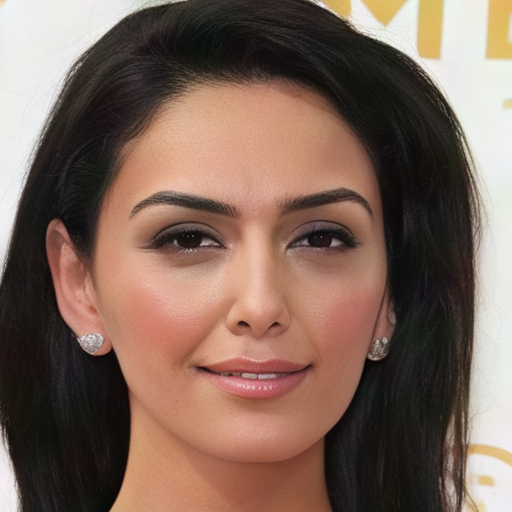} \hspace{-4mm} &
    \includegraphics[width=0.098\textwidth]{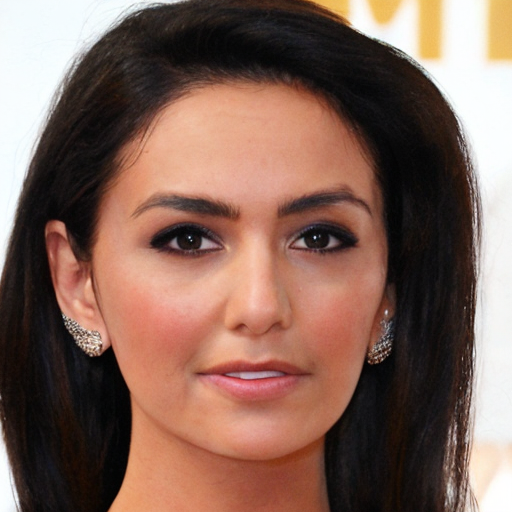} \hspace{-4mm} &
    \includegraphics[width=0.098\textwidth]{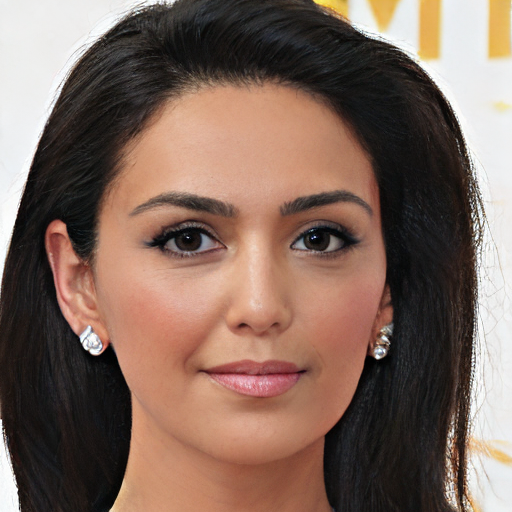} \hspace{-4mm} &
    \includegraphics[width=0.098\textwidth]{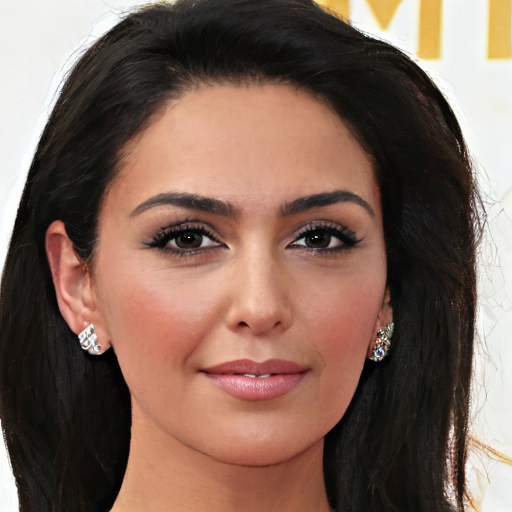} \hspace{-4mm} 
    \\
    Ref - \imgnote \hspace{-4mm} &
    HQ - \imgnote \hspace{-4mm} &
    LQ - \imgnote \hspace{-4mm} &
    PGDiff~\cite{yang2023pgdiff} \hspace{-4mm} &
    CodeFormer~\cite{zhou2022codeformer} \hspace{-4mm} & 
    DAEFR~\cite{tsai2024daefr} \hspace{-4mm} &
    FaceMe~\cite{liu2025faceme} \hspace{-4mm} & 
    MGFR~\cite{tao2025overcoming} \hspace{-4mm} &
    OSDFace~\cite{wang2025osdface} \hspace{-4mm} &
    HonestFace (ours) \hspace{-4mm}
    \\
    \end{tabular}
    \end{adjustbox}
}
\renewcommand{\imgid}{320}
\renewcommand{\imgnote}{320}
\scalebox{0.98}{
    \hspace{-0.4cm}
    \begin{adjustbox}{valign=t}
    \begin{tabular}{ccccccccccc}
    \includegraphics[width=0.098\textwidth]{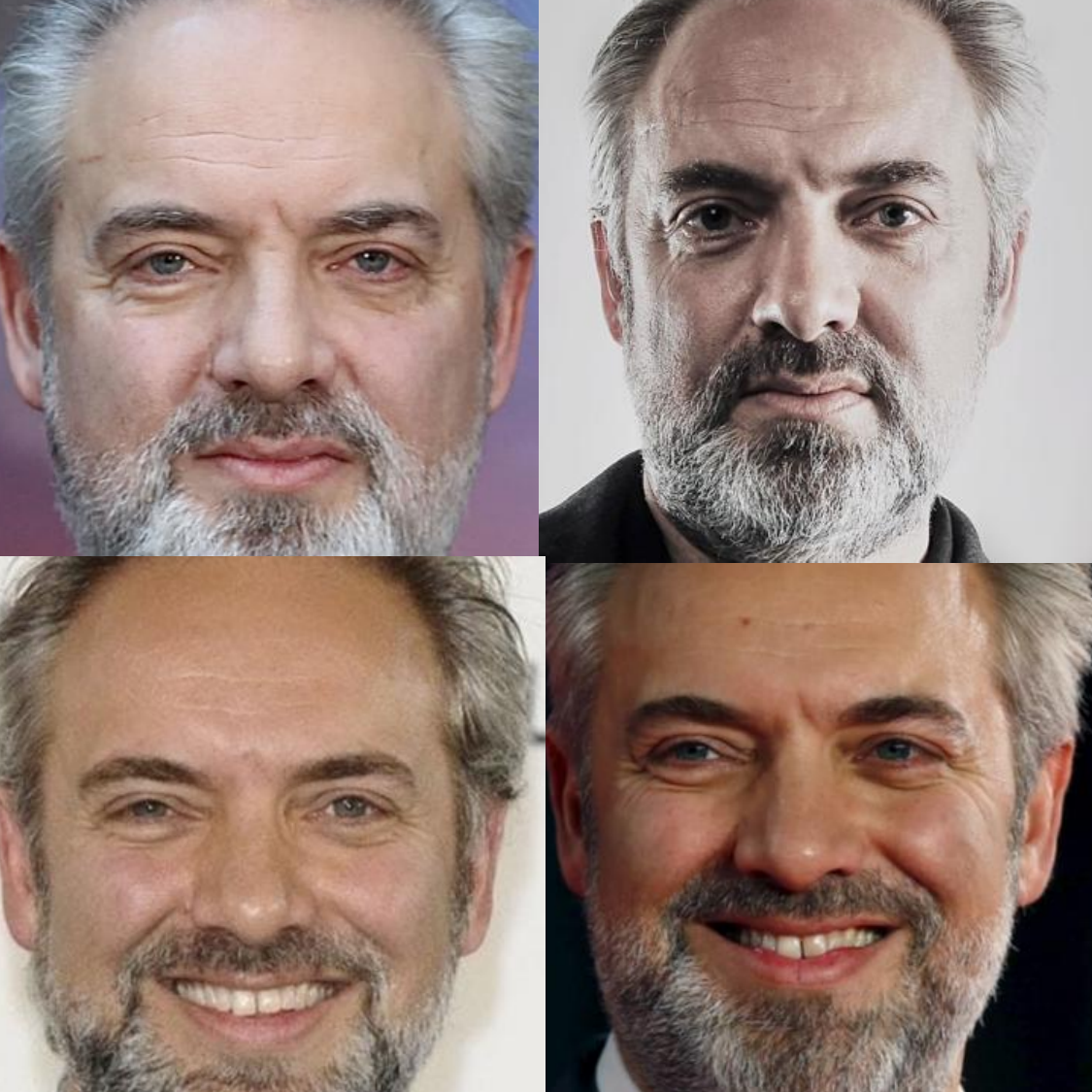} \hspace{-4mm} &
    \includegraphics[width=0.098\textwidth]{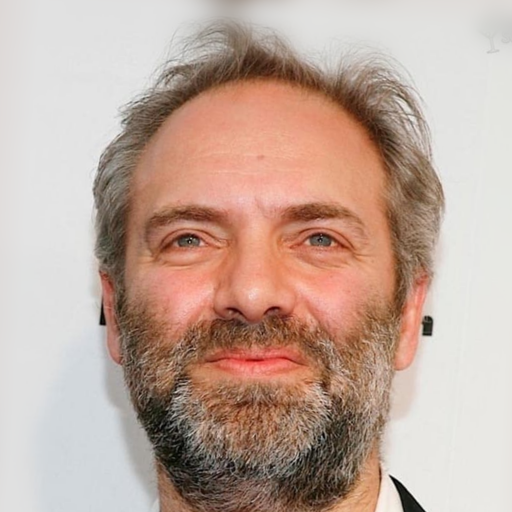} \hspace{-4mm} &
    \includegraphics[width=0.098\textwidth]{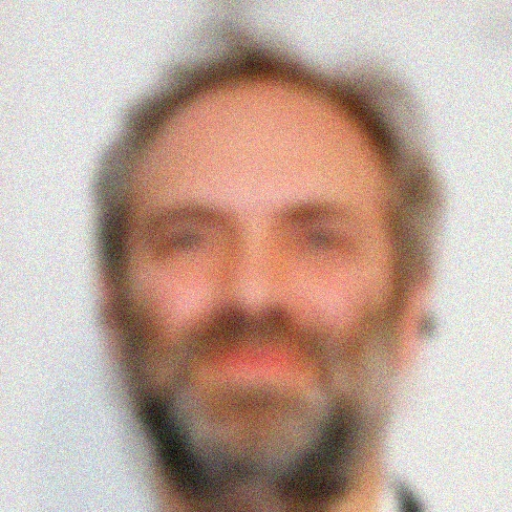} \hspace{-4mm} &
    \includegraphics[width=0.098\textwidth]{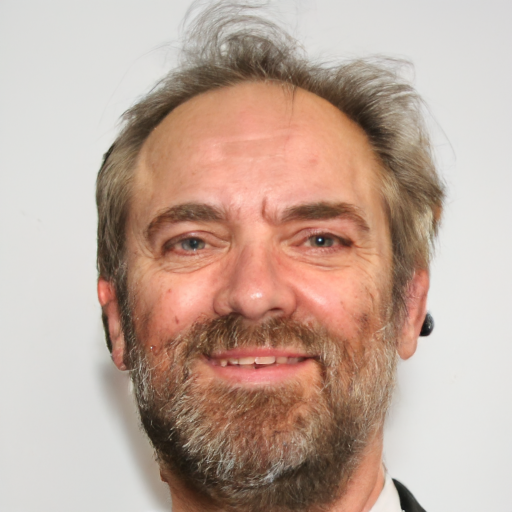} \hspace{-4mm} &
    \includegraphics[width=0.098\textwidth]{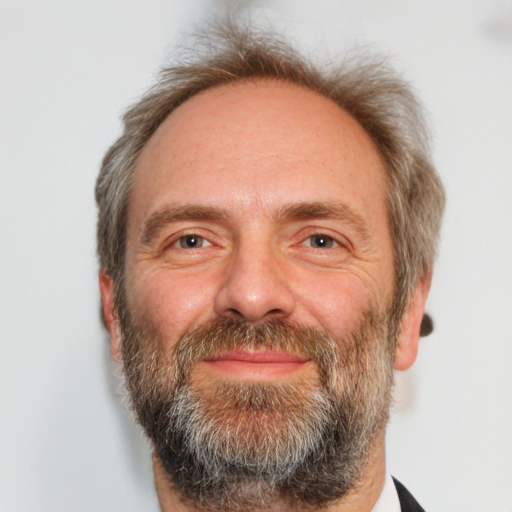} \hspace{-4mm} &
    \includegraphics[width=0.098\textwidth]{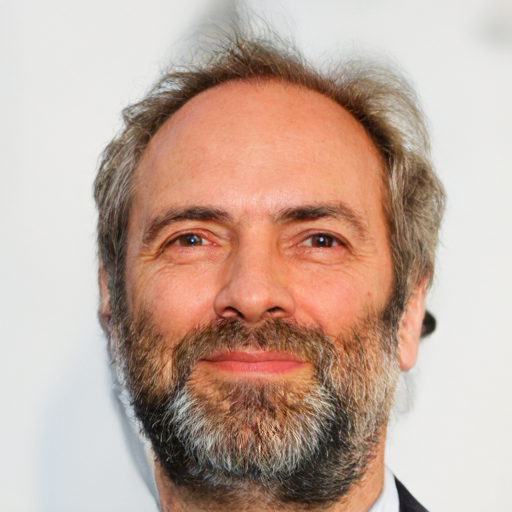} \hspace{-4mm} &
    \includegraphics[width=0.098\textwidth]{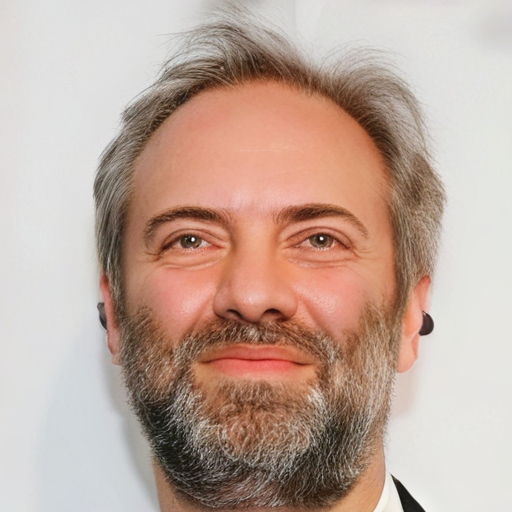} \hspace{-4mm} &
    \includegraphics[width=0.098\textwidth]{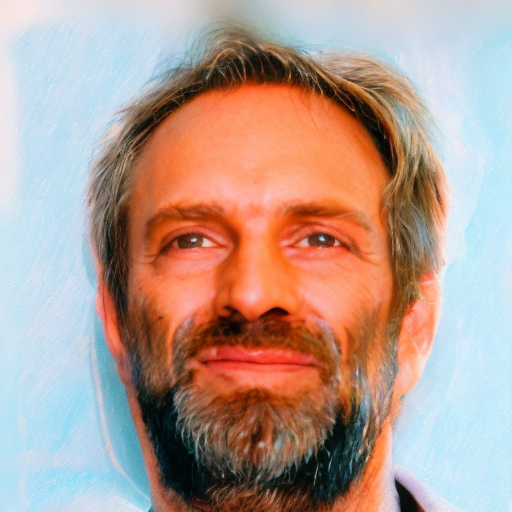} \hspace{-4mm} &
    \includegraphics[width=0.098\textwidth]{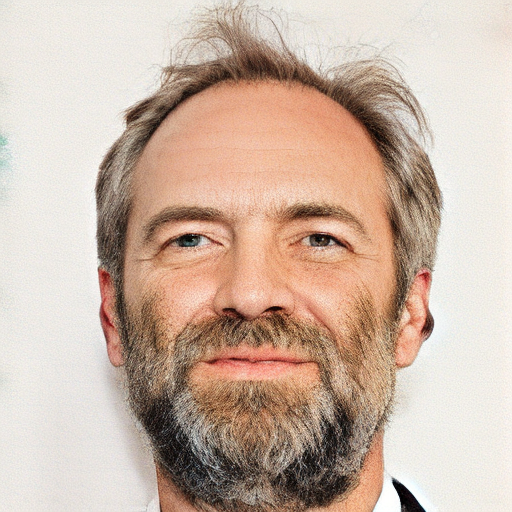} \hspace{-4mm} &
    \includegraphics[width=0.098\textwidth]{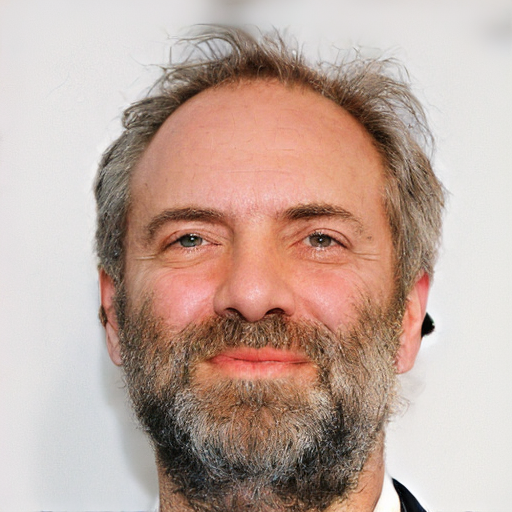} \hspace{-4mm} 
    \\
    Ref - \imgnote \hspace{-4mm} &
    HQ - \imgnote \hspace{-4mm} &
    LQ - \imgnote \hspace{-4mm} &
    PGDiff~\cite{yang2023pgdiff} \hspace{-4mm} &
    CodeFormer~\cite{zhou2022codeformer} \hspace{-4mm} & 
    DAEFR~\cite{tsai2024daefr} \hspace{-4mm} &
    FaceMe~\cite{liu2025faceme} \hspace{-4mm} & 
    MGFR~\cite{tao2025overcoming} \hspace{-4mm} &
    OSDFace~\cite{wang2025osdface} \hspace{-4mm} &
    HonestFace (ours) \hspace{-4mm}
    \\
    \end{tabular}
    \end{adjustbox}
}
\vspace{-3.5mm}
\caption{Visual comparison of synthetic datasets CelebHQRef-Test and Reface-Test. Please zoom in for a better view.}
\label{fig:vis-reface}
\vspace{-3.5mm}
\end{figure*}

\begin{table*}[t]
\footnotesize
\setlength{\tabcolsep}{0.5mm} 
\newcolumntype{C}{>{\centering\arraybackslash}X}
\newcolumntype{Y}{>{\centering\arraybackslash}X}
\begin{center}
\begin{tabularx}{0.8\textwidth}{c|c|CC|CCCC|CCCC}
\toprule[0.15em]
\rowcolor{color3} Datasets & Methods & LPIPS$\downarrow$ & DISTS$\downarrow$ & C-IQA$\uparrow$ & M-IQA$\uparrow$ & MUSIQ$\uparrow$ & FID$\downarrow$ & Deg.$\downarrow$ & LMD$\downarrow$ \\

\midrule[0.15em]
\multirow{6}{*}{\makecell{CelebHQRef-Test\,}} & ASFFNet~\cite{li2020asffnet512}  & 
0.1969 & \bb{0.1411} & 0.6067  &  0.5346 & 71.181 & \bb{45.618} & 35.636 & 3.6047 \\
& DMDNet~\cite{li2022dmdnet}  &
0.2191 & 0.1499 & 0.6417  &  0.5165 & 71.119 & 52.406 & 34.258 & 2.6437 \\
& PGDiff~\cite{yang2023pgdiff} & 
0.2674 & 0.1559 & 0.5735  &  0.4953 & 70.762 & 60.693 & 46.248 & 3.6478 \\
& FaceMe~\cite{liu2025faceme}  &
\bb{0.1945} & 0.1412 & \bb{0.6606}  &  0.5758 & \bb{71.598} & 46.423 & \bb{30.578} & \bb{2.0107} \\
& MGFR~\cite{tao2025overcoming}  &
0.2311 & 0.1520 & 0.6224  & \bb{0.6475} & 70.626 & 52.103 & 38.322 & 2.5274 \\
& \,HonestFace (ours)\, &
\rr{0.1809} & \rr{0.1246} & \rr{0.6673} & \rr{0.6858} & \rr{71.996} & \rr{41.021} & \rr{26.106} & \rr{1.7798} \\
\midrule[0.15em]
\multirow{6}{*}{\makecell{Reface-Test\,}} & ASFFNet~\cite{li2020asffnet512}  & 
\bb{0.2012} & \bb{0.1422} & 0.6043 & 0.5188 & 70.071 & 32.046 & 36.268 & 3.4611 \\ 
& DMDNet~\cite{li2022dmdnet}  &
0.2234 & 0.1510 & 0.6394 & 0.5028 & 70.278 & 36.867 & 35.282 & 2.6880 \\
& PGDiff~\cite{yang2023pgdiff} & 
0.2701 & 0.1575 & 0.5632 & 0.4822 & \bb{70.607} & 43.946 & 46.923 & 3.4423 \\
& FaceMe~\cite{liu2025faceme}  &
0.2021 & 0.1473 & \bb{0.6595} & 0.5881 & 70.426 & \bb{31.483} & \bb{32.021} & \bb{1.9793} \\
& MGFR~\cite{tao2025overcoming}  &
0.2307 & 0.1535 & 0.6467 & \bb{0.6436} & 70.464 & 34.167 & 38.762 & 2.4641 \\
& \,HonestFace (ours)\, &
\rr{0.1784} & \rr{0.1235} & \rr{0.6818} & \rr{0.6717} & \rr{71.608} & \rr{25.755} & \rr{28.173} & \rr{1.7369} \\ 
\bottomrule[0.15em]
\end{tabularx}
\end{center}
\vspace{-0.2mm}
\footnotesize
\caption{Comparison with the reference-based methods. C-IQA stands for CLIP-IQA, and M-IQA stands for MANIQA.} 
\vspace{-8.5mm}
\label{tab:reference}
\end{table*}

\vspace{-2.0mm}
\subsection{Face Alignment Guidance}\label{sec:losses}
\vspace{-0.2mm}

Although the adversarial loss aligns the distributions of restored and real images, it does not focus on subtle facial properties. For example, the faithful reproduction of textures in hair and skin—such as fine wrinkles and unique blemishes, is crucial for ``honesty''.  Therefore, our model employs several loss functions to encourage both global harmony and local coherence. These include an identity loss, a global perceptual loss, and our proposed masked face alignment (MFA) method for enhancing localized perceptual fidelity.

\vspace{1mm}
\noindent\textbf{Identity Preservation Loss}. 
Preserving facial identity is essential for truthful face restoration. Face recognition models~\cite{deng2019arcface,kim2022adaface} map aligned faces to distinctive identity embeddings. To reduce bias from any single model, we use an ensemble of $M$ distinct pre-trained facial identity extractors, denoted as $\mathcal{F} = \{f_1, f_2, \dots, f_M\}$. Each $f_i \in \mathcal{F}$ maps an input face to an identity embedding. We define the identity loss $\mathcal{L}_{\text{ID}}$ to enforce identity consistency between the restored image $\hat{x}_H$ and its HQ counterpart $x_H$ across all extractors:
\begin{equation}
    \mathcal{L}_{\text{ID}} = \sum_{i=1}^{M} \left(1 - \frac{f_i(x_H) \cdot f_i(\hat{x}_H)}{\|f_i(x_H)\|_2 \cdot \|f_i(\hat{x}_H)\|_2}\right).
\end{equation}

\vspace{1mm}
\noindent\textbf{Global Perceptual Loss}. 
To perceptually align the overall structure and texture of the restored image $\hat{x}_H$ with the ground truth $x_H$, we use a global perceptual loss $\mathcal{L}_{\text{Per}}$ based on DISTS~\cite{ding2020dists}, a metric that correlates well with subjective image-quality judgments. Following prior works~\cite{li2024dfosd,wang2025osdface}, we adopt its edge-aware variant, EA-DISTS, to further emphasize structural consistency:
\begin{equation}
    \mathcal{L}_{\text{Per}} = \mathcal{L}_{\text{dists}}(x_H, \hat{x}_H) + \mathcal{L}_{\text{dists}}(\mathcal{S}(x_H), \mathcal{S}(\hat{x}_H)),
\end{equation}
where $\mathcal{L}_{\text{dists}}$ denotes the DISTS score, and $\mathcal{S}(\cdot)$ is the Sobel operator used to extract edge maps.

\vspace{1mm}
\noindent\textbf{Local Perceptual Loss via Masked Face Alignment}. 
Human attention prioritizes distinctive facial details, such as moles, scars, and wrinkles around the eyes and mouth, whereas cheeks and hair require only natural textures rather than exact replication. Previous approaches~\cite{wang2025osdface} often treated all regions equally, potentially expending model capacity on less critical areas while missing fine details.
To address this, our proposed masked face alignment (MFA), shown in Fig.~\ref{fig:mfa}, introduces a localized perceptual loss $\mathcal{L}_{\text{MFA}}$ that strategically focuses on perceptually significant facial regions. 

Firstly, MFA creates an attention mask derived from heatmap predictions on the input image.
Heatmap generation is a well-established technique in tasks such as facial landmark detection~\cite{Payer2016,bulat2018super,wang2019adaptive,mccouat2022contour}. These methods typically perform spatial feature matching to capture local details. Each heatmap channel corresponds to a distinct landmark or facial region. By aggregating these channel-wise heatmaps, we subsequently construct a composite mask highlighting perceptually important areas. Additionally, heatmap generation techniques are robust to severe image degradation~\cite{bulat2018super}, thereby allowing us to obtain reliable attention masks even from LQ inputs.

Specifically, we use a pre-trained heatmap detection model $X_{\text{HM}}$ to process the LQ input $x_L$.
The channel-wise maximum produces a single-channel attention map:
\begin{equation}
    \mathcal{M}_\text{raw} = \text{max}_{\text{channel}}(X_{\text{HM}}(x_L)) \in \mathbb{R}^{H \times W}. 
\end{equation}

To enhance the contrast of the attention map and render it suitable for blending, we apply an exponential transformation with a sharpening parameter $k > 0$, followed by min-max normalization stabilized by a small constant $\epsilon > 0$. The final normalized attention mask $\mathcal{M} \in [0,1]^{H \times W}$ is then obtained as:
\begin{equation}
    \label{eq:heatmap_norm}
    \mathcal{M} = \frac{1 - e^{-k \cdot \mathcal{M}_\text{raw}} - \min(1 - e^{-k \cdot \mathcal{M}_\text{raw}})}{\max(1 - e^{-k \cdot \mathcal{M}_\text{raw}}) - \min(1 - e^{-k \cdot \mathcal{M}_\text{raw}}) + \epsilon}.
\end{equation}

\begin{figure*}[t]
\footnotesize
\centering
\newcommand{\imgid}{gsl067_large}
\newcommand{\imgnote}{01}
\scalebox{0.98}{
    \hspace{-0.4cm}
    \begin{adjustbox}{valign=t}
    \begin{tabular}{ccccccccccc}
    \includegraphics[width=0.098\textwidth]{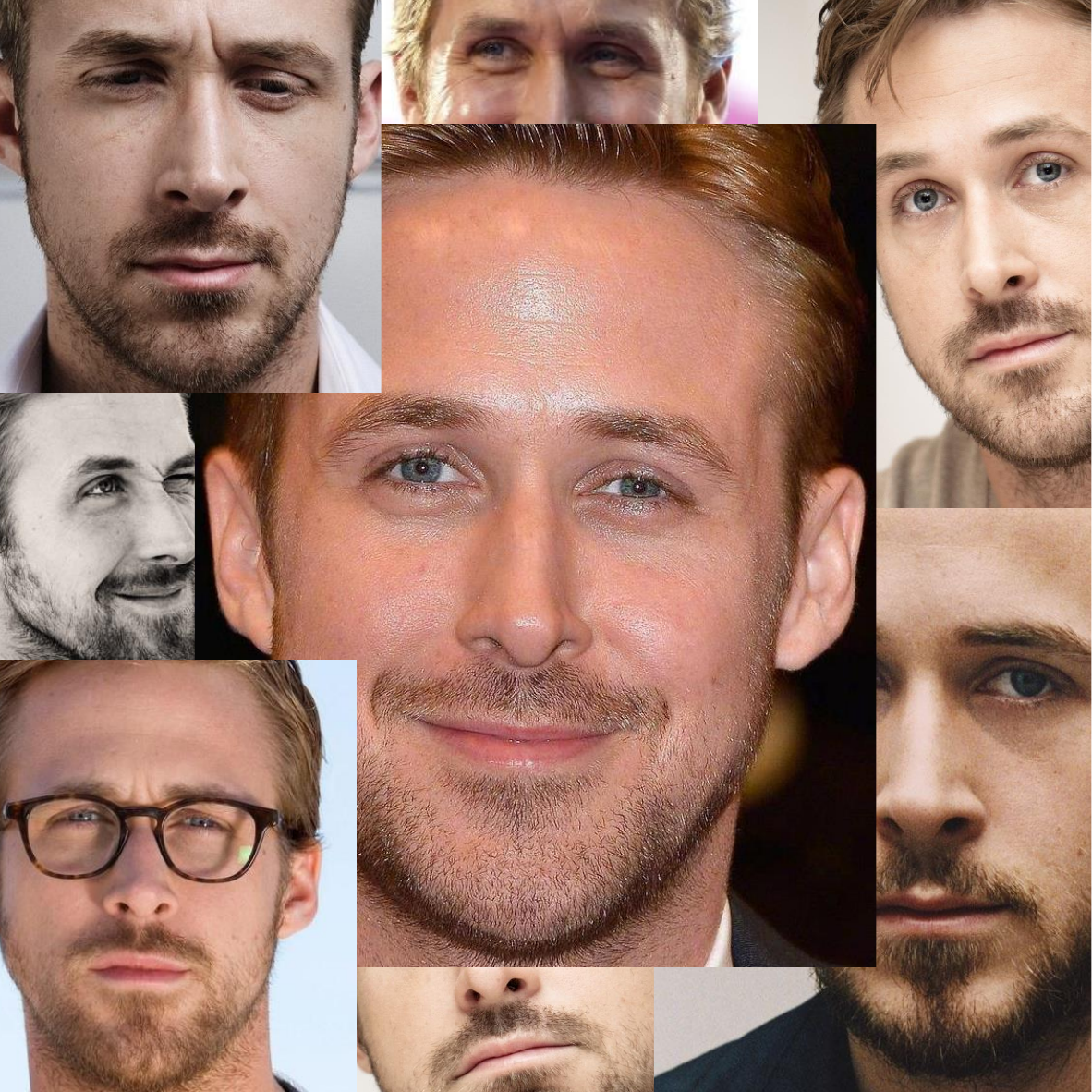} \hspace{-4mm} &
    \includegraphics[width=0.098\textwidth]{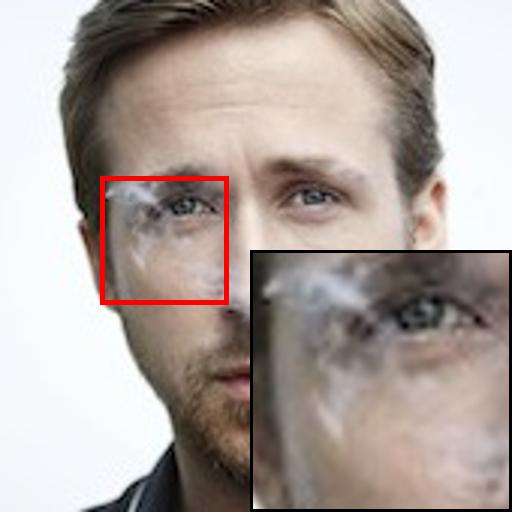} \hspace{-4mm} &
    \includegraphics[width=0.098\textwidth]{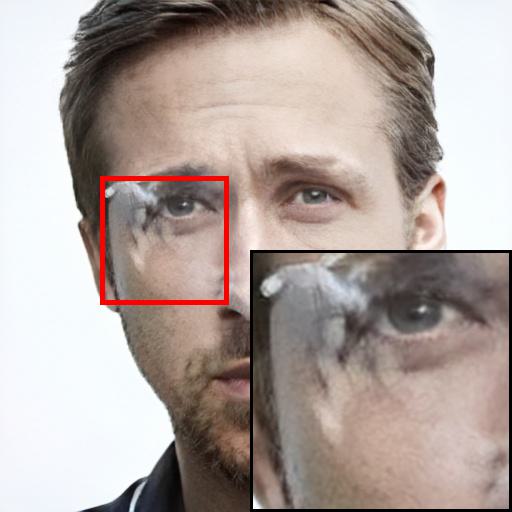} \hspace{-4mm} &
    \includegraphics[width=0.098\textwidth]{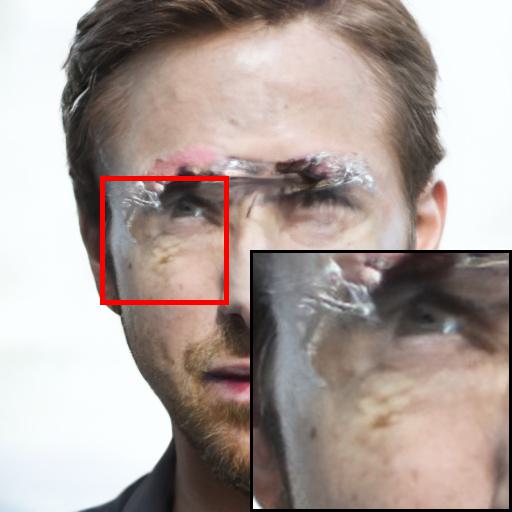} \hspace{-4mm} &
    \includegraphics[width=0.098\textwidth]{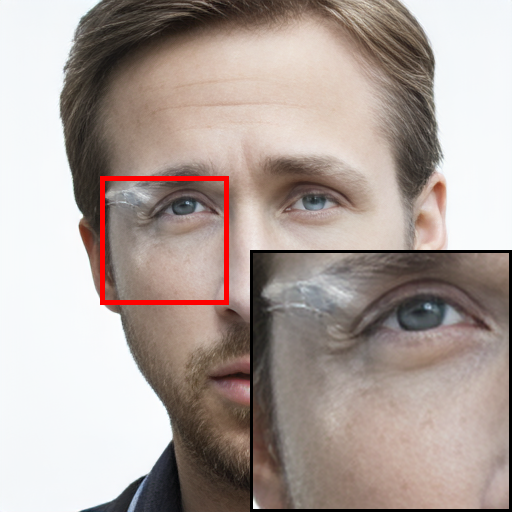} \hspace{-4mm} &
    \includegraphics[width=0.098\textwidth]{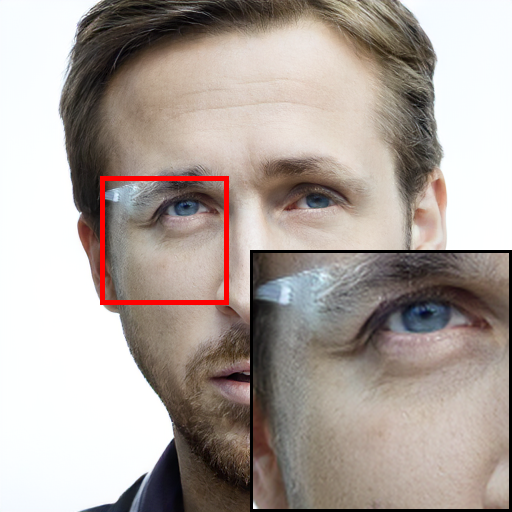} \hspace{-4mm} &
    \includegraphics[width=0.098\textwidth]{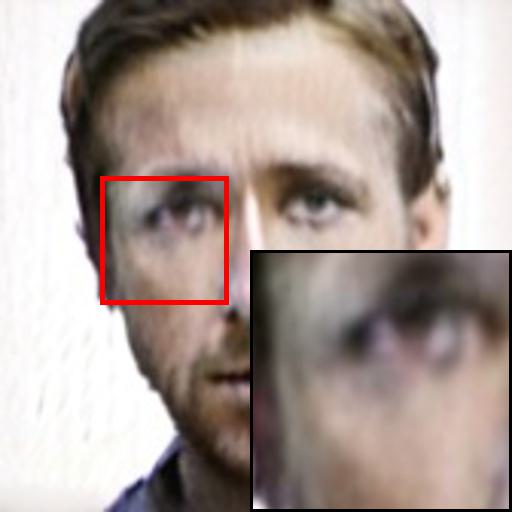} \hspace{-4mm} &
    \includegraphics[width=0.098\textwidth]{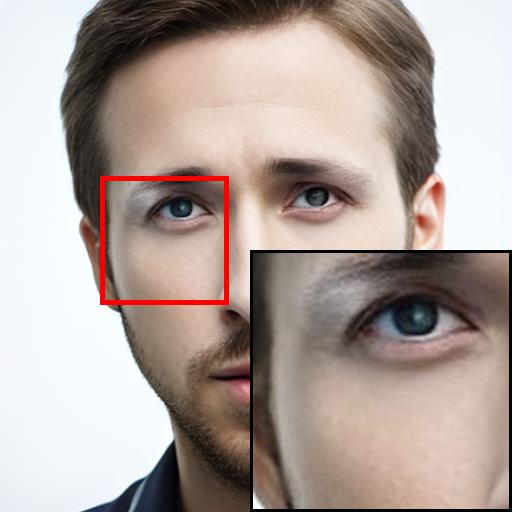} \hspace{-4mm} &
    \includegraphics[width=0.098\textwidth]{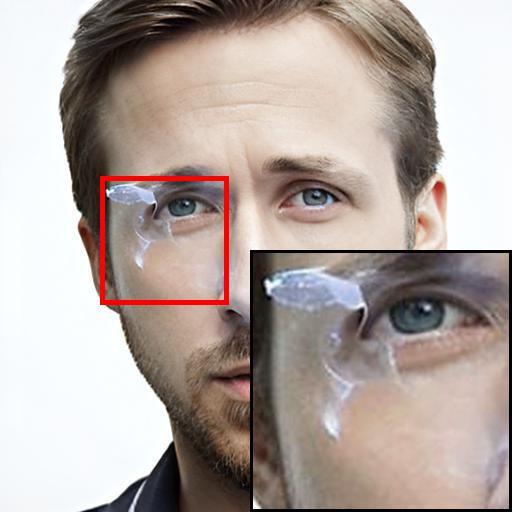} \hspace{-4mm} &
    \includegraphics[width=0.098\textwidth]{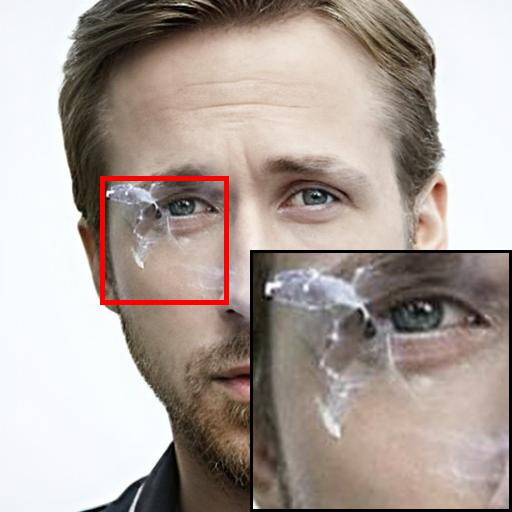} \hspace{-4mm} 
    \\
    Ref - \imgnote \hspace{-4mm} &
    LQ - \imgnote \hspace{-4mm} &
    ASFFNet~\cite{li2020asffnet512} \hspace{-4mm} &
    PGDiff~\cite{yang2023pgdiff} \hspace{-4mm} &
    CodeFormer~\cite{zhou2022codeformer} \hspace{-4mm} & 
    DAEFR~\cite{tsai2024daefr} \hspace{-4mm} &
    FaceMe~\cite{liu2025faceme} \hspace{-4mm} & 
    OSEDiff~\cite{wu2024osediff} \hspace{-4mm} &
    OSDFace~\cite{wang2025osdface} \hspace{-4mm} &
    HonestFace (ours) \hspace{-4mm}
    \\
    \end{tabular}
    \end{adjustbox}
}
\renewcommand{\imgid}{zs033_large}
\renewcommand{\imgnote}{02}
\scalebox{0.98}{
    \hspace{-0.4cm}
    \begin{adjustbox}{valign=t}
    \begin{tabular}{ccccccccccc}
    \includegraphics[width=0.098\textwidth]{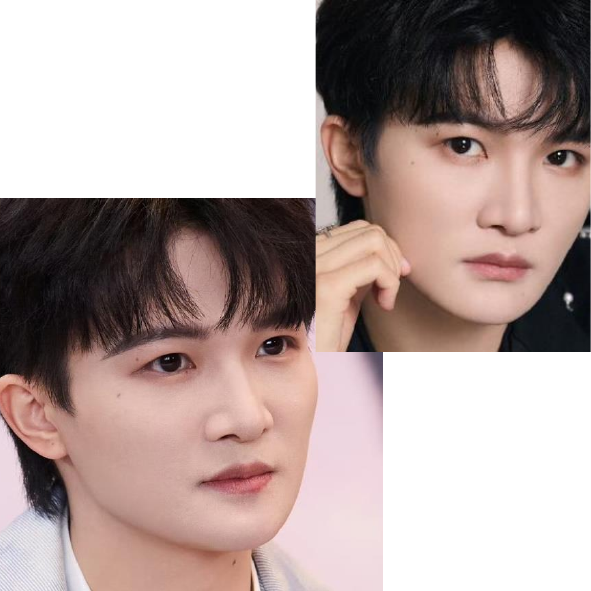} \hspace{-4mm} &
    \includegraphics[width=0.098\textwidth]{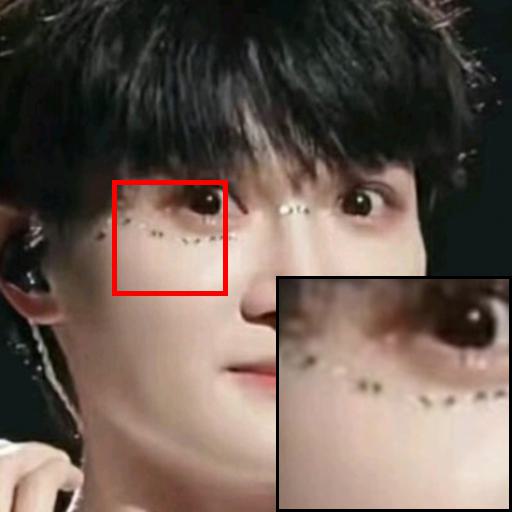} \hspace{-4mm} &
    \includegraphics[width=0.098\textwidth]{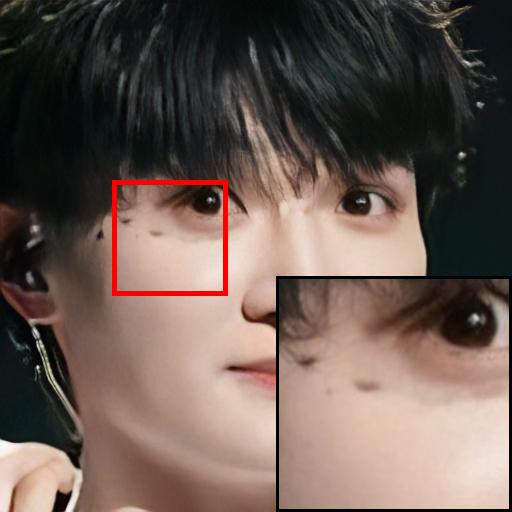} \hspace{-4mm} &
    \includegraphics[width=0.098\textwidth]{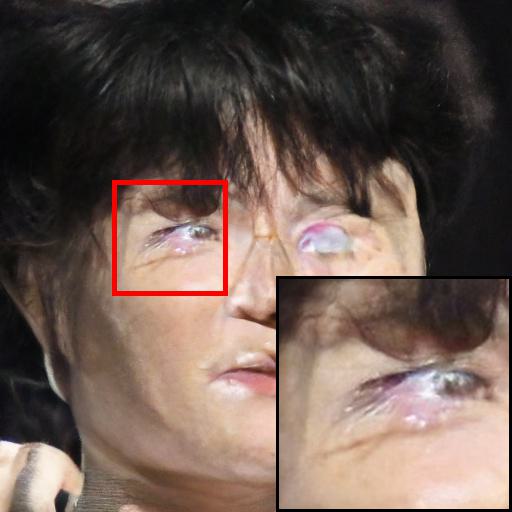} \hspace{-4mm} &
    \includegraphics[width=0.098\textwidth]{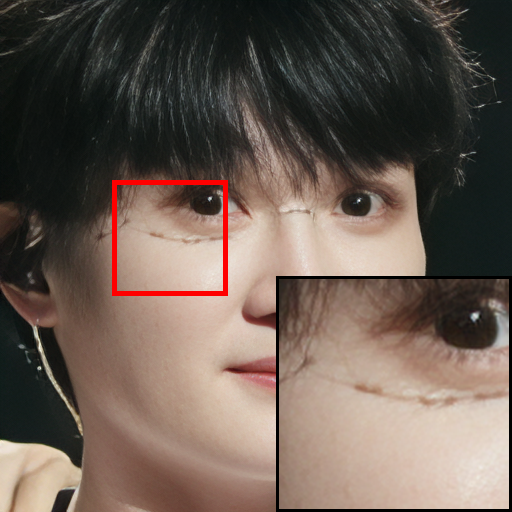} \hspace{-4mm} &
    \includegraphics[width=0.098\textwidth]{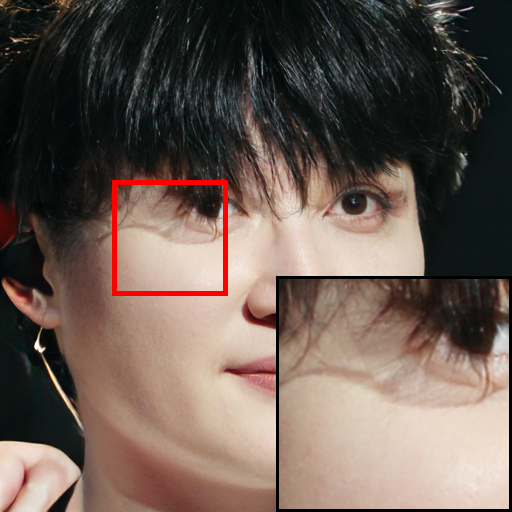} \hspace{-4mm} &
    \includegraphics[width=0.098\textwidth]{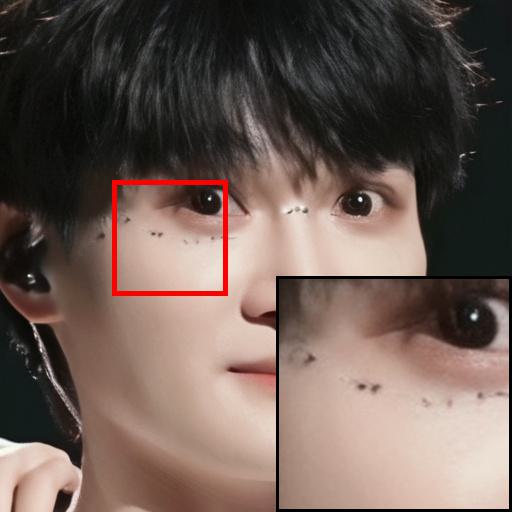} \hspace{-4mm} &
    \includegraphics[width=0.098\textwidth]{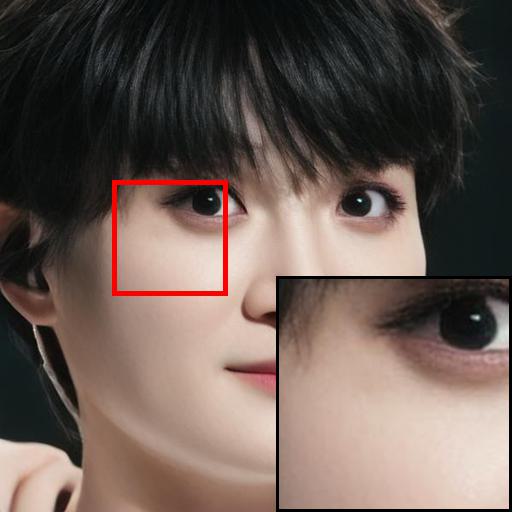} \hspace{-4mm} &
    \includegraphics[width=0.098\textwidth]{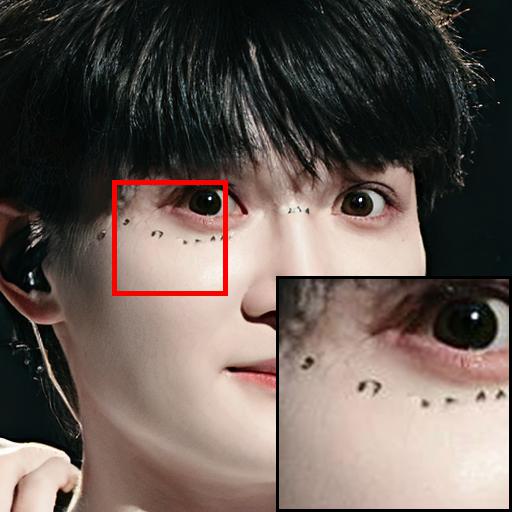} \hspace{-4mm} &
    \includegraphics[width=0.098\textwidth]{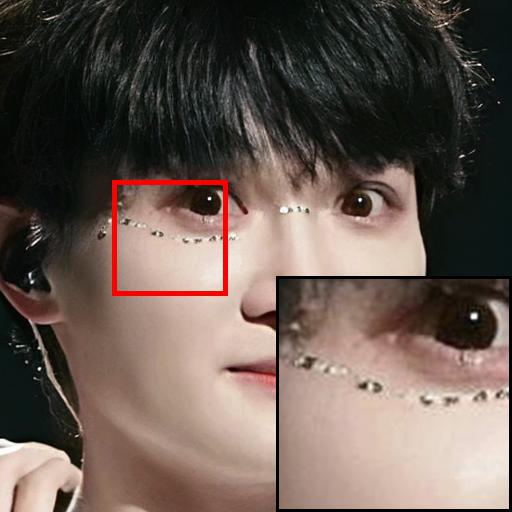} \hspace{-4mm} 
    \\
    Ref - \imgnote \hspace{-4mm} &
    LQ - \imgnote \hspace{-4mm} &
    ASFFNet~\cite{li2020asffnet512} \hspace{-4mm} &
    PGDiff~\cite{yang2023pgdiff} \hspace{-4mm} &
    CodeFormer~\cite{zhou2022codeformer} \hspace{-4mm} & 
    DAEFR~\cite{tsai2024daefr} \hspace{-4mm} &
    FaceMe~\cite{liu2025faceme} \hspace{-4mm} & 
    OSEDiff~\cite{wu2024osediff} \hspace{-4mm} &
    OSDFace~\cite{wang2025osdface} \hspace{-4mm} &
    HonestFace (ours) \hspace{-4mm}
    \\
    \end{tabular}
    \end{adjustbox}
}
\renewcommand{\imgid}{hcy033_large}
\renewcommand{\imgnote}{03}
\scalebox{0.98}{
    \hspace{-0.4cm}
    \begin{adjustbox}{valign=t}
    \begin{tabular}{ccccccccccc}
    \includegraphics[width=0.098\textwidth]{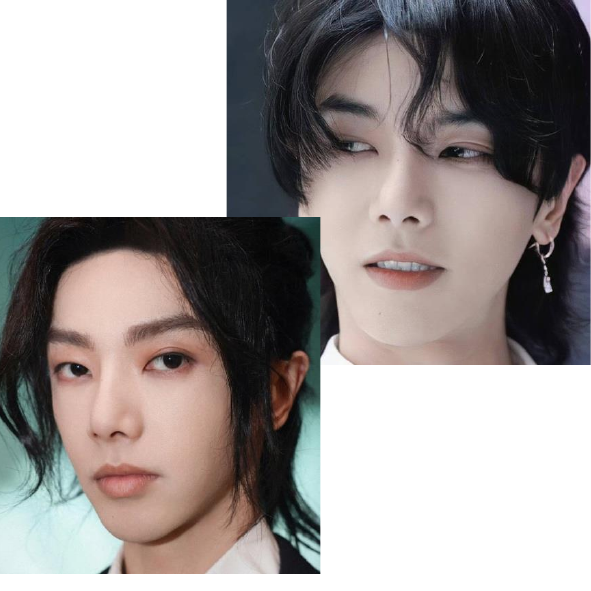} \hspace{-4mm} &
    \includegraphics[width=0.098\textwidth]{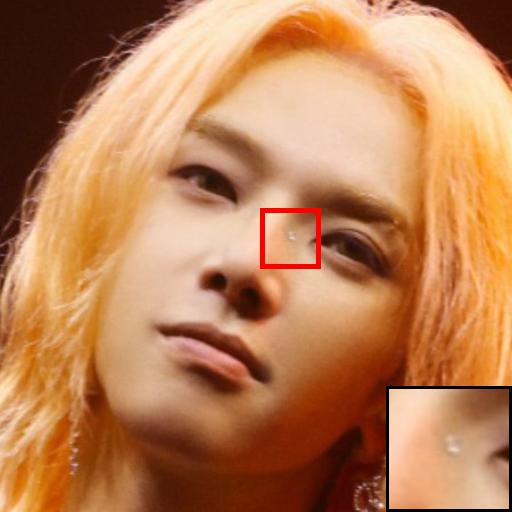} \hspace{-4mm} &
    \includegraphics[width=0.098\textwidth]{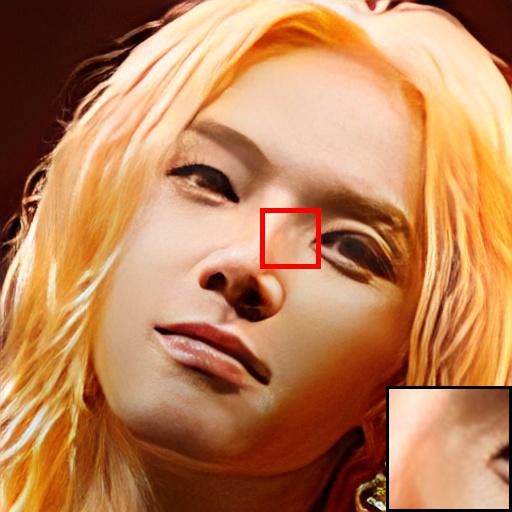} \hspace{-4mm} &
    \includegraphics[width=0.098\textwidth]{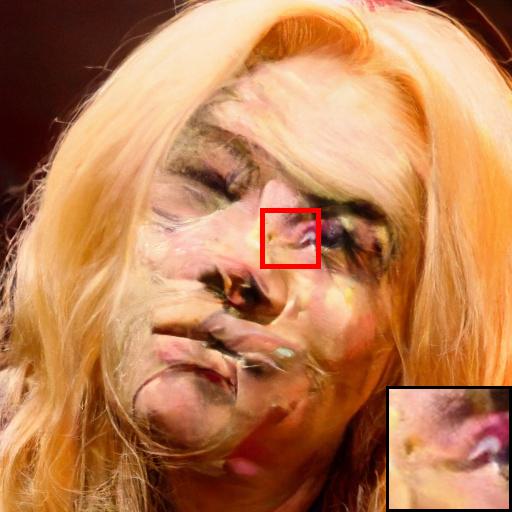} \hspace{-4mm} &
    \includegraphics[width=0.098\textwidth]{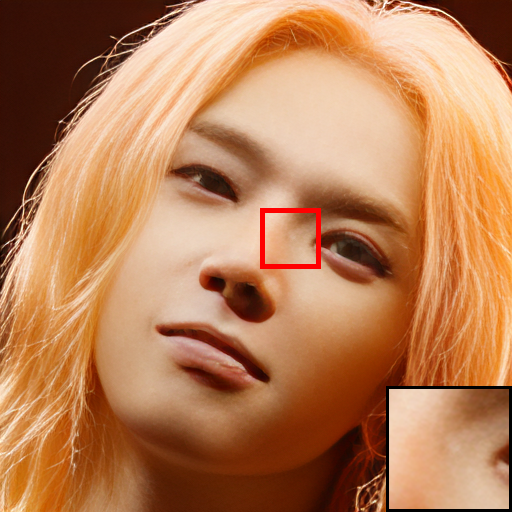} \hspace{-4mm} &
    \includegraphics[width=0.098\textwidth]{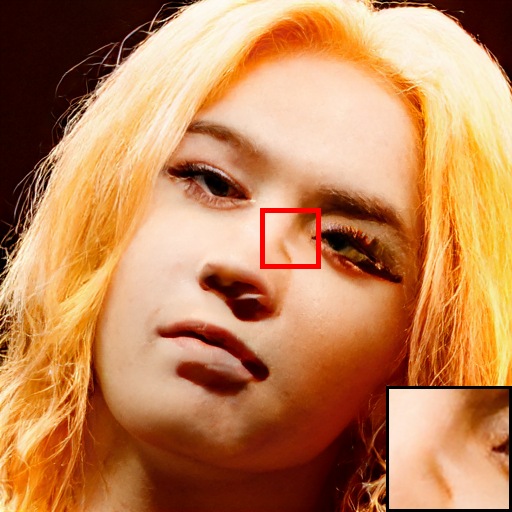} \hspace{-4mm} &
    \includegraphics[width=0.098\textwidth]{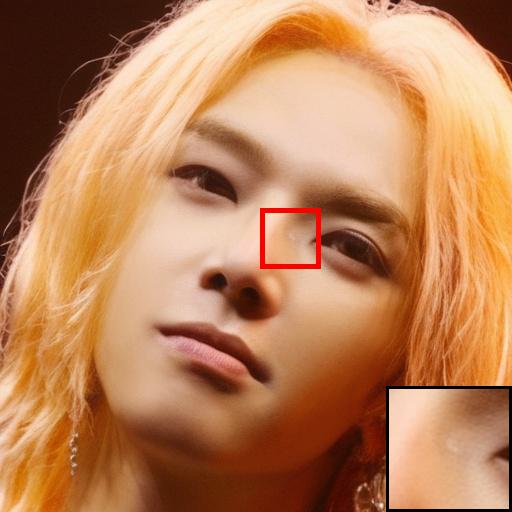} \hspace{-4mm} &
    \includegraphics[width=0.098\textwidth]{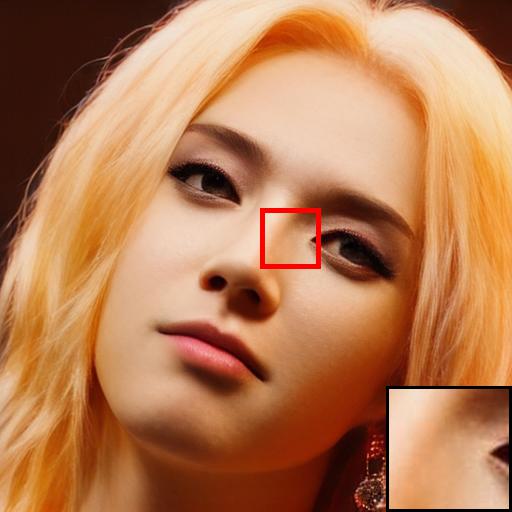} \hspace{-4mm} &
    \includegraphics[width=0.098\textwidth]{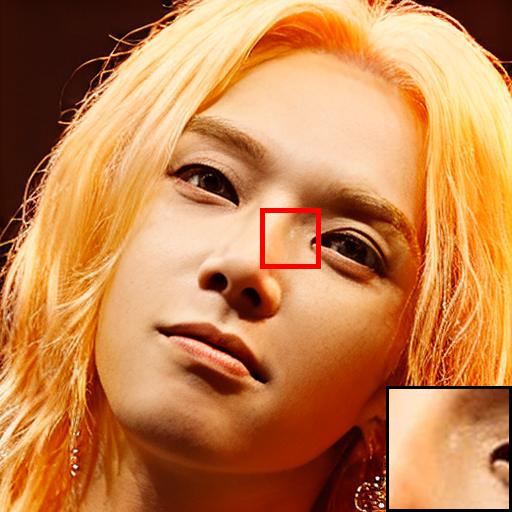} \hspace{-4mm} &
    \includegraphics[width=0.098\textwidth]{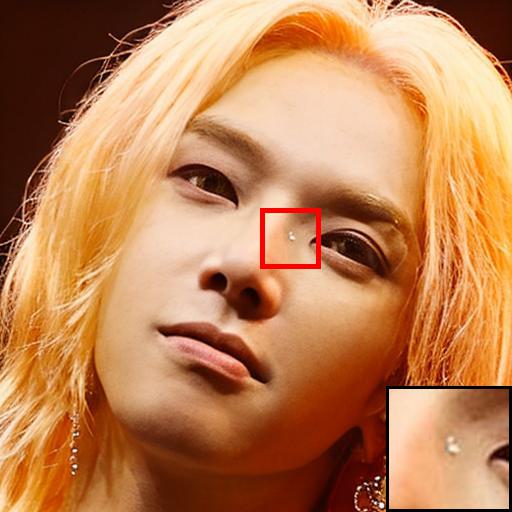} \hspace{-4mm} 
    \\
    Ref - \imgnote \hspace{-4mm} &
    LQ - \imgnote \hspace{-4mm} &
    ASFFNet~\cite{li2020asffnet512} \hspace{-4mm} &
    PGDiff~\cite{yang2023pgdiff} \hspace{-4mm} &
    CodeFormer~\cite{zhou2022codeformer} \hspace{-4mm} & 
    DAEFR~\cite{tsai2024daefr} \hspace{-4mm} &
    FaceMe~\cite{liu2025faceme} \hspace{-4mm} & 
    OSEDiff~\cite{wu2024osediff} \hspace{-4mm} &
    OSDFace~\cite{wang2025osdface} \hspace{-4mm} &
    HonestFace (ours) \hspace{-4mm}
    \\
    \end{tabular}
    \end{adjustbox}
}
\renewcommand{\imgid}{bjt013_large}
\renewcommand{\imgnote}{04}
\scalebox{0.98}{
    \hspace{-0.4cm}
    \begin{adjustbox}{valign=t}
    \begin{tabular}{ccccccccccc}
    \includegraphics[width=0.098\textwidth]{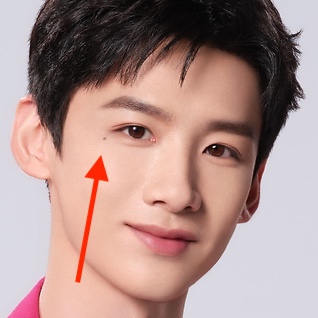} \hspace{-4mm} &
    \includegraphics[width=0.098\textwidth]{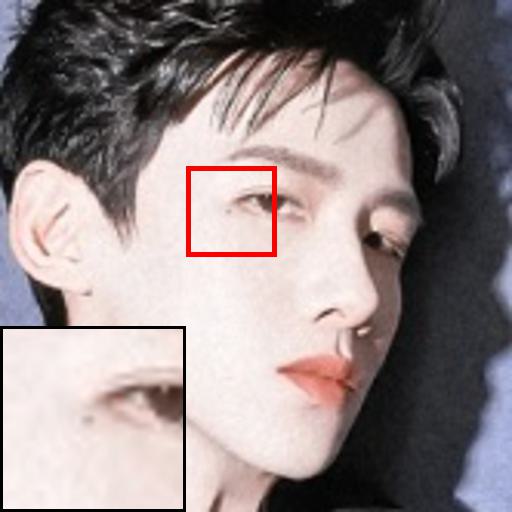} \hspace{-4mm} &
    \includegraphics[width=0.098\textwidth]{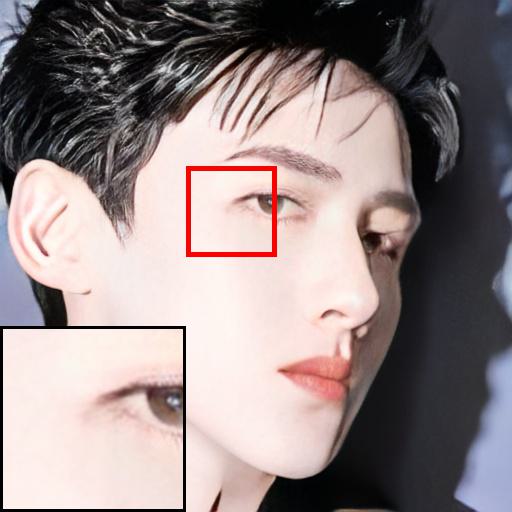} \hspace{-4mm} &
    \includegraphics[width=0.098\textwidth]{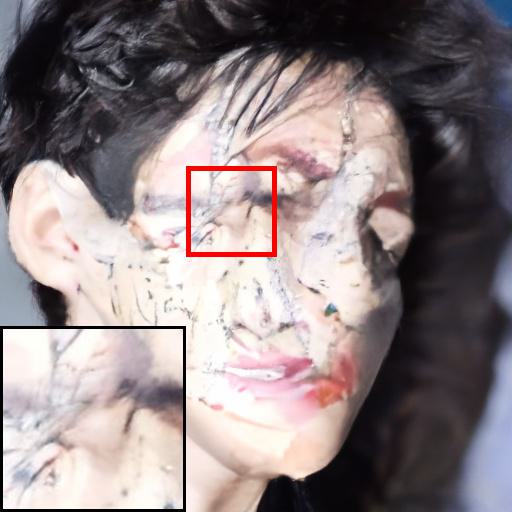} \hspace{-4mm} &
    \includegraphics[width=0.098\textwidth]{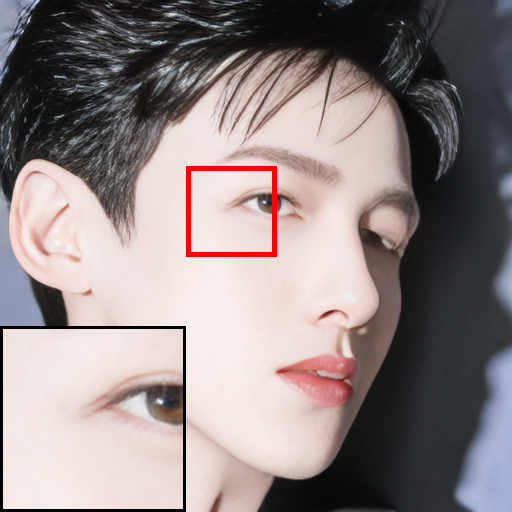} \hspace{-4mm} &
    \includegraphics[width=0.098\textwidth]{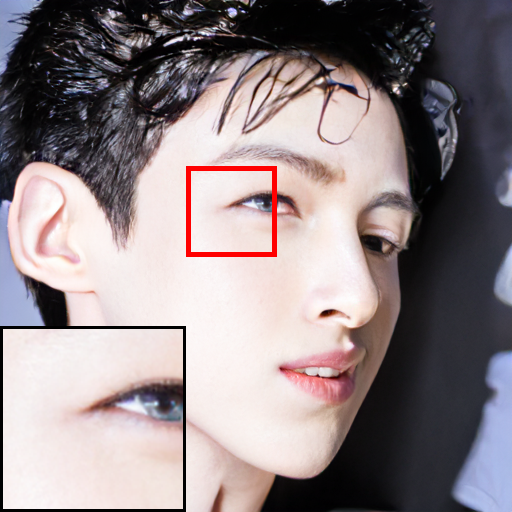} \hspace{-4mm} &
    \includegraphics[width=0.098\textwidth]{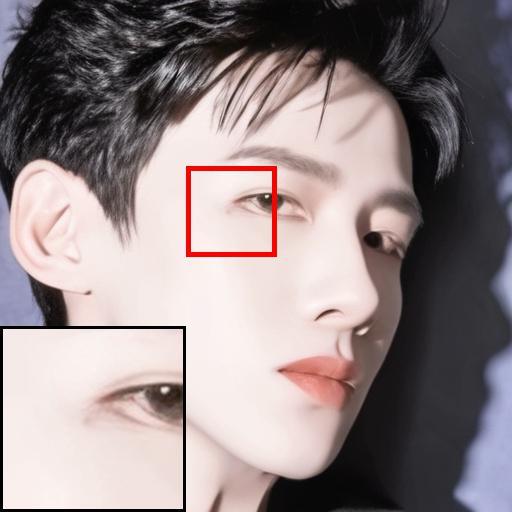} \hspace{-4mm} &
    \includegraphics[width=0.098\textwidth]{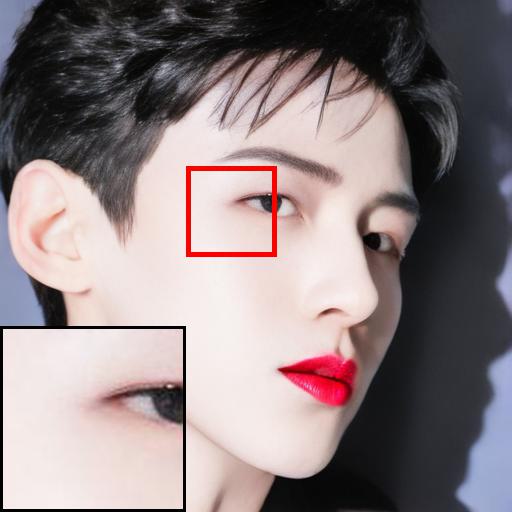} \hspace{-4mm} &
    \includegraphics[width=0.098\textwidth]{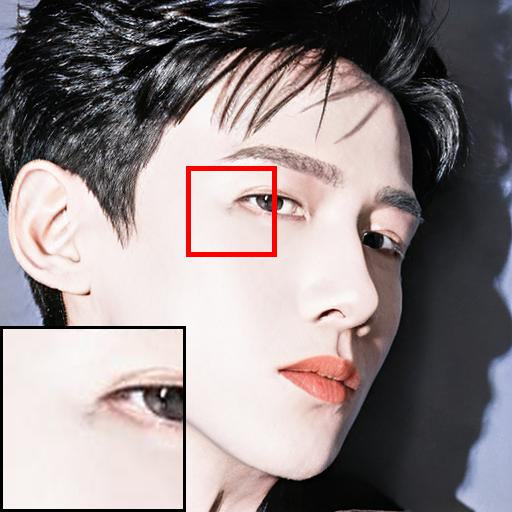} \hspace{-4mm} &
    \includegraphics[width=0.098\textwidth]{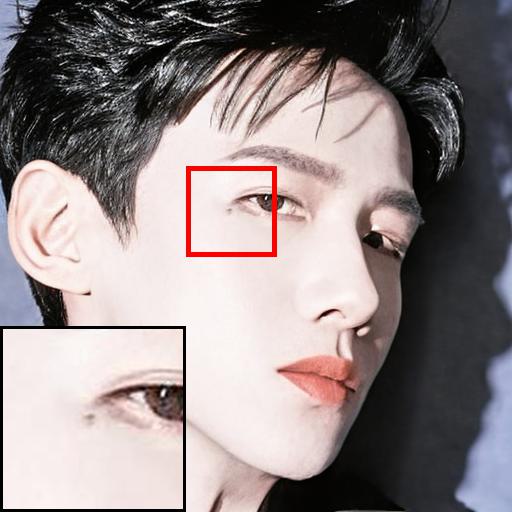} \hspace{-4mm} 
    \\
    Ref - \imgnote \hspace{-4mm} &
    LQ - \imgnote \hspace{-4mm} &
    ASFFNet~\cite{li2020asffnet512} \hspace{-4mm} &
    PGDiff~\cite{yang2023pgdiff} \hspace{-4mm} &
    CodeFormer~\cite{zhou2022codeformer} \hspace{-4mm} & 
    DAEFR~\cite{tsai2024daefr} \hspace{-4mm} &
    FaceMe~\cite{liu2025faceme} \hspace{-4mm} & 
    OSEDiff~\cite{wu2024osediff} \hspace{-4mm} &
    OSDFace~\cite{wang2025osdface} \hspace{-4mm} &
    HonestFace (ours) \hspace{-4mm}
    \\
    \end{tabular}
    \end{adjustbox}
}
\vspace{-2mm}
\caption{Visual comparison on the real-world MultiRefCeleb-Test dataset. Please zoom in for a better view.}
\label{fig:vis-realworld}
\vspace{-2mm}
\end{figure*}

\begin{table*}[t]
\vspace{-0.5mm}
\footnotesize
\setlength{\tabcolsep}{0.4mm}

\newcolumntype{?}{!{\vrule width 1pt}}
\newcolumntype{C}{>{\centering\arraybackslash}X}
\begin{center}
\scalebox{1.0}{
\begin{tabularx}{\textwidth}{l|*{6}{C}|*{6}{C}|*{2}{C}}
\toprule[0.15em]
\rowcolor{color3} & \multicolumn{6}{c|}{CelebHQRef-Test} & \multicolumn{6}{c|}{Reface-Test} & \multicolumn{2}{c}{\,\,\,MultiRefCeleb-Test} \\
\rowcolor{color3}
\multirow{-2}{*}{Methods} & PSNR$\uparrow$ & SSIM$\uparrow$ & LPIPS$\downarrow$ & DISTS$\downarrow$ & M-IQA$\uparrow$ & FID$\downarrow$ & PSNR$\uparrow$ & SSIM$\uparrow$ & LPIPS$\downarrow$ & DISTS$\downarrow$ & M-IQA$\uparrow$ & FID$\downarrow$ & M-IQA$\uparrow$ & FID$\downarrow$ \\
\midrule[0.15em]

CodeFormer~\cite{zhou2022codeformer} & 
\rr{25.49} & 0.7100 & \rr{0.1791} & 0.1419 & 0.5827 & 48.37 &
\rr{25.65} & 0.7082 & \bb{0.1872} & 0.1497 & 0.5699 & 33.57 & 
0.5483 & 60.61 \\

DAEFR~\cite{tsai2024daefr} & 
22.85 & 0.6549 & 0.2058 & 0.1439 & 0.6019 & 49.35 &
23.07 & 0.6574 & 0.2092 & 0.1498 & 0.5842 & 34.10 & 
0.5541 & \rr{48.52} \\
\midrule
OSEDiff~\cite{wu2024osediff} & 
24.07 & \bb{0.7105} & 0.2077 & 0.1533 & 0.5229 & 57.06 &
24.21 & \bb{0.7110} & 0.2076 & 0.1553 & 0.5122 & 46.31 & 
\bb{0.7057} & 84.85 \\
OSDFace~\cite{wang2025osdface} & 
23.51 & 0.6742 & 0.2050 & \bb{0.1373} & \bb{0.6316} & \bb{46.51} &
23.54 & 0.6718 & 0.2156 & \bb{0.1481} & \bb{0.6257} & \bb{32.59} & 
\rr{0.7148} & 63.11 \\
\midrule
HonestFace (ours) & 
\bb{25.17} & \rr{0.7147} & \bb{0.1809} & \rr{0.1246} & \rr{0.6858} & \rr{41.02} &
\bb{25.25} & \rr{0.7118} & \rr{0.1784} & \rr{0.1235} & \rr{0.6717} & \rr{25.75} & 
0.6649 & \bb{57.32} \\
\bottomrule[0.15em]
\end{tabularx}}
\end{center}
\caption{Quantitative comparison with the no-reference methods. M-IQA stands for MANIQA. }
\label{tab:RealWorld}
\vspace{-8mm}
\end{table*}

The normalized mask $\mathcal{M}$ is then used to blend the ground truth $x_H$ and restored image $\hat{x}_H$ with a neutral background $B \in \mathbb{R}^{H \times W \times 3}$. This procedure focuses the subsequent perceptual loss calculation on the identified salient regions:
\begin{equation}
\begin{aligned}
    x_{\text{MFA},H} &= \mathcal{M} \odot x_H + (1-\mathcal{M}) \odot B,\\
    \hat{x}_{\text{MFA},H} &= \mathcal{M} \odot \hat{x}_H + (1-\mathcal{M}) \odot B,
\end{aligned}
\end{equation}
where $\odot$ denotes element-wise multiplication. It targets perceptually important areas for targeted color and texture fidelity, mitigating global color shifts and facial texture inconsistencies.

The local perceptual loss, $\mathcal{L}_{\text{MFA}}$, is then computed using the LPIPS metric~\cite{zhang2018lpips}, known for its correlation with human perception of image similarity, on these masked images:
\begin{equation}
    \mathcal{L}_{\text{MFA}} = \mathcal{L}_{\text{lpips}}(x_{\text{MFA},H}, \hat{x}_{\text{MFA},H}).
\end{equation}

The targeted loss directs the model to focus on the regions identified by the heatmap, ensuring that subtle textures and natural colors are reconstructed accurately.

\vspace{-1.5mm}
\section{Experiments}\label{sec:exp}
\vspace{-0.2mm}
\subsection{Experimental Settings}
\vspace{-0.2mm}

\noindent\textbf{Datasets}.
HonestFace is fine-tuned on the FFHQ-Ref~\cite{hsiao2024refldm}, Reface-HQ training dataset~\cite{tao2025overcoming} and the last 905 identities from CelebRef-HQ~\cite{li2022dmdnet}.
FFHQ-Ref comprises 20,406 face images from FFHQ~\cite{karras2019ffhq} and their corresponding reference images.
Reface-HQ contains 23,500 high-quality face images from 5,250 identities, with 4,870 used for training and 380 reserved for testing (Reface-Test). CelebRef-HQ includes 1,005 identities and a total of 10,555 images. All images are resized to 512$\times$512 before training. Synthetic low-quality data is generated using the same degradation pipeline as VQFR~\cite{gu2022vqfr}.
For inference, we evaluate on Reface-Test and the first 100 identities from CelebRef-HQ using the same degradation config as training.

Furthermore, we introduce \textit{MultiRefCeleb-Test}, a large-scale test dataset. Most existing benchmarks use high-quality images with synthetic degradation. In contrast, our dataset includes over 9,000 real-world images of 400+ celebrities from public sources, including self-collected images in Fig.~\ref{fig:teaser}. As illustrated in Fig.~\ref{fig:dataset}, each identity within the dataset is provided with multiple high-quality references and several low-quality queries. We quantified degradation using the weighted NR-IQA score~\cite{ntire2025face,ntire2026face}. All images score below 3, indicating severe quality loss. The distribution is: 0.4\% ($<$1), 3.5\% (1--1.5), {15.8\%} (1.5--2), {42.7\%} (2--2.5), and {37.6\%} (2.5--3). 

\vspace{1mm}
\noindent\textbf{Metrics}.
We evaluate pixel-level restoration fidelity using PSNR and SSIM.
Perceptual metrics include LPIPS~\cite{zhang2018lpips}, DISTS~\cite{ding2020dists}, and no-reference methods such as CLIP-IQA~\cite{wang2022clipiqa}, MANIQA~\cite{yang2022maniqa}, and MUSIQ~\cite{ke2021musiq}.
For the synthetic data, FID is calculated between the ground truth and restored faces, while for our new real-world MultiRefCeleb-Test, FID is computed against the FFHQ dataset, to ensure fair comparison with prior work~\cite{gu2022vqfr, liu2025faceme, tsai2024daefr, wang2025osdface}.
 To assess identity consistency, we use the angular difference between ArcFace embeddings (``Deg.'') and the L2 landmark distance (``LMD'').

\begin{table*}[t]
        \centering
        \footnotesize 
        \captionsetup[subtable]{skip=2pt}
        \setlength{\tabcolsep}{1mm} 
        \newcolumntype{C}{>{\centering\arraybackslash}X}
        \newcolumntype{S}{>{\centering\arraybackslash}c}
    
	    \subfloat[\small Real-world test: MultiRefCeleb-Test.\label{tab:real-world}]{ 
        \scalebox{0.86}{
			\begin{tabular}{S|c c c c c c}
				\toprule
				\rowcolor{color3} Methods & C-IQA$\uparrow$ & M-IQA$\uparrow$ & MUSIQ$\uparrow$ & FID$\downarrow$ \\
				\midrule
				  ASFFNet~\cite{li2020asffnet512} & 0.5899 & \bb{0.6154} & 67.999 & \bb{59.504}  \\
			      DMDNet~\cite{li2022dmdnet}      & \rr{0.6741} & 0.5313 & \bb{70.466} & 64.954  \\
			      PGDiff~\cite{yang2023pgdiff}    & 0.4850 & 0.4867 & 62.407 & 83.092  \\
                  FaceMe~\cite{liu2025faceme}     & 0.6443 & 0.5866 & 66.936 & 59.821  \\
                  HonestFace (ours)               & \bb{0.6516} & \rr{0.6649} & \rr{73.716} & \rr{57.316}  \\
				\bottomrule
	        \end{tabular}
            }
        }\hspace{1.5mm}\vspace{0mm}
        \subfloat[\small MS-SWD~\cite{he2024ms-swd} metric for color IQA.\label{tab:msswd}]{
		    \scalebox{1.0}{\hspace{-1.5mm}
			\begin{tabular}{l| c c}
				\toprule
				\rowcolor{color3} Methods & CelebHQRef-Test & Reface-Test \\
				\midrule
				MGFR~\cite{tao2025overcoming} & 0.8000 & 0.7674 \\
				OSEDiff~\cite{wu2024osediff}  & 0.4646 & 0.4388 \\
                    OSDFace~\cite{wang2025osdface}& 0.5634 & 0.5716 \\
                    HonestFace (ours)             & \rr{0.3461} & \rr{0.3356} \\
				\bottomrule
	        \end{tabular}}
        }\hspace{0mm}\vspace{0mm}
        \subfloat[\small Ablation study of IDE. \label{tab:ide}]{
		    \scalebox{0.86}{
			\begin{tabular}{S|SS|c c c c c c}
				\toprule
				\rowcolor{color3} FFE & FIE w. Arc & FIE w. Ada & LPIPS$\downarrow$ & DISTS$\downarrow$ & Deg.$\downarrow$ & LMD$\downarrow$ \\
				\midrule
				  \xmark & \xmark & \xmark & 0.1855 & 0.1282 & 30.052 & 1.9013 \\
			      \cmark & \xmark & \xmark & 0.1835 & 0.1265 & 29.322 & 1.8873 \\
			      \xmark & \cmark & \xmark & 0.1863 & 0.1288 & 27.338 & 1.8790 \\
                  \cmark & \xmark & \cmark & 0.1816 & 0.1259 & 27.490 & 1.8130 \\
                  \cmark & \cmark & \xmark & \rr{0.1809} & \rr{0.1246} & \rr{26.106} & \rr{1.7798} \\
				\bottomrule
	        \end{tabular}
            }
        }\hspace{0mm}\vspace{0.5mm}
        \setlength{\tabcolsep}{0.5mm} 
        \subfloat[\small Ablation study of MFA. \label{tab:mfa}]{ \vspace{-1mm}
            \scalebox{0.98}{
                \begin{tabular}{SSS|c c c}
                    \toprule
                    \rowcolor{color3} w. LPIPS & w. DISTS & w. MSE & C-IQA$\uparrow$ & M-IQA$\uparrow$ & Deg.$\downarrow$ \\
                    \midrule
                    \xmark & \xmark & \xmark & 0.6585 & 0.6822 & 28.271  \\
                \xmark & \xmark & \cmark & 0.6533 & 0.6532 & 27.276 \\
                \xmark & \cmark & \xmark & 0.6582 & 0.6811 & 27.324 \\
                    \cmark & \xmark & \xmark & \rr{0.6673} & \rr{0.6858} & \rr{26.106} \\
                    \bottomrule
                \end{tabular}
                }
        }\hspace{1mm}\vspace{0mm}
        \subfloat[\small Complexity comparison during inference. \label{tab:time}]{\vspace{-1mm}
            \scalebox{0.98}{
                \begin{tabular}{l|c c c c c}
                    \toprule
                    \rowcolor{color3} Methods & PGDiff~\cite{yang2023pgdiff} & MGFR~\cite{tao2025overcoming} & OSEDiff~\cite{wu2024osediff} &  OSDFace~\cite{wang2025osdface} & HonestFace (ours) \\
                    \midrule
                    Step     & 1,000   & 50                & 1     & 1     & 1     \\
                    Time (s) & 85.81   & 6.9               & 0.13  & 0.10  & 0.13  \\
                    Param (M)& 176.4   & 2,029.3            & 1,302 & 978.4 & 1,059 \\
                        MACs (G) & 480,997 & $\approx$2,672    & 2,269 & 2,132 & 2,282 \\
                    \bottomrule
                \end{tabular}
            }
        }\hspace{0mm}\vspace{3mm}
	\caption{Comprehensive quantitative evaluations. (a) Quantitative comparisons with state-of-the-art reference-based methods on the real-world dataset. (b-e) More experiments for color shift evaluation, ablation studies, and inference complexity comparison. All ablation study experiments are conducted on CelebHQRef-Test. C-IQA stands for CLIP-IQA and M-IQA stands for MANIQA.}
	\label{tab:ablations}\vspace{-7mm}
\end{table*}

\vspace{1mm}
\noindent\textbf{Implementation Details}.
We adopt OSDFace~\cite{wang2025osdface} as our baseline model, which has demonstrated excellent performance in blind face restoration. For fine-tuning $\varepsilon_\theta$, \ie, the UNet, we utilize LoRA~\cite{hu2022lora} with both the rank and alpha set to $16$. The discriminator is based on LoRA-finetuned SDXL~\cite{podell2024sdxl} following OSDFace~\cite{wang2025osdface}. The training process spans 90K iterations across 4 NVIDIA A6000 GPUs, with a batch size of $4$. The prompt embedding consists of 77 tokens ($n$\,$=$\,$77$), where $n_1$\,$=$\,$4$, $n_2$\,$=$\,$7$, and $n_3$\,$=$\,$77$. For the face identity encoder, we use ArcFace~\cite{deng2019arcface} as $X_\text{FR}$. In terms of face alignment guidance, we leverage both ArcFace~\cite{deng2019arcface} and AdaFace~\cite{kim2022adaface} for feature extraction, and $k$\,$=$\,$10$ for MFA sharpening. As for loss functions, we use $L_\text{MFA}$ ($\lambda$$=$$4$), $L_\text{Per}$ ($\lambda$$=$$1$), $L_\text{ID}$ ($\lambda$$=$$2$), and $L_\mathcal{G}$ ($\lambda$$=$$0.005$) to balance magnitudes for losses.
The model is optimized by AdamW~\cite{loshchilov2018AdamW} with a learning rate of $1$$\times$$10^{-4}$ throughout the entire fine-tuning stage.

\vspace{-1.5mm}
\subsection{Main Results}
\vspace{-0.2mm}
To assess the performance in generating realistic facial textures and preserving identity consistency, we compare HonestFace with state-of-the-art reference-based face restoration methods, including ASFFNet~\cite{li2020asffnet512}, DMDNet~\cite{li2022dmdnet}, PGDiff~\cite{yang2023pgdiff}, FaceMe~\cite{liu2025faceme}, and MGFR~\cite{tao2025overcoming}. Additionally, to emphasize the texture honesty, we further evaluate transformer-based methods, such as CodeFormer~\cite{zhou2022codeformer} and DAEFR~\cite{tsai2024daefr}, as well as one-step diffusion methods like OSEDiff~\cite{wu2024osediff} and OSDFace~\cite{wang2025osdface}. It is worth noting that PGDiff~\cite{yang2023pgdiff} exhibits limitations when dealing with severe real-world degradation. Comparable issues are observed
in Fig.~3 of FaceMe~\cite{liu2025faceme} and Figs.~26, 27 of InterLCM~\cite{li2025interlcm} under similarly severe degradations.

\vspace{1mm}
\noindent\textbf{Quantitative Results}. \textcolor{red}{Red} and \textcolor{cvprblue}{blue} indicate the best and second-best results.
The comparisons with reference-based face restoration methods shown in Tabs.~\ref{tab:reference} and \ref{tab:real-world} show that HonestFace achieves state-of-the-art results across most metrics. In particular, identity-aware metrics such as Deg., LMD demonstrate HonestFace’s strong identity fidelity. Our method also excels on perceptual metrics.

A broader comparison with more methods is presented in Tab.~\ref{tab:RealWorld}. Since these competing methods operate in a reference-free manner, we focus the comparison solely on image quality. Pixel-level metrics demonstrate that our method outperforms other diffusion-based approaches, while higher perceptual-level scores indicate a superior restoration of textures and fine details.

\vspace{1mm}
\noindent\textbf{Qualitative Results}.
Visual comparisons in Figs.~\ref{fig:vis-reface},~\ref{fig:vis-realworld} show that HonestFace yields superior visual results. It avoids the over-smoothing artifacts seen in FaceMe~\cite{liu2025faceme}, and does not alter eye or skin color, unlike MGFR~\cite{tao2025overcoming} and OSDFace~\cite{wang2025osdface}. HonestFace effectively preserves both identity and expression, especially in key facial regions critical for recognition, such as the eyes, mouth, and skin wrinkles. For instance, in the fourth example of Fig.~\ref{fig:vis-realworld}, HonestFace accurately reconstructs a mole at the corner of the eye, while other methods either miss or distort this feature.

\vspace{-1.5mm}
\subsection{Complexity Analysis}
\vspace{-0.2mm}
Table~\ref{tab:time} provides a detailed comparison of model complexity, including sampling steps, inference time, parameter count, and multiply-accumulate operations (MACs). All tests are conducted on an NVIDIA A6000 GPU with 512$\times$512 inputs for consistency. Although HonestFace adds new modules to support multi-reference face restoration, it remains competitive with other one-step diffusion face restoration models. The additional IDE incurs only a marginal overhead
(7\% params, 6\% MACs) compared to baseline OSDFace, which is acceptable given the performance gains (about 10\%).

\vspace{-1.5mm}
\subsection{Color Shift} 
\vspace{-0.2mm}
Perceptual color difference remains a key challenge in image restoration. Due to the strong generative capability, diffusion-based methods tend to produce noticeable color inconsistencies, as clearly illustrated in Figs.~\ref{fig:vis-reface},~\ref{fig:vis-realworld}. 
We further evaluate color fidelity using {MS-SWD}~\cite{he2024ms-swd}, a color-aware metric. As shown in Tab.~\ref{tab:msswd}, experimental results demonstrate that our approach outperforms other methods, under both reference-based and no-reference settings.

\vspace{-1.5mm}
\subsection{Ablation Studies}
\vspace{-0.2mm}

\noindent\textbf{Identity Embedder (IDE)}.
To determine the optimal IDE structure and verify its effectiveness, we conduct a detailed ablation study in Tab.~\ref{tab:ide}. The quantitative results show that adding the facial feature extractor (FFE) improves eye feature consistency, while adding the face identity encoder (FIE) reduces angular differences in recognition embeddings. Combining FFE and FIE leads to deeper identity understanding and better fidelity consistency.

\vspace{1mm}
\noindent\textbf{Masked Face Alignment (MFA)}.
We evaluate the impact of heatmap weighting on texture and identity consistency, as shown in Tab.~\ref{tab:mfa}. We also explore different loss functions for $x_{\text{MFA},H}$ and $\hat{x}_{\text{MFA},H}$, and ultimately choose LPIPS as the local perceptual loss. It is also worth mentioning that using MSE as the loss causes over-smoothing in facial regions, making it unsuitable here.

\vspace{-1.5mm}
\section{Conclusion}
\vspace{-0.2mm}
In this paper, we address the critical challenges in ``honest'' face restoration: achieving natural, realistic textures while maintaining strict identity fidelity. We propose HonestFace, a novel one-step diffusion model that effectively integrates multiple reference images. 
HonestFace incorporates the identity embedder for robust identity preservation and masked face alignment to enhance perceptual naturalness. Furthermore, we develop a new real-world benchmark MultiRefCeleb-Test for comprehensive evaluation.
Our experiments demonstrate that HonestFace achieves state-of-the-art performance, delivering faces that are faithful to the subject's identity and exhibit realistic textures under challenging real-world degradations.

\section*{Acknowledgements}

This work is supported by the National Natural Science Foundation of China (62501386, 625B2116), CCF-Tencent Rhino-Bird Open Research Fund, and CAAI-Tencent Rhino-Bird Open Research Fund. This work is also sponsored by AI Hundred Schools Program and is carried out using the Ascend AI technology stack. 
This work is also supported by SJTU Shenlan Program Key Project SL2023ZD203.

\bibliographystyle{ACM-Reference-Format}
\balance
\bibliography{bib_conf}


\begin{thebibliography}{82}


\ifx \showCODEN    \undefined \def \showCODEN     #1{\unskip}     \fi
\ifx \showISBNx    \undefined \def \showISBNx     #1{\unskip}     \fi
\ifx \showISBNxiii \undefined \def \showISBNxiii  #1{\unskip}     \fi
\ifx \showISSN     \undefined \def \showISSN      #1{\unskip}     \fi
\ifx \showLCCN     \undefined \def \showLCCN      #1{\unskip}     \fi
\ifx \shownote     \undefined \def \shownote      #1{#1}          \fi
\ifx \showarticletitle \undefined \def \showarticletitle #1{#1}   \fi
\ifx \showURL      \undefined \def \showURL       {\relax}        \fi
\providecommand\bibfield[2]{#2}
\providecommand\bibinfo[2]{#2}
\providecommand\natexlab[1]{#1}
\providecommand\showeprint[2][]{arXiv:#2}

\bibitem[{Black Forest Labs}(2024)]%
        {flux2024}
\bibfield{author}{\bibinfo{person}{{Black Forest Labs}}.}
  \bibinfo{year}{2024}\natexlab{}.
\newblock \bibinfo{title}{FLUX}.
\newblock
  \bibinfo{howpublished}{\url{https://github.com/black-forest-labs/flux}}.
\newblock


\bibitem[Bulat and Tzimiropoulos(2018)]%
        {bulat2018super}
\bibfield{author}{\bibinfo{person}{Adrian Bulat} {and}
  \bibinfo{person}{Georgios Tzimiropoulos}.} \bibinfo{year}{2018}\natexlab{}.
\newblock \showarticletitle{{Super-fan}: {Integrated} facial landmark
  localization and super-resolution of real-world low resolution faces in
  arbitrary poses with {GANs}}. In \bibinfo{booktitle}{\emph{CVPR}}.
\newblock


\bibitem[Chan et~al\mbox{.}(2021)]%
        {chan2021glean}
\bibfield{author}{\bibinfo{person}{Kelvin~CK Chan}, \bibinfo{person}{Xintao
  Wang}, \bibinfo{person}{Xiangyu Xu}, \bibinfo{person}{Jinwei Gu}, {and}
  \bibinfo{person}{Chen~Change Loy}.} \bibinfo{year}{2021}\natexlab{}.
\newblock \showarticletitle{{GLEAN}: Generative Latent Bank for Large-Factor
  Image Super-Resolution}. In \bibinfo{booktitle}{\emph{CVPR}}.
\newblock


\bibitem[Chen et~al\mbox{.}(2021)]%
        {ChenPSFRGAN}
\bibfield{author}{\bibinfo{person}{Chaofeng Chen}, \bibinfo{person}{Xiaoming
  Li}, \bibinfo{person}{Yang Lingbo}, \bibinfo{person}{Xianhui Lin},
  \bibinfo{person}{Lei Zhang}, {and} \bibinfo{person}{Kwan-Yee~K. Wong}.}
  \bibinfo{year}{2021}\natexlab{}.
\newblock \showarticletitle{Progressive Semantic-Aware Style Transformation for
  Blind Face Restoration}. In \bibinfo{booktitle}{\emph{CVPR}}.
\newblock


\bibitem[Chen et~al\mbox{.}(2024)]%
        {chen2024pixartalpha}
\bibfield{author}{\bibinfo{person}{Junsong Chen}, \bibinfo{person}{Jincheng
  YU}, \bibinfo{person}{Chongjian GE}, \bibinfo{person}{Lewei Yao},
  \bibinfo{person}{Enze Xie}, \bibinfo{person}{Zhongdao Wang},
  \bibinfo{person}{James Kwok}, \bibinfo{person}{Ping Luo},
  \bibinfo{person}{Huchuan Lu}, {and} \bibinfo{person}{Zhenguo Li}.}
  \bibinfo{year}{2024}\natexlab{}.
\newblock \showarticletitle{PixArt-\${\textbackslash}alpha\$: Fast Training of
  Diffusion Transformer for Photorealistic Text-to-Image Synthesis}. In
  \bibinfo{booktitle}{\emph{ICLR}}.
\newblock


\bibitem[Chen et~al\mbox{.}(2023a)]%
        {chen2023BFRffusion}
\bibfield{author}{\bibinfo{person}{Xiaoxu Chen}, \bibinfo{person}{Jingfan Tan},
  \bibinfo{person}{Tao Wang}, \bibinfo{person}{Kaihao Zhang},
  \bibinfo{person}{Wenhan Luo}, {and} \bibinfo{person}{Xiaocun Cao}.}
  \bibinfo{year}{2023}\natexlab{a}.
\newblock \showarticletitle{Towards Real-World Blind Face Restoration with
  Generative Diffusion Prior}.
\newblock \bibinfo{journal}{\emph{arXiv preprint arXiv:2312.15736}}
  (\bibinfo{year}{2023}).
\newblock


\bibitem[Chen et~al\mbox{.}(2018)]%
        {chen2018fsrnet}
\bibfield{author}{\bibinfo{person}{Yu Chen}, \bibinfo{person}{Ying Tai},
  \bibinfo{person}{Xiaoming Liu}, \bibinfo{person}{Chunhua Shen}, {and}
  \bibinfo{person}{Jian Yang}.} \bibinfo{year}{2018}\natexlab{}.
\newblock \showarticletitle{{FSRNet}: End-to-End Learning Face Super-Resolution
  with Facial Priors}. In \bibinfo{booktitle}{\emph{CVPR}}.
\newblock


\bibitem[Chen et~al\mbox{.}(2025)]%
        {ntire2025face}
\bibfield{author}{\bibinfo{person}{Zheng Chen}, \bibinfo{person}{Jingkai Wang},
  \bibinfo{person}{Kai Liu}, \bibinfo{person}{Jue Gong}, \bibinfo{person}{Lei
  Sun}, \bibinfo{person}{Zongwei Wu}, \bibinfo{person}{Radu Timofte},
  \bibinfo{person}{Yulun Zhang}, {et~al\mbox{.}}}
  \bibinfo{year}{2025}\natexlab{}.
\newblock \showarticletitle{NTIRE 2025 Challenge on Real-World Face
  Restoration: Methods and Results}. In \bibinfo{booktitle}{\emph{CVPRW}}.
\newblock


\bibitem[Chen et~al\mbox{.}(2023b)]%
        {chen2023promptsr}
\bibfield{author}{\bibinfo{person}{Zheng Chen}, \bibinfo{person}{Yulun Zhang},
  \bibinfo{person}{Jinjin Gu}, \bibinfo{person}{Xin Yuan},
  \bibinfo{person}{Linghe Kong}, \bibinfo{person}{Guihai Chen}, {and}
  \bibinfo{person}{Xiaokang Yang}.} \bibinfo{year}{2023}\natexlab{b}.
\newblock \showarticletitle{Image Super-Resolution with Text Prompt Diffusion}.
\newblock \bibinfo{journal}{\emph{arXiv preprint arXiv:2303.06373}}
  (\bibinfo{year}{2023}).
\newblock


\bibitem[Deng et~al\mbox{.}(2019)]%
        {deng2019arcface}
\bibfield{author}{\bibinfo{person}{Jiankang Deng}, \bibinfo{person}{Jia Guo},
  \bibinfo{person}{Xue Niannan}, {and} \bibinfo{person}{Stefanos Zafeiriou}.}
  \bibinfo{year}{2019}\natexlab{}.
\newblock \showarticletitle{{ArcFace}: Additive Angular Margin Loss for Deep
  Face Recognition}. In \bibinfo{booktitle}{\emph{CVPR}}.
\newblock


\bibitem[Ding et~al\mbox{.}(2020)]%
        {ding2020dists}
\bibfield{author}{\bibinfo{person}{Keyan Ding}, \bibinfo{person}{Kede Ma},
  \bibinfo{person}{Shiqi Wang}, {and} \bibinfo{person}{Eero~P Simoncelli}.}
  \bibinfo{year}{2020}\natexlab{}.
\newblock \showarticletitle{Image Quality Assessment: Unifying Structure and
  Texture Similarity}.
\newblock \bibinfo{journal}{\emph{IEEE TPAMI}} (\bibinfo{year}{2020}).
\newblock


\bibitem[Ding et~al\mbox{.}(2024)]%
        {ding2024restoration}
\bibfield{author}{\bibinfo{person}{Zheng Ding}, \bibinfo{person}{Xuaner Zhang},
  \bibinfo{person}{Zhuowen Tu}, {and} \bibinfo{person}{Zhihao Xia}.}
  \bibinfo{year}{2024}\natexlab{}.
\newblock \showarticletitle{Restoration by generation with constrained priors}.
  In \bibinfo{booktitle}{\emph{CVPR}}.
\newblock


\bibitem[Dogan et~al\mbox{.}(2019)]%
        {dogan2019exemplar}
\bibfield{author}{\bibinfo{person}{Berk Dogan}, \bibinfo{person}{Shuhang Gu},
  {and} \bibinfo{person}{Radu Timofte}.} \bibinfo{year}{2019}\natexlab{}.
\newblock \showarticletitle{Exemplar guided face image super-resolution without
  facial landmarks}. In \bibinfo{booktitle}{\emph{CVPRW}}.
\newblock


\bibitem[Dong et~al\mbox{.}(2025)]%
        {dong2024tsdsr}
\bibfield{author}{\bibinfo{person}{Linwei Dong}, \bibinfo{person}{Qingnan Fan},
  \bibinfo{person}{Yihong Guo}, \bibinfo{person}{Zhonghao Wang},
  \bibinfo{person}{Qi Zhang}, \bibinfo{person}{Jinwei Chen},
  \bibinfo{person}{Yawei Luo}, {and} \bibinfo{person}{Changqing Zou}.}
  \bibinfo{year}{2025}\natexlab{}.
\newblock \showarticletitle{{TSD-SR}: One-Step Diffusion with Target Score
  Distillation for Real-World Image Super-Resolution}. In
  \bibinfo{booktitle}{\emph{CVPR}}.
\newblock


\bibitem[Gatys et~al\mbox{.}(2015)]%
        {Gatys2015Texture}
\bibfield{author}{\bibinfo{person}{Leon~A. Gatys},
  \bibinfo{person}{Alexander~S. Ecker}, {and} \bibinfo{person}{Matthias
  Bethge}.} \bibinfo{year}{2015}\natexlab{}.
\newblock \showarticletitle{Texture synthesis using convolutional neural
  networks}. In \bibinfo{booktitle}{\emph{NeurIPS}}.
\newblock


\bibitem[Gu et~al\mbox{.}(2020)]%
        {gu2020image}
\bibfield{author}{\bibinfo{person}{Jinjin Gu}, \bibinfo{person}{Yujun Shen},
  {and} \bibinfo{person}{Bolei Zhou}.} \bibinfo{year}{2020}\natexlab{}.
\newblock \showarticletitle{Image processing using multi-code gan prior}. In
  \bibinfo{booktitle}{\emph{CVPR}}.
\newblock


\bibitem[Gu et~al\mbox{.}(2022)]%
        {gu2022vqfr}
\bibfield{author}{\bibinfo{person}{Yuchao Gu}, \bibinfo{person}{Xintao Wang},
  \bibinfo{person}{Liangbin Xie}, \bibinfo{person}{Chao Dong},
  \bibinfo{person}{Gen Li}, \bibinfo{person}{Ying Shan}, {and}
  \bibinfo{person}{Ming-Ming Cheng}.} \bibinfo{year}{2022}\natexlab{}.
\newblock \showarticletitle{{VQFR}: Blind Face Restoration with
  Vector-Quantized Dictionary and Parallel Decoder}. In
  \bibinfo{booktitle}{\emph{ECCV}}.
\newblock


\bibitem[He et~al\mbox{.}(2024)]%
        {he2024ms-swd}
\bibfield{author}{\bibinfo{person}{Jiaqi He}, \bibinfo{person}{Zhihua Wang},
  \bibinfo{person}{Leon Wang}, \bibinfo{person}{Tsein-I Liu},
  \bibinfo{person}{Yuming Fang}, \bibinfo{person}{Qilin Sun}, {and}
  \bibinfo{person}{Kede Ma}.} \bibinfo{year}{2024}\natexlab{}.
\newblock \showarticletitle{Multiscale Sliced {Wasserstein} Distances as
  Perceptual Color Difference Measures}. In \bibinfo{booktitle}{\emph{ECCV}}.
\newblock


\bibitem[Hsiao et~al\mbox{.}(2024)]%
        {hsiao2024refldm}
\bibfield{author}{\bibinfo{person}{Chi-Wei Hsiao}, \bibinfo{person}{Yu-Lun
  Liu}, \bibinfo{person}{Cheng-Kun Yang}, \bibinfo{person}{Sheng-Po Kuo},
  \bibinfo{person}{Kevin Jou}, {and} \bibinfo{person}{Chia-Ping Chen}.}
  \bibinfo{year}{2024}\natexlab{}.
\newblock \showarticletitle{ReF-LDM: A Latent Diffusion Model for
  Reference-based Face Image Restoration}. In
  \bibinfo{booktitle}{\emph{NeurIPS}}.
\newblock


\bibitem[Hu et~al\mbox{.}(2022)]%
        {hu2022lora}
\bibfield{author}{\bibinfo{person}{Edward~J Hu}, \bibinfo{person}{Yelong Shen},
  \bibinfo{person}{Phillip Wallis}, \bibinfo{person}{Zeyuan Allen-Zhu},
  \bibinfo{person}{Yuanzhi Li}, \bibinfo{person}{Shean Wang},
  \bibinfo{person}{Lu Wang}, {and} \bibinfo{person}{Weizhu Chen}.}
  \bibinfo{year}{2022}\natexlab{}.
\newblock \showarticletitle{Lo{RA}: Low-Rank Adaptation of Large Language
  Models}. In \bibinfo{booktitle}{\emph{ICLR}}.
\newblock


\bibitem[Karras et~al\mbox{.}(2019)]%
        {karras2019ffhq}
\bibfield{author}{\bibinfo{person}{Tero Karras}, \bibinfo{person}{Samuli
  Laine}, {and} \bibinfo{person}{Timo Aila}.} \bibinfo{year}{2019}\natexlab{}.
\newblock \showarticletitle{A style-based generator architecture for generative
  adversarial networks}. In \bibinfo{booktitle}{\emph{CVPR}}.
\newblock


\bibitem[Ke et~al\mbox{.}(2021)]%
        {ke2021musiq}
\bibfield{author}{\bibinfo{person}{Junjie Ke}, \bibinfo{person}{Qifei Wang},
  \bibinfo{person}{Yilin Wang}, \bibinfo{person}{Peyman Milanfar}, {and}
  \bibinfo{person}{Feng Yang}.} \bibinfo{year}{2021}\natexlab{}.
\newblock \showarticletitle{{ MUSIQ: Multi-scale Image Quality Transformer }}.
  In \bibinfo{booktitle}{\emph{ICCV}}.
\newblock


\bibitem[Kim et~al\mbox{.}(2019)]%
        {kim2019progressive}
\bibfield{author}{\bibinfo{person}{Deokyun Kim}, \bibinfo{person}{Minseon Kim},
  \bibinfo{person}{Gihyun Kwon}, {and} \bibinfo{person}{Dae-Shik Kim}.}
  \bibinfo{year}{2019}\natexlab{}.
\newblock \showarticletitle{Progressive face super-resolution via attention to
  facial landmark}. In \bibinfo{booktitle}{\emph{BMVC}}.
\newblock


\bibitem[Kim et~al\mbox{.}(2022)]%
        {kim2022adaface}
\bibfield{author}{\bibinfo{person}{Minchul Kim}, \bibinfo{person}{Anil~K Jain},
  {and} \bibinfo{person}{Xiaoming Liu}.} \bibinfo{year}{2022}\natexlab{}.
\newblock \showarticletitle{{AdaFace}: Quality Adaptive Margin for Face
  Recognition}. In \bibinfo{booktitle}{\emph{CVPR}}.
\newblock


\bibitem[Li et~al\mbox{.}(2025a)]%
        {li2025fluxsr}
\bibfield{author}{\bibinfo{person}{Jianze Li}, \bibinfo{person}{Jiezhang Cao},
  \bibinfo{person}{Yong Guo}, \bibinfo{person}{Wenbo Li}, {and}
  \bibinfo{person}{Yulun Zhang}.} \bibinfo{year}{2025}\natexlab{a}.
\newblock \showarticletitle{One Diffusion Step to Real-World Super-Resolution
  via Flow Trajectory Distillation}. In \bibinfo{booktitle}{\emph{ICML}}.
\newblock


\bibitem[Li et~al\mbox{.}(2025b)]%
        {li2024dfosd}
\bibfield{author}{\bibinfo{person}{Jianze Li}, \bibinfo{person}{Jiezhang Cao},
  \bibinfo{person}{Zichen Zou}, \bibinfo{person}{Xiongfei Su},
  \bibinfo{person}{Xin Yuan}, \bibinfo{person}{Yulun Zhang},
  \bibinfo{person}{Yong Guo}, {and} \bibinfo{person}{Xiaokang Yang}.}
  \bibinfo{year}{2025}\natexlab{b}.
\newblock \showarticletitle{Unleashing the Power of One-Step Diffusion based
  Image Super-Resolution via a Large-Scale Diffusion Discriminator}.
\newblock \bibinfo{journal}{\emph{NeurIPS}} (\bibinfo{year}{2025}).
\newblock


\bibitem[Li et~al\mbox{.}(2025c)]%
        {li2025interlcm}
\bibfield{author}{\bibinfo{person}{Senmao Li}, \bibinfo{person}{Kai Wang},
  \bibinfo{person}{Joost van~de Weijer}, \bibinfo{person}{Fahad~Shahbaz Khan},
  \bibinfo{person}{Chun-Le Guo}, \bibinfo{person}{Shiqi Yang},
  \bibinfo{person}{Yaxing Wang}, \bibinfo{person}{Jian Yang}, {and}
  \bibinfo{person}{Ming-Ming Cheng}.} \bibinfo{year}{2025}\natexlab{c}.
\newblock \showarticletitle{InterLCM: Low-Quality Images as Intermediate States
  of Latent Consistency Models for Effective Blind Face Restoration}. In
  \bibinfo{booktitle}{\emph{ICLR}}.
\newblock


\bibitem[Li et~al\mbox{.}(2020a)]%
        {li2020blind}
\bibfield{author}{\bibinfo{person}{Xiaoming Li}, \bibinfo{person}{Chaofeng
  Chen}, \bibinfo{person}{Shangchen Zhou}, \bibinfo{person}{Xianhui Lin},
  \bibinfo{person}{Wangmeng Zuo}, {and} \bibinfo{person}{Lei Zhang}.}
  \bibinfo{year}{2020}\natexlab{a}.
\newblock \showarticletitle{Blind face restoration via deep multi-scale
  component dictionaries}. In \bibinfo{booktitle}{\emph{ECCV}}.
\newblock


\bibitem[Li et~al\mbox{.}(2020b)]%
        {li2020asffnet512}
\bibfield{author}{\bibinfo{person}{Xiaoming Li}, \bibinfo{person}{Wenyu Li},
  \bibinfo{person}{Dongwei Ren}, \bibinfo{person}{Hongzhi Zhang},
  \bibinfo{person}{Meng Wang}, {and} \bibinfo{person}{Wangmeng Zuo}.}
  \bibinfo{year}{2020}\natexlab{b}.
\newblock \showarticletitle{Enhanced blind face restoration with multi-exemplar
  images and adaptive spatial feature fusion}. In
  \bibinfo{booktitle}{\emph{CVPR}}.
\newblock


\bibitem[Li et~al\mbox{.}(2018a)]%
        {li2018gfrnet}
\bibfield{author}{\bibinfo{person}{Xiaoming Li}, \bibinfo{person}{Ming Liu},
  \bibinfo{person}{Yuting Ye}, \bibinfo{person}{Wangmeng Zuo},
  \bibinfo{person}{Liang Lin}, {and} \bibinfo{person}{Ruigang Yang}.}
  \bibinfo{year}{2018}\natexlab{a}.
\newblock \showarticletitle{Learning Warped Guidance for Blind Face
  Restoration}. In \bibinfo{booktitle}{\emph{ECCV}}.
\newblock


\bibitem[Li et~al\mbox{.}(2018b)]%
        {li2018learning}
\bibfield{author}{\bibinfo{person}{Xiaoming Li}, \bibinfo{person}{Ming Liu},
  \bibinfo{person}{Yuting Ye}, \bibinfo{person}{Wangmeng Zuo},
  \bibinfo{person}{Liang Lin}, {and} \bibinfo{person}{Ruigang Yang}.}
  \bibinfo{year}{2018}\natexlab{b}.
\newblock \showarticletitle{Learning warped guidance for blind face
  restoration}. In \bibinfo{booktitle}{\emph{ECCV}}.
\newblock


\bibitem[Li et~al\mbox{.}(2022)]%
        {li2022dmdnet}
\bibfield{author}{\bibinfo{person}{Xiaoming Li}, \bibinfo{person}{Shiguang
  Zhang}, \bibinfo{person}{Shangchen Zhou}, \bibinfo{person}{Lei Zhang}, {and}
  \bibinfo{person}{Wangmeng Zuo}.} \bibinfo{year}{2022}\natexlab{}.
\newblock \showarticletitle{Learning dual memory dictionaries for blind face
  restoration}.
\newblock \bibinfo{journal}{\emph{IEEE TPAMI}} (\bibinfo{year}{2022}).
\newblock


\bibitem[Lin et~al\mbox{.}(2024)]%
        {lin2024diffbir}
\bibfield{author}{\bibinfo{person}{Xinqi Lin}, \bibinfo{person}{Jingwen He},
  \bibinfo{person}{Ziyan Chen}, \bibinfo{person}{Zhaoyang Lyu},
  \bibinfo{person}{Bo Dai}, \bibinfo{person}{Fanghua Yu},
  \bibinfo{person}{Wanli Ouyang}, \bibinfo{person}{Yu Qiao}, {and}
  \bibinfo{person}{Chao Dong}.} \bibinfo{year}{2024}\natexlab{}.
\newblock \showarticletitle{{DiffBIR}: Towards Blind Image Restoration with
  Generative Diffusion Prior}. In \bibinfo{booktitle}{\emph{ECCV}}.
\newblock


\bibitem[Liu et~al\mbox{.}(2025)]%
        {liu2025faceme}
\bibfield{author}{\bibinfo{person}{Siyu Liu}, \bibinfo{person}{Zheng-Peng
  Duan}, \bibinfo{person}{Jia OuYang}, \bibinfo{person}{Jiayi Fu},
  \bibinfo{person}{Hyunhee Park}, \bibinfo{person}{Zikun Liu},
  \bibinfo{person}{Chun-Le Guo}, {and} \bibinfo{person}{Chongyi Li}.}
  \bibinfo{year}{2025}\natexlab{}.
\newblock \showarticletitle{{FaceMe}: Robust Blind Face Restoration with
  Personal Identification}. In \bibinfo{booktitle}{\emph{AAAI}}.
\newblock


\bibitem[Loshchilov and Hutter(2019)]%
        {loshchilov2018AdamW}
\bibfield{author}{\bibinfo{person}{Ilya Loshchilov} {and}
  \bibinfo{person}{Frank Hutter}.} \bibinfo{year}{2019}\natexlab{}.
\newblock \showarticletitle{Decoupled Weight Decay Regularization}. In
  \bibinfo{booktitle}{\emph{ICLR}}.
\newblock


\bibitem[Luo et~al\mbox{.}(2023)]%
        {luo2023diffinstruct}
\bibfield{author}{\bibinfo{person}{Weijian Luo}, \bibinfo{person}{Tianyang Hu},
  \bibinfo{person}{Shifeng Zhang}, \bibinfo{person}{Jiacheng Sun},
  \bibinfo{person}{Zhenguo Li}, {and} \bibinfo{person}{Zhihua Zhang}.}
  \bibinfo{year}{2023}\natexlab{}.
\newblock \showarticletitle{Diff-instruct: A universal approach for
  transferring knowledge from pre-trained diffusion models}. In
  \bibinfo{booktitle}{\emph{NeurIPS}}.
\newblock


\bibitem[Mahendran and Vedaldi(2015)]%
        {Mahendran2015Understanding}
\bibfield{author}{\bibinfo{person}{Aravindh Mahendran} {and}
  \bibinfo{person}{Andrea Vedaldi}.} \bibinfo{year}{2015}\natexlab{}.
\newblock \showarticletitle{{ Understanding deep image representations by
  inverting them }}. In \bibinfo{booktitle}{\emph{CVPR}}.
\newblock


\bibitem[McCouat and Voiculescu(2022)]%
        {mccouat2022contour}
\bibfield{author}{\bibinfo{person}{James McCouat} {and} \bibinfo{person}{Irina
  Voiculescu}.} \bibinfo{year}{2022}\natexlab{}.
\newblock \showarticletitle{Contour-hugging heatmaps for landmark detection}.
  In \bibinfo{booktitle}{\emph{CVPR}}.
\newblock


\bibitem[Menon et~al\mbox{.}(2020)]%
        {menon2020pulse}
\bibfield{author}{\bibinfo{person}{Sachit Menon}, \bibinfo{person}{Alexandru
  Damian}, \bibinfo{person}{Shijia Hu}, \bibinfo{person}{Nikhil Ravi}, {and}
  \bibinfo{person}{Cynthia Rudin}.} \bibinfo{year}{2020}\natexlab{}.
\newblock \showarticletitle{Pulse: Self-supervised photo upsampling via latent
  space exploration of generative models}. In \bibinfo{booktitle}{\emph{CVPR}}.
\newblock


\bibitem[Nitzan et~al\mbox{.}(2022)]%
        {nitzan2022mystyle}
\bibfield{author}{\bibinfo{person}{Yotam Nitzan}, \bibinfo{person}{Kfir
  Aberman}, \bibinfo{person}{Qiurui He}, \bibinfo{person}{Orly Liba},
  \bibinfo{person}{Michal Yarom}, \bibinfo{person}{Yossi Gandelsman},
  \bibinfo{person}{Inbar Mosseri}, \bibinfo{person}{Yael Pritch}, {and}
  \bibinfo{person}{Daniel Cohen-Or}.} \bibinfo{year}{2022}\natexlab{}.
\newblock \showarticletitle{Mystyle: A personalized generative prior}.
\newblock \bibinfo{journal}{\emph{ACM TOG}} (\bibinfo{year}{2022}).
\newblock


\bibitem[Payer et~al\mbox{.}(2016)]%
        {Payer2016}
\bibfield{author}{\bibinfo{person}{Christian Payer}, \bibinfo{person}{Darko
  {\v{S}}tern}, \bibinfo{person}{Horst Bischof}, {and} \bibinfo{person}{Martin
  Urschler}.} \bibinfo{year}{2016}\natexlab{}.
\newblock \showarticletitle{Regressing Heatmaps for Multiple Landmark
  Localization Using {CNNs}}. In \bibinfo{booktitle}{\emph{MICCAI}}.
\newblock


\bibitem[Podell et~al\mbox{.}(2024)]%
        {podell2024sdxl}
\bibfield{author}{\bibinfo{person}{Dustin Podell}, \bibinfo{person}{Zion
  English}, \bibinfo{person}{Kyle Lacey}, \bibinfo{person}{Andreas Blattmann},
  \bibinfo{person}{Tim Dockhorn}, \bibinfo{person}{Jonas M{\"u}ller},
  \bibinfo{person}{Joe Penna}, {and} \bibinfo{person}{Robin Rombach}.}
  \bibinfo{year}{2024}\natexlab{}.
\newblock \showarticletitle{{SDXL}: Improving Latent Diffusion Models for
  High-Resolution Image Synthesis}. In \bibinfo{booktitle}{\emph{ICLR}}.
\newblock


\bibitem[Qiu et~al\mbox{.}(2023)]%
        {qiu2023diffbfr}
\bibfield{author}{\bibinfo{person}{Xinmin Qiu}, \bibinfo{person}{Congying Han},
  \bibinfo{person}{Zicheng Zhang}, \bibinfo{person}{Bonan Li},
  \bibinfo{person}{Tiande Guo}, {and} \bibinfo{person}{Xuecheng Nie}.}
  \bibinfo{year}{2023}\natexlab{}.
\newblock \showarticletitle{{DiffBFR}: Bootstrapping Diffusion Model for Blind
  Face Restoration}. In \bibinfo{booktitle}{\emph{ACM MM}}.
\newblock


\bibitem[Ramesh et~al\mbox{.}(2021)]%
        {pmlr-v139-ramesh21a}
\bibfield{author}{\bibinfo{person}{Aditya Ramesh}, \bibinfo{person}{Mikhail
  Pavlov}, \bibinfo{person}{Gabriel Goh}, \bibinfo{person}{Scott Gray},
  \bibinfo{person}{Chelsea Voss}, \bibinfo{person}{Alec Radford},
  \bibinfo{person}{Mark Chen}, {and} \bibinfo{person}{Ilya Sutskever}.}
  \bibinfo{year}{2021}\natexlab{}.
\newblock \showarticletitle{Zero-Shot Text-to-Image Generation}. In
  \bibinfo{booktitle}{\emph{ICML}}.
\newblock


\bibitem[Rombach et~al\mbox{.}(2022)]%
        {rombach2022ldm}
\bibfield{author}{\bibinfo{person}{Robin Rombach}, \bibinfo{person}{Andreas
  Blattmann}, \bibinfo{person}{Dominik Lorenz}, \bibinfo{person}{Patrick
  Esser}, {and} \bibinfo{person}{Bjorn Ommer}.}
  \bibinfo{year}{2022}\natexlab{}.
\newblock \showarticletitle{{ High-Resolution Image Synthesis with Latent
  Diffusion Models }}. In \bibinfo{booktitle}{\emph{CVPR}}.
\newblock


\bibitem[Sauer et~al\mbox{.}(2024a)]%
        {Sauer2024LADD}
\bibfield{author}{\bibinfo{person}{Axel Sauer}, \bibinfo{person}{Frederic
  Boesel}, \bibinfo{person}{Tim Dockhorn}, \bibinfo{person}{Andreas Blattmann},
  \bibinfo{person}{Patrick Esser}, {and} \bibinfo{person}{Robin Rombach}.}
  \bibinfo{year}{2024}\natexlab{a}.
\newblock \showarticletitle{Fast High-Resolution Image Synthesis with Latent
  Adversarial Diffusion Distillation}. In \bibinfo{booktitle}{\emph{ACM
  SIGGRAPH Asia}}.
\newblock


\bibitem[Sauer et~al\mbox{.}(2024b)]%
        {Sauer2023ADD}
\bibfield{author}{\bibinfo{person}{Axel Sauer}, \bibinfo{person}{Dominik
  Lorenz}, \bibinfo{person}{Andreas Blattmann}, {and} \bibinfo{person}{Robin
  Rombach}.} \bibinfo{year}{2024}\natexlab{b}.
\newblock \showarticletitle{Adversarial Diffusion Distillation}. In
  \bibinfo{booktitle}{\emph{ECCV}}.
\newblock


\bibitem[Shen et~al\mbox{.}(2018)]%
        {shen2018deep}
\bibfield{author}{\bibinfo{person}{Ziyi Shen}, \bibinfo{person}{Wei-Sheng Lai},
  \bibinfo{person}{Tingfa Xu}, \bibinfo{person}{Jan Kautz}, {and}
  \bibinfo{person}{Ming-Hsuan Yang}.} \bibinfo{year}{2018}\natexlab{}.
\newblock \showarticletitle{Deep semantic face deblurring}. In
  \bibinfo{booktitle}{\emph{CVPR}}.
\newblock


\bibitem[Simonyan and Zisserman(2015)]%
        {simonyan2015vgg}
\bibfield{author}{\bibinfo{person}{K Simonyan} {and} \bibinfo{person}{A
  Zisserman}.} \bibinfo{year}{2015}\natexlab{}.
\newblock \showarticletitle{Very Deep Convolutional Networks for Large-Scale
  Image Recognition}. In \bibinfo{booktitle}{\emph{ICLR}}.
\newblock


\bibitem[Suin and Chellappa(2024)]%
        {Suin2024CLRFace}
\bibfield{author}{\bibinfo{person}{Maitreya Suin} {and} \bibinfo{person}{Rama
  Chellappa}.} \bibinfo{year}{2024}\natexlab{}.
\newblock \showarticletitle{{CLR-Face}: Conditional Latent Refinement for Blind
  Face Restoration Using Score-Based Diffusion Models}. In
  \bibinfo{booktitle}{\emph{IJCAI}}.
\newblock


\bibitem[Sun et~al\mbox{.}(2025)]%
        {sun2024pisasr}
\bibfield{author}{\bibinfo{person}{Lingchen Sun}, \bibinfo{person}{Rongyuan
  Wu}, \bibinfo{person}{Zhiyuan Ma}, \bibinfo{person}{Shuaizheng Liu},
  \bibinfo{person}{Qiaosi Yi}, {and} \bibinfo{person}{Lei Zhang}.}
  \bibinfo{year}{2025}\natexlab{}.
\newblock \showarticletitle{Pixel-level and Semantic-level Adjustable
  Super-resolution: A Dual-LoRA Approach}. In \bibinfo{booktitle}{\emph{CVPR}}.
\newblock


\bibitem[Tao et~al\mbox{.}(2025)]%
        {tao2025overcoming}
\bibfield{author}{\bibinfo{person}{Keda Tao}, \bibinfo{person}{Jinjin Gu},
  \bibinfo{person}{Yulun Zhang}, \bibinfo{person}{Xiucheng Wang}, {and}
  \bibinfo{person}{Nan Cheng}.} \bibinfo{year}{2025}\natexlab{}.
\newblock \showarticletitle{Overcoming False Illusions in Real-World Face
  Restoration with Multi-Modal Guided Diffusion Model}. In
  \bibinfo{booktitle}{\emph{ICLR}}.
\newblock


\bibitem[Tsai et~al\mbox{.}(2024)]%
        {tsai2024daefr}
\bibfield{author}{\bibinfo{person}{Yu-Ju Tsai}, \bibinfo{person}{Yu-Lun Liu},
  \bibinfo{person}{Lu Qi}, \bibinfo{person}{Kelvin~CK Chan}, {and}
  \bibinfo{person}{Ming-Hsuan Yang}.} \bibinfo{year}{2024}\natexlab{}.
\newblock \showarticletitle{Dual Associated Encoder for Face Restoration}. In
  \bibinfo{booktitle}{\emph{ICLR}}.
\newblock


\bibitem[Varanka et~al\mbox{.}(2024)]%
        {varanka2024pfstorer}
\bibfield{author}{\bibinfo{person}{Tuomas Varanka}, \bibinfo{person}{Tapani
  Toivonen}, \bibinfo{person}{Soumya Tripathy}, \bibinfo{person}{Guoying Zhao},
  {and} \bibinfo{person}{Erman Acar}.} \bibinfo{year}{2024}\natexlab{}.
\newblock \showarticletitle{{PFStorer}: Personalized face restoration and
  super-resolution}. In \bibinfo{booktitle}{\emph{CVPR}}.
\newblock


\bibitem[Wan et~al\mbox{.}(2020)]%
        {wan2020bringing}
\bibfield{author}{\bibinfo{person}{Ziyu Wan}, \bibinfo{person}{Bo Zhang},
  \bibinfo{person}{Dongdong Chen}, \bibinfo{person}{Pan Zhang},
  \bibinfo{person}{Dong Chen}, \bibinfo{person}{Jing Liao}, {and}
  \bibinfo{person}{Fang Wen}.} \bibinfo{year}{2020}\natexlab{}.
\newblock \showarticletitle{Bringing old photos back to life}. In
  \bibinfo{booktitle}{\emph{CVPR}}.
\newblock


\bibitem[Wang et~al\mbox{.}(2023a)]%
        {wang2022clipiqa}
\bibfield{author}{\bibinfo{person}{Jianyi Wang}, \bibinfo{person}{Kelvin~CK
  Chan}, {and} \bibinfo{person}{Chen~Change Loy}.}
  \bibinfo{year}{2023}\natexlab{a}.
\newblock \showarticletitle{Exploring CLIP for Assessing the Look and Feel of
  Images}. In \bibinfo{booktitle}{\emph{AAAI}}.
\newblock


\bibitem[Wang et~al\mbox{.}(2026a)]%
        {ntire2026face}
\bibfield{author}{\bibinfo{person}{Jingkai Wang}, \bibinfo{person}{Jue Gong},
  \bibinfo{person}{Zheng Chen}, \bibinfo{person}{Kai Liu},
  \bibinfo{person}{Jiatong Li}, \bibinfo{person}{Yulun Zhang},
  \bibinfo{person}{Radu Timofte}, \bibinfo{person}{Jiachen Tu},
  \bibinfo{person}{Yaokun Shi}, \bibinfo{person}{Guoyi Xu},
  \bibinfo{person}{Yaoxin Jiang}, \bibinfo{person}{Jiajia Liu},
  \bibinfo{person}{Yingsi Chen}, \bibinfo{person}{Yijiao Liu},
  \bibinfo{person}{Hui Li}, \bibinfo{person}{Yu Wang},
  \bibinfo{person}{Congchao Zhu}, \bibinfo{person}{Alexandru-Gabriel
  Lefterache}, \bibinfo{person}{Anamaria Radoi}, \bibinfo{person}{Chuanyue
  Yan}, \bibinfo{person}{Tao Lu}, \bibinfo{person}{Yanduo Zhang},
  \bibinfo{person}{Kanghui Zhao}, \bibinfo{person}{Jiaming Wang},
  \bibinfo{person}{Yuqi Li}, \bibinfo{person}{WenBo Xiong},
  \bibinfo{person}{Yifei Chen}, \bibinfo{person}{Xian Hu}, \bibinfo{person}{Wei
  Deng}, \bibinfo{person}{Daiguo Zhou}, \bibinfo{person}{Sujith~Roy V},
  \bibinfo{person}{Claudia Jesuraj}, \bibinfo{person}{Vikas B},
  \bibinfo{person}{Spoorthi LC}, \bibinfo{person}{Nikhil Akalwadi},
  \bibinfo{person}{Ramesh~Ashok Tabib}, \bibinfo{person}{Uma Mudenagudi},
  \bibinfo{person}{Yuxuan Jiang}, \bibinfo{person}{Chengxi Zeng},
  \bibinfo{person}{Tianhao Peng}, \bibinfo{person}{Fan Zhang},
  \bibinfo{person}{David Bull}, \bibinfo{person}{Wei Zhou},
  \bibinfo{person}{Linfeng Li}, \bibinfo{person}{Hongyu Huang},
  \bibinfo{person}{Hoyoung Lee}, \bibinfo{person}{SangYun Oh},
  \bibinfo{person}{ChangYoung Jeong}, \bibinfo{person}{Axi Niu},
  \bibinfo{person}{Jinyang Zhang}, \bibinfo{person}{Zhenguo Wu},
  \bibinfo{person}{Senyan Qing}, \bibinfo{person}{Jinqiu Sun}, {and}
  \bibinfo{person}{Yanning Zhang}.} \bibinfo{year}{2026}\natexlab{a}.
\newblock \showarticletitle{The Second Challenge on Real-World Face Restoration
  at NTIRE 2026: Methods and Results}. In \bibinfo{booktitle}{\emph{CVPRW}}.
\newblock


\bibitem[Wang et~al\mbox{.}(2025)]%
        {wang2025osdface}
\bibfield{author}{\bibinfo{person}{Jingkai Wang}, \bibinfo{person}{Jue Gong},
  \bibinfo{person}{Lin Zhang}, \bibinfo{person}{Zheng Chen},
  \bibinfo{person}{Xing Liu}, \bibinfo{person}{Hong Gu},
  \bibinfo{person}{Yutong Liu}, \bibinfo{person}{Yulun Zhang}, {and}
  \bibinfo{person}{Xiaokang Yang}.} \bibinfo{year}{2025}\natexlab{}.
\newblock \showarticletitle{{OSDFace}: One-Step Diffusion Model for Face
  Restoration}. In \bibinfo{booktitle}{\emph{CVPR}}.
\newblock


\bibitem[Wang et~al\mbox{.}(2026b)]%
        {wang2026strsr}
\bibfield{author}{\bibinfo{person}{Jingkai Wang}, \bibinfo{person}{Yixin Tang},
  \bibinfo{person}{Jue Gong}, \bibinfo{person}{Jiatong Li},
  \bibinfo{person}{Shu Li}, \bibinfo{person}{Libo Liu},
  \bibinfo{person}{Jianliang Lan}, \bibinfo{person}{Yutong Liu}, {and}
  \bibinfo{person}{Yulun Zhang}.} \bibinfo{year}{2026}\natexlab{b}.
\newblock \showarticletitle{Spectral and Trajectory Regularization for
  Diffusion Transformer Super-Resolution}. In \bibinfo{booktitle}{\emph{ECCV}}.
\newblock


\bibitem[Wang et~al\mbox{.}(2024)]%
        {wang2024stablesr}
\bibfield{author}{\bibinfo{person}{Jianyi Wang}, \bibinfo{person}{Zongsheng
  Yue}, \bibinfo{person}{Shangchen Zhou}, \bibinfo{person}{Kelvin~C.K. Chan},
  {and} \bibinfo{person}{Chen~Change Loy}.} \bibinfo{year}{2024}\natexlab{}.
\newblock \showarticletitle{Exploiting Diffusion Prior for Real-World Image
  Super-Resolution}.
\newblock \bibinfo{journal}{\emph{IJCV}} (\bibinfo{year}{2024}).
\newblock


\bibitem[Wang et~al\mbox{.}(2019)]%
        {wang2019adaptive}
\bibfield{author}{\bibinfo{person}{Xinyao Wang}, \bibinfo{person}{Liefeng Bo},
  {and} \bibinfo{person}{Li Fuxin}.} \bibinfo{year}{2019}\natexlab{}.
\newblock \showarticletitle{Adaptive wing loss for robust face alignment via
  heatmap regression}. In \bibinfo{booktitle}{\emph{ICCV}}.
\newblock


\bibitem[Wang et~al\mbox{.}(2021)]%
        {wang2021gfpgan}
\bibfield{author}{\bibinfo{person}{Xintao Wang}, \bibinfo{person}{Yu Li},
  \bibinfo{person}{Honglun Zhang}, {and} \bibinfo{person}{Ying Shan}.}
  \bibinfo{year}{2021}\natexlab{}.
\newblock \showarticletitle{Towards Real-World Blind Face Restoration with
  Generative Facial Prior}. In \bibinfo{booktitle}{\emph{CVPR}}.
\newblock


\bibitem[Wang et~al\mbox{.}(2023b)]%
        {wang2023prolificdreamer}
\bibfield{author}{\bibinfo{person}{Zhengyi Wang}, \bibinfo{person}{Cheng Lu},
  \bibinfo{person}{Yikai Wang}, \bibinfo{person}{Fan Bao},
  \bibinfo{person}{Chongxuan Li}, \bibinfo{person}{Hang Su}, {and}
  \bibinfo{person}{Jun Zhu}.} \bibinfo{year}{2023}\natexlab{b}.
\newblock \showarticletitle{ProlificDreamer: High-Fidelity and Diverse
  Text-to-3D Generation with Variational Score Distillation}. In
  \bibinfo{booktitle}{\emph{NeurIPS}}.
\newblock


\bibitem[Wang et~al\mbox{.}(2023c)]%
        {wang2023restoreformer++}
\bibfield{author}{\bibinfo{person}{Zhouxia Wang}, \bibinfo{person}{Jiawei
  Zhang}, \bibinfo{person}{Tianshui Chen}, \bibinfo{person}{Wenping Wang},
  {and} \bibinfo{person}{Ping Luo}.} \bibinfo{year}{2023}\natexlab{c}.
\newblock \showarticletitle{RestoreFormer++: Towards Real-World Blind Face
  Restoration from Undegraded Key-Value Pairs}.
\newblock \bibinfo{journal}{\emph{IEEE TPAMI}} (\bibinfo{year}{2023}).
\newblock


\bibitem[Wang et~al\mbox{.}(2023d)]%
        {wang2023dr2}
\bibfield{author}{\bibinfo{person}{Zhixin Wang}, \bibinfo{person}{Xiaoyun
  Zhang}, \bibinfo{person}{Ziying Zhang}, \bibinfo{person}{Huangjie Zheng},
  \bibinfo{person}{Mingyuan Zhou}, \bibinfo{person}{Ya Zhang}, {and}
  \bibinfo{person}{Yanfeng Wang}.} \bibinfo{year}{2023}\natexlab{d}.
\newblock \showarticletitle{DR2: Diffusion-based Robust Degradation Remover for
  Blind Face Restoration}. In \bibinfo{booktitle}{\emph{CVPR}}.
\newblock


\bibitem[Wu et~al\mbox{.}(2024a)]%
        {wu2024osediff}
\bibfield{author}{\bibinfo{person}{Rongyuan Wu}, \bibinfo{person}{Lingchen
  Sun}, \bibinfo{person}{Zhiyuan Ma}, {and} \bibinfo{person}{Lei Zhang}.}
  \bibinfo{year}{2024}\natexlab{a}.
\newblock \showarticletitle{One-Step Effective Diffusion Network for Real-World
  Image Super-Resolution}. In \bibinfo{booktitle}{\emph{NeurIPS}}.
\newblock


\bibitem[Wu et~al\mbox{.}(2024b)]%
        {wu2024seesr}
\bibfield{author}{\bibinfo{person}{Rongyuan Wu}, \bibinfo{person}{Tao Yang},
  \bibinfo{person}{Lingchen Sun}, \bibinfo{person}{Zhengqiang Zhang},
  \bibinfo{person}{Shuai Li}, {and} \bibinfo{person}{Lei Zhang}.}
  \bibinfo{year}{2024}\natexlab{b}.
\newblock \showarticletitle{{SeeSR}: Towards Semantics-Aware Real-World Image
  Super-Resolution}. In \bibinfo{booktitle}{\emph{CVPR}}.
\newblock


\bibitem[Wu et~al\mbox{.}(2025)]%
        {wu2025sasw}
\bibfield{author}{\bibinfo{person}{Weilin Wu}, \bibinfo{person}{Shifan Yang},
  \bibinfo{person}{Qizhao Lin}, \bibinfo{person}{XingHong Chen},
  \bibinfo{person}{Kunping Yang}, \bibinfo{person}{Jing Wang}, {and}
  \bibinfo{person}{Guannan Chen}.} \bibinfo{year}{2025}\natexlab{}.
\newblock \showarticletitle{A Novel Perspective on Low-Light Image Enhancement:
  Leveraging Artifact Regularization and Walsh-Hadamard Transform}. In
  \bibinfo{booktitle}{\emph{ACM MM}}.
\newblock


\bibitem[Xie et~al\mbox{.}(2024a)]%
        {xie2024sana}
\bibfield{author}{\bibinfo{person}{Enze Xie}, \bibinfo{person}{Junsong Chen},
  \bibinfo{person}{Junyu Chen}, \bibinfo{person}{Han Cai},
  \bibinfo{person}{Haotian Tang}, \bibinfo{person}{Yujun Lin},
  \bibinfo{person}{Zhekai Zhang}, \bibinfo{person}{Muyang Li},
  \bibinfo{person}{Ligeng Zhu}, \bibinfo{person}{Yao Lu}, {and}
  \bibinfo{person}{Song Han}.} \bibinfo{year}{2024}\natexlab{a}.
\newblock \showarticletitle{Sana: Efficient High-Resolution Image Synthesis
  with Linear Diffusion Transformer}.
\newblock \bibinfo{journal}{\emph{arXiv preprint arXiv:2410.10629}}
  (\bibinfo{year}{2024}).
\newblock


\bibitem[Xie et~al\mbox{.}(2024b)]%
        {xie2024pltrans}
\bibfield{author}{\bibinfo{person}{Lianxin Xie}, \bibinfo{person}{Csbingbing
  Zheng}, \bibinfo{person}{Wen Xue}, \bibinfo{person}{Le Jiang},
  \bibinfo{person}{Cheng Liu}, \bibinfo{person}{Si Wu}, {and}
  \bibinfo{person}{Hau~San Wong}.} \bibinfo{year}{2024}\natexlab{b}.
\newblock \showarticletitle{Learning Degradation-Unaware Representation with
  Prior-Based Latent Transformations for Blind Face Restoration}. In
  \bibinfo{booktitle}{\emph{CVPR}}.
\newblock


\bibitem[Yang et~al\mbox{.}(2020)]%
        {yang2020hifacegan}
\bibfield{author}{\bibinfo{person}{Lingbo Yang}, \bibinfo{person}{Shanshe
  Wang}, \bibinfo{person}{Siwei Ma}, \bibinfo{person}{Wen Gao},
  \bibinfo{person}{Chang Liu}, \bibinfo{person}{Pan Wang}, {and}
  \bibinfo{person}{Peiran Ren}.} \bibinfo{year}{2020}\natexlab{}.
\newblock \showarticletitle{Hifacegan: Face renovation via collaborative
  suppression and replenishment}. In \bibinfo{booktitle}{\emph{ACM MM}}.
\newblock


\bibitem[Yang et~al\mbox{.}(2023)]%
        {yang2023pgdiff}
\bibfield{author}{\bibinfo{person}{Peiqing Yang}, \bibinfo{person}{Shangchen
  Zhou}, \bibinfo{person}{Qingyi Tao}, {and} \bibinfo{person}{Chen~Change
  Loy}.} \bibinfo{year}{2023}\natexlab{}.
\newblock \showarticletitle{{PGDiff}: Guiding Diffusion Models for Versatile
  Face Restoration via Partial Guidance}. In
  \bibinfo{booktitle}{\emph{NeurIPS}}.
\newblock


\bibitem[Yang et~al\mbox{.}(2022)]%
        {yang2022maniqa}
\bibfield{author}{\bibinfo{person}{Sidi Yang}, \bibinfo{person}{Tianhe Wu},
  \bibinfo{person}{Shuwei Shi}, \bibinfo{person}{Shanshan Lao},
  \bibinfo{person}{Yuan Gong}, \bibinfo{person}{Mingdeng Cao},
  \bibinfo{person}{Jiahao Wang}, {and} \bibinfo{person}{Yujiu Yang}.}
  \bibinfo{year}{2022}\natexlab{}.
\newblock \showarticletitle{{MANIQA}: Multi-dimension Attention Network for
  No-Reference Image Quality Assessment}. In \bibinfo{booktitle}{\emph{CVPRW}}.
\newblock


\bibitem[Yang et~al\mbox{.}(2021)]%
        {Yang2021GPEN}
\bibfield{author}{\bibinfo{person}{Tao Yang}, \bibinfo{person}{Peiran Ren},
  \bibinfo{person}{Xuansong Xie}, {and} \bibinfo{person}{Lei Zhang}.}
  \bibinfo{year}{2021}\natexlab{}.
\newblock \showarticletitle{GAN Prior Embedded Network for Blind Face
  Restoration in the Wild}. In \bibinfo{booktitle}{\emph{CVPR}}.
\newblock


\bibitem[Ying et~al\mbox{.}(2024)]%
        {ying2024restorerid}
\bibfield{author}{\bibinfo{person}{Jiacheng Ying}, \bibinfo{person}{Mushui
  Liu}, \bibinfo{person}{Zhe Wu}, \bibinfo{person}{Runming Zhang},
  \bibinfo{person}{Zhu Yu}, \bibinfo{person}{Siming Fu},
  \bibinfo{person}{Si-Yuan Cao}, \bibinfo{person}{Chao Wu},
  \bibinfo{person}{Yunlong Yu}, {and} \bibinfo{person}{Hui-Liang Shen}.}
  \bibinfo{year}{2024}\natexlab{}.
\newblock \showarticletitle{{RestorerID}: Towards Tuning-Free Face Restoration
  with ID Preservation}.
\newblock \bibinfo{journal}{\emph{arXiv preprint arXiv:2411.14125}}
  (\bibinfo{year}{2024}).
\newblock


\bibitem[Yu et~al\mbox{.}(2024)]%
        {yu2024supir}
\bibfield{author}{\bibinfo{person}{Fanghua Yu}, \bibinfo{person}{Jinjin Gu},
  \bibinfo{person}{Zheyuan Li}, \bibinfo{person}{Jinfan Hu},
  \bibinfo{person}{Xiangtao Kong}, \bibinfo{person}{Xintao Wang},
  \bibinfo{person}{Jingwen He}, \bibinfo{person}{Yu Qiao}, {and}
  \bibinfo{person}{Chao Dong}.} \bibinfo{year}{2024}\natexlab{}.
\newblock \showarticletitle{Scaling Up to Excellence: Practicing Model Scaling
  for Photo-Realistic Image Restoration In the Wild}. In
  \bibinfo{booktitle}{\emph{CVPR}}.
\newblock


\bibitem[Yu et~al\mbox{.}(2018)]%
        {yu2018super}
\bibfield{author}{\bibinfo{person}{Xin Yu}, \bibinfo{person}{Basura Fernando},
  \bibinfo{person}{Richard Hartley}, {and} \bibinfo{person}{Fatih Porikli}.}
  \bibinfo{year}{2018}\natexlab{}.
\newblock \showarticletitle{Super-resolving very low-resolution face images
  with supplementary attributes}. In \bibinfo{booktitle}{\emph{CVPR}}.
\newblock


\bibitem[Yue and Loy(2024)]%
        {yue2024difface}
\bibfield{author}{\bibinfo{person}{Zongsheng Yue} {and}
  \bibinfo{person}{Chen~Change Loy}.} \bibinfo{year}{2024}\natexlab{}.
\newblock \showarticletitle{{ DifFace: Blind Face Restoration with Diffused
  Error Contraction }}.
\newblock \bibinfo{journal}{\emph{IEEE TPAMI}} (\bibinfo{year}{2024}).
\newblock


\bibitem[Zeng et~al\mbox{.}(2023)]%
        {zeng2023mystylepp}
\bibfield{author}{\bibinfo{person}{Libing Zeng}, \bibinfo{person}{Lele Chen},
  \bibinfo{person}{Yi Xu}, {and} \bibinfo{person}{Nima~Khademi Kalantari}.}
  \bibinfo{year}{2023}\natexlab{}.
\newblock \showarticletitle{Mystyle++: A controllable personalized generative
  prior}. In \bibinfo{booktitle}{\emph{ACM SIGGRAPH Asia}}.
\newblock


\bibitem[Zhang et~al\mbox{.}(2025)]%
        {zhang2025instantrestore}
\bibfield{author}{\bibinfo{person}{Howard Zhang}, \bibinfo{person}{Yuval
  Alaluf}, \bibinfo{person}{Sizhuo Ma}, \bibinfo{person}{Achuta Kadambi},
  \bibinfo{person}{Jian Wang}, {and} \bibinfo{person}{Kfir Aberman}.}
  \bibinfo{year}{2025}\natexlab{}.
\newblock \showarticletitle{{InstantRestore}: Single-Step Personalized Face
  Restoration with Shared-Image Attention}. In
  \bibinfo{booktitle}{\emph{SIGGRAPH}}.
\newblock


\bibitem[Zhang et~al\mbox{.}(2018)]%
        {zhang2018lpips}
\bibfield{author}{\bibinfo{person}{Richard Zhang}, \bibinfo{person}{Phillip
  Isola}, \bibinfo{person}{Alexei~A. Efros}, \bibinfo{person}{Eli Shechtman},
  {and} \bibinfo{person}{Oliver Wang}.} \bibinfo{year}{2018}\natexlab{}.
\newblock \showarticletitle{The Unreasonable Effectiveness of Deep Features as
  a Perceptual Metric}. In \bibinfo{booktitle}{\emph{CVPR}}.
\newblock


\bibitem[Zhou et~al\mbox{.}(2022)]%
        {zhou2022codeformer}
\bibfield{author}{\bibinfo{person}{Shangchen Zhou},
  \bibinfo{person}{Kelvin~C.K. Chan}, \bibinfo{person}{Chongyi Li}, {and}
  \bibinfo{person}{Chen~Change Loy}.} \bibinfo{year}{2022}\natexlab{}.
\newblock \showarticletitle{Towards Robust Blind Face Restoration with Codebook
  Lookup TransFormer}. In \bibinfo{booktitle}{\emph{NeurIPS}}.
\newblock


\end{thebibliography}

\end{document}